\documentclass[lettersize,journal]{IEEEtran}
\usepackage{amsmath,amsfonts}
\usepackage{algorithmic}
\usepackage{array}
\usepackage{textcomp}
\usepackage{stfloats}

\usepackage{verbatim}
\usepackage{mathtools}
\usepackage{tensor}
\usepackage{bm}
\usepackage{colortbl}
\usepackage{xargs}
\usepackage{xcolor}  % Coloured text etc.
\usepackage[colorinlistoftodos,prependcaption,textsize=tiny]{todonotes}
\newcommandx{\Shida}[2][1=]{\todo[inline,linecolor=blue,backgroundcolor=blue!25,bordercolor=blue,#1]{#2}}
\newcommandx{\Sen}[2][1=]{\todo[inline,linecolor=red,backgroundcolor=red!25,bordercolor=red,#1]{#2}}
\newcommandx{\SenDraft}[2][1=]{\todo[inline,linecolor=red,backgroundcolor=green!25,bordercolor=red,#1]{#2}}

\newcolumntype{M}[1]{>{\centering\arraybackslash}m{#1}}

\usepackage{graphicx}
\usepackage{subfigure}
\usepackage{makecell}

\mathtoolsset{showonlyrefs}
\usepackage{url}
\usepackage{adjustbox}

\makeatletter
\DeclareRobustCommand{\iscircle}{\mathord{\mathpalette\is@circle\relax}}
\newcommand\is@circle[2]{%
  \begingroup
  \sbox\z@{\raisebox{\depth}{$\m@th#1\bigcirc$}}%
  \sbox\tw@{$#1\square$}%
  \resizebox{!}{\ht\tw@}{\usebox{\z@}}%
  \endgroup
}
\makeatother

\begin{document}
    \title{\fontsize{23}{27}\selectfont AQUA-SLAM: Tightly-Coupled Underwater Acoustic-Visual-Inertial SLAM with Sensor Calibration}
    \author{Shida Xu, Kaicheng Zhang and Sen Wang
    \thanks{The authors are with Department of Electrical and Electronic Engineering \& I-X, Imperial College London, SW7 2AZ London, U.K. (e-mail: \{s.xu23, k.zhang23, sen.wang\}@imperial.ac.uk).}
    \thanks{Shida Xu and Kaicheng Zhang are also with the School of Engineering and Physical Sciences, Heriot-Watt University, EH14 4AS Edinburgh, U.K.}
    \thanks{Corresponding author: Sen Wang}
    }

% \markboth{IEEE TRANSACTIONS ON ROBOTICS}%
% {How to Use the IEEEtran \LaTeX \ Templates}

\maketitle

\begin{abstract}
    Underwater environments pose significant challenges for visual Simultaneous Localization and Mapping (SLAM) systems due to limited visibility, inadequate illumination, and sporadic loss of structural features in images. Addressing these challenges, this paper introduces a novel, tightly-coupled Acoustic-Visual-Inertial SLAM approach, termed AQUA-SLAM, to fuse a Doppler Velocity Log (DVL), a stereo camera, and an Inertial Measurement Unit (IMU) within a graph optimization framework. Moreover, we propose an efficient sensor calibration technique, encompassing multi-sensor extrinsic calibration (among the DVL, camera and IMU) and DVL transducer misalignment calibration, with a fast linear approximation procedure for real-time online execution. The proposed methods are extensively evaluated in a tank environment with ground truth, and validated for offshore applications in the North Sea. The results demonstrate that our method surpasses current state-of-the-art underwater and visual-inertial SLAM systems in terms of localization accuracy and robustness. The proposed system will be made open-source for the community.
%        \Sen{To revise}
\end{abstract}


% \iffalse
\section{Introduction}
    \label{sec:introduction}

Autonomous Underwater Vehicles (AUVs) are critical tools in offshore applications and ocean science, offering the capability to operate autonomously in challenging and often hazardous underwater environments. These vehicles are indispensable for tasks such as seabed mapping, pipeline and cable inspections, biological and environmental monitoring, and the maintenance of underwater infrastructure. A key application area is the detailed visual inspection of subsea structures, including offshore wind turbine foundations, where precise localization and mapping are paramount for effective operation.
% Autonomous navigation is essential for AUVs, especially in underwater environments where accurate and robust positioning and perception capabilities are vital. The field of
Considering cameras are widely equipped on underwater robots, visual Simultaneous Localization and Mapping (SLAM) techniques emerge as natural solutions.

Yet, underwater environments pose substantial unique challenges to visual SLAM techniques. The rapid attenuation of light energy in water severely limits the visibility of optical camera sensors, especially in murky water conditions.
% Since light energy attenuates rapidly in water, the visibility of optical camera sensors is significantly limited underwater, especially in murky water conditions.
Moreover, underwater vision often suffers from poor lighting and blizzards of ``marine snow'' caused by small particles of organic matter in water, severely reducing image quality with  increased motion blur and dynamic image regions.
Additionally, ocean current disturbances to the robots can frequently push them away, which causes underwater structures to be occasionally outside the camera's field of view leading to intermittent loss of visual tracking.
Therefore, although visual SLAM techniques have recently made tremendous progress in terrestrial settings \cite{mur2017orb2,engel2017dso,luo2022hybrid}, their performance and robustness are inevitably compromised in underwater due to the complex and dynamic nature of aquatic environments.

% due to the scarcity of nearby  providing texture for visual tracking,
% heavily rely on the quality of images and the existence of visual texture, which
% Optical sensors, while effective on land, face significant limitations underwater due to rapid light energy attenuation, leading to drastically reduced effective ranges. Additionally, particulate matter in water introduces noise into sensor data, compromising the efficiency of vision-based SLAM systems. The underwater environment often features expansive areas with few discernible features and poor lighting, severely affecting image quality and impeding accurate pose estimation. These conditions, common in aquatic environments, pose substantial challenges for vision-based SLAM systems. In systems relying solely on vision, such as those described in \cite{mur2017orb2}, these visual impairments can lead to tracking failures.

% using optical and inertial sensors,
% underwater electromagnetic wave or LiDAR attenuate dramatically, limited propagation range

\begin{figure}
    \centering
    \includegraphics[width=1\linewidth,height=7.7cm]{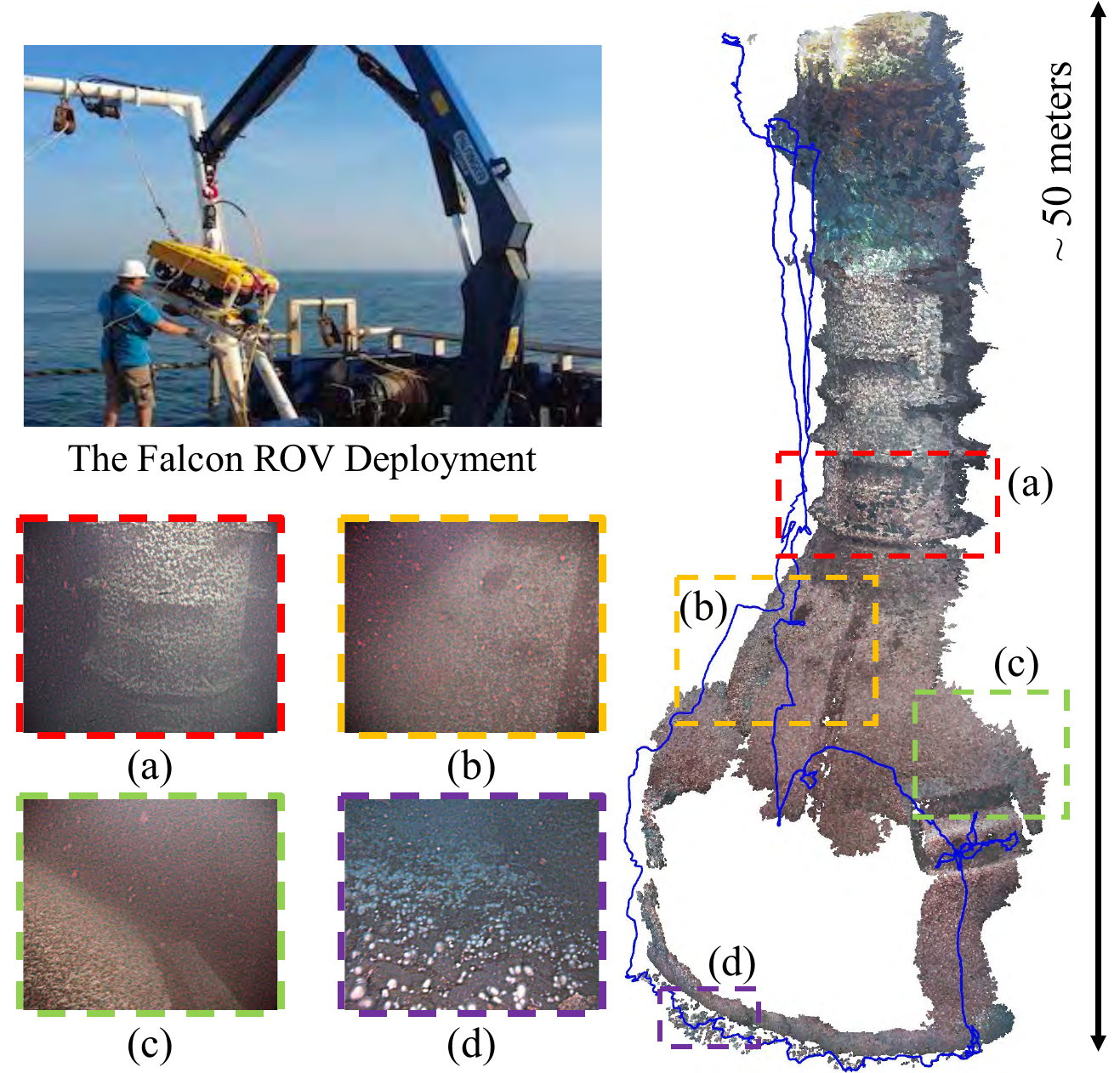}
    \caption{Estimated trajectory (\textcolor{blue}{\rule[.5ex]{1em}{1pt}}) and dense 3D reconstruction of an offshore structure using the proposed AQUA-SLAM algorithm. Images (a-d) show the challenging underwater conditions for a SLAM system using a camera.}
    \label{fig:offshore_big}
\end{figure}

Fusing visual SLAM with an Inertial Measurement Unit (IMU), known as visual-inertial SLAM (VI-SLAM) \cite{lin2018VINS-Mono,orbslam3}, can alleviate some of the challenges arising from transient, noise-affected visual inputs from an optical camera, such as momentary motion blur. Therefore, the accuracy and robustness of underwater SLAM systems, particularly against short-term visual disruptions, can be substantially enhanced \cite{rahman2022svin2}.
However, most of the challenges for underwater vision, such as the limited visibility and the ``marine snow'', are long-term effects that last at least from tens of seconds to a few minutes before being mitigated.
VI-SLAM also encounters its own set of problems underwater. The low signal-to-noise ratio in accelerometers' measurements and the need for double integration to estimate translation greatly amplify inherent noises, making VI-SLAM systems particularly vulnerable to unreliability in visually challenging underwater conditions.
Meanwhile, linear acceleration measurements are coupled with gravitational forces and IMU biases.
Therefore, a proper initialization process is required to estimate gravity direction and biases. However, this process often depends on optimal visual conditions and fully excited motion, which are difficult in underwater environments and for underwater vehicles moving against water. % Therefore, the robustness of VI-SLAM systems is significantly compromised in underwater environments.

% . Consequently, this technology has markedly progressed in addressing the SLAM challenge in most terrestrial applications, bringing the resolution of the terrestrial SLAM problem close to a fully solved state.

% and visual-inertial
% VI-SLAM integrates an Inertial Measurement Unit (IMU) to improve pose estimation, mitigating

To address these problems, this paper focuses on the sensor fusion of a stereo camera, an IMU and a Doppler Velocity Log (DVL), an underwater acoustic sensor measuring a linear velocity to the seabed \cite{Rudolph2012dvl_model} (see details on DVL in Section \ref{section:dvl_measurement_model}). % Its use of acoustic waves makes it capable of traversing significantly greater distances, thus offering an extended effective range.
%A DVL sensor % mainly provides linear velocity measurements, and it 
%is often used in a loosely-coupled manner as an odometry \cite{kim2013hull_inspection,vargas2021robust}. There is little literature on 1) investigating how to fully leverage DVL for 6 Degree-of-Freedom (DoF) underwater visual-inertial SLAM in a tightly-coupled fashion, while 2) addressing multi-sensor extrinsic calibration which is essential to fuse a sensor suite of a DVL, a camera and an IMU.
Current state-of-the-art approaches on fusing underwater localization sensors in the form of cameras, inertial and DVL sensors include 1) DVL integration with cameras without IMU \cite{huang2023tightly_visualdvl}, 2) filtering-based methods \cite{zhao2023tightly_icewater}, 3) DVL integration with LiDAR and visual SLAM for marine surface scenarios \cite{Thoms2023dvl_slam},  and 4) loosely-coupled integration of cameras, gyroscope and DVL, e.g., \cite{vargas2021robust}. To the best of our knowledge, a SLAM system with rigorous modelling of DVL and tightly-coupled fusion of DVL, camera, and IMU data within a graph optimization framework, specifically designed for underwater scenarios, has not been previously explored in the literature. Specifically, compared to our previous work \cite{vargas2021robust,xu2021underwater} which loosely fused cameras, DVL and gyroscope data, the new rigorous DVL modeling, tightly-coupled formulation, and accelerometer integration significantly improve performance. Furthermore, these enhancements enable more accurate and efficient extrinsic calibration and facilitate the calibration of DVL transducer orientation, which was not possible in the previous approach.

Therefore, this paper firstly models DVL's transducer and velocity measurements rigorously, and derives a tightly-coupled acoustic-visual-inertial graph optimization for underwater SLAM. % This method integrates data from a DVL, IMU, and camera within a graph optimization framework.
By making full use of the complementary strengths of these three sensing modalities, the approach aims to create a robust SLAM system capable of overcoming the challenges posed by underwater environments: the DVL provides reliable velocity measurements in underwater environments confining accelerometer-caused velocity drifts and enables dead-reckoning in visually degraded scenarios; the stereo camera offers high-accuracy localization capacity under good visual conditions and loop-closure for long-term drift correction; the IMU delivers reliable short-term motion estimation and renders absolute roll and pitch angles observable.
% \Shida{maybe consider how to turn to calibration topic naturely?}

Secondly, for the multi-sensor extrinsic calibration of DVL, camera and IMU, existing calibration mechanisms, such as hand-eye calibration \cite{strobl2006handeye}, %facilitate the determination of extrinsic calibration parameters between sensors. However, these mechanisms
are suboptimal in underwater scenarios for two primary reasons: 1) they often necessitate a pre-established configuration with a calibration pattern being consistently visible - a challenging or even unviable requirement in underwater environments; 2) the calibration accuracy of these mechanisms is often compromised, attributable to the loosely correlated nature of the process and the inherent degradation of underwater image quality. Furthermore, the extrinsic parameters are susceptible to variation over extended periods, such as several weeks, due to the continuous exposure to water drag forces and wave/current impacts. For users who have no dedicated expertise or underwater facilities to carry out sensor calibration, an automatic online calibration system only relying on features of surrounding scenes is more appealing.

Misalignment calibration of DVL's transducers is also of paramount importance to avoid errors in velocity measurements.
Recent works \cite{xu2022dvl_calibration,fu2022dvl_calibration} have attempted to calibrate a DVL sensor in the context of strapdown inertial navigation systems (SINS), concentrating on their extrinsic calibration. Notably, there is a lack of research exploring vision-facilitated calibration of DVL transducers.

To address these problems on the multi-sensor calibration and the DVL misalignment calibration, this paper proposes a novel online sensor calibration algorithm that is designed to calibrate the extrinsic parameters between DVL, camera and IMU, and to correct the alignment of DVL's transducers.

\subsection{Contributions}

The main contributions of this work include:
\begin{itemize}
    \item A novel underwater acoustic-visual-inertial SLAM algorithm, termed AQUA-SLAM, to tightly fuse DVL, camera and IMU sensors within a graph optimization framework. To the best of our knowledge, %our rigor modelling and pre-integration formulation of DVL measurements for tightly-coupled fusion with visual-inertial sensors are original 
    this is the first tightly-coupled graph-based SLAM system designed for underwater environments that integrates DVL, IMU, and camera data in a graph-based optimization framework.
    \item An efficient online sensor calibration algorithm for both DVL-camera-IMU extrinsic calibration and DVL transducer misalignment calibration, with a rapid linear approximation method designed to enable its real-time execution.
    % A distinctive feature of our approach is the integration of a rapid linear approximation procedure, which significantly accelerates the calibration process, enabling it to meet the demands of real-time execution.
    % \item A meticulously redesigned Acoustic-Visual-Inertial SLAM system, which is based on the ORB SLAM3 framework\cite{orbslam3}
    \item Extensive real-world experiments conducted in a water tank, along with offshore validation in the North Sea, demonstrating that our proposed method outperforms the state-of-the-art underwater and visual-inertial SLAM systems, and is viable for offshore applications (see result in Fig. \ref{fig:offshore_big}).
\end{itemize}
Our source code implementation will be released for the community upon the acceptance of the paper\footnote{\url{http://github.com/SenseRoboticsLab/AQUA-SLAM}}.
% The tank dataset with GT and corresponding tool set will be released in another dataset paper.
%\Sen{create gitpage and add the URL}

The rest of the paper is organized as follows: Section \ref{sec:relatedwork} reviews the literature on underwater visual SLAM and the sensor calibration, followed by a problem formulation in Section \ref{sec:preliminaries}. Section \ref{section:Camera DVL IMU Fusion} and \ref{sec:Extrinsic Calibration} describe the proposed AQUA-SLAM and sensor calibration algorithms, respectively. System implementation is detailed in Section \ref{sec:system_imp}. Experiment evaluation is presented in Section \ref{sec:experiments} before drawing conclusions in Section \ref{sec:cons}.

%\Sen{To revise}

\section{Related Work}\label{sec:relatedwork}
In this section, we review three topics related to our work: underwater visual SLAM, underwater vision based extrinsic sensor calibration and DVL calibration.

\subsection{Underwater Visual SLAM}

\subsubsection{Methods Using DVL}
In the early years, most research on underwater visual SLAM modeled DVL measurements as odometry, rather than directly incorporating the sensor's raw data. Eustice et al. \cite{underwater_vslam_ekf1} introduced a sensor fusion framework integrating navigation data with 5-DoF relative pose measurements % from cameras 
for vehicle motion in underwater environments, utilizing an augmented state Kalman filter. Ozog et al. proposed a SLAM method employing a sparse point cloud derived from a DVL, based on a piecewise-planar model \cite{ozog2013dvl_camera_piecewise_planar}. The underwater visual SLAM system proposed in \cite{kim2013hull_inspection} was based on a pose graph framework with DVL modeled as odometry constraints for hull inspection. This approach was extended to employ piecewise-planar panels for 3D reconstruction of curved ship hull surfaces \cite{underwater_vslam_graph2}. Fiducial markers were also incorporated into a visual SLAM framework alongside DVL, IMU, and depth sensor in \cite{westman2018camera_dvl_extrinsic}. 

Visual SLAM system integrating with DVL has gained more attention recently. A DVL, a stereo camera and a gyroscope were fused in a loosely-coupled fashion in \cite{vargas2021robust,xu2021underwater}, enabling reasonable pose estimation in challenging underwater environments. However, its DVL was still used in a loosely-coupled manner. Meanwhile, it did not incorporate an accelerometer and assumed zero bias for the gyroscope, resulting in unbounded roll and pitch estimates. Thoms et al. tightly integrated DVL into a LiDAR-visual-inertial SLAM system for Unmanned Surface Vehicles (USV) \cite{Thoms2023dvl_slam}. However, their method is designed for USVs operating on 2D water surface and may not be directly applicable to underwater environments. A tightly-coupled multi-sensor fusion framework for camera, IMU, DVL and a depth sensor, based on the Multi-State Constraint Kalman Filter (MSCKF), was proposed in \cite{zhao2023tightly_icewater}. It is a filtering based method, different from our graph optimization based method in this work. Huang et al. proposed a tightly-coupled visual-DVL fusion method which integrates the velocity measurements from a DVL into a visual odometry for improved localization accuracy \cite{huang2023tightly_visualdvl}. However, the lack of IMU integration might compromise the robustness of orientation estimation in challenging visual conditions. Importantly, none of these works addressed the sensor calibration problem simultaneously.
% Unlike \cite{ozog2013dvl_camera_piecewise_planar}, which models very sparse point cloud generated by DVL as piecewise-planar,

In contrast, the approach proposed in this paper focuses on a tightly-coupled integration of DVL, IMU and camera data, formulating the problem as a graph optimization to achieve accurate and robust localization and mapping in challenging underwater scenarios. To the best of our knowledge, this is the first tightly-coupled graph-based SLAM system that integrates DVL, IMU, and camera data in a unified framework for underwater environments. Furthermore we model the sensor calibration problem simultaneously as an online calibration module, which is essential for multi-sensor fusion.
 %While \cite{kim2013hull_inspection} explored , our research concentrates on the integration of DVL within a graph-based SLAM framework and sensor calibration.
% In contrast to \cite{westman2018camera_dvl_extrinsic}, which employs fiducial markers for pose estimation, our method utilizes natural scene features. Our previous work \cite{xu2021underwater}
%  In contrast, the proposed method integrates DVL, cameras, and an IMU, effectively avoiding accumulated drift in roll and pitch. Compared with \cite{Thoms2023dvl_slam} which is designed for USVs, our method mainly focus on underwater environment.

\subsubsection{Methods Without Using DVL}
Recent works attempted to enhance the accuracy and robustness of underwater visual SLAM by integrating other sensing modalities. Rahman et al. proposed the sonar visual inertial (SVIN) SLAM system \cite{rahman2018svin}, which integrated a downward-facing mechanical scanning sonar, a stereo camera and an IMU under a tightly-coupled framework based on OKVIS \cite{okvis}. They extended this work to SVIN2 \cite{rahman2022svin2,rahman2019svin2} which additionally included a water-pressure depth sensor and a loop closure module. More recently, Joshi et al. proposed a state estimation switching strategy that can detect failures in SVIN2 and seamlessly transition to a model-based approach using the robot's kinematics and proprioceptive sensors to maintain pose estimation \cite{joshi2023switching}. However, none of these methods investigate the incorporation of DVL data.

\subsection{Underwater Vision Based Extrinsic Calibration}
As previously discussed, extrinsic sensor calibration is vital for multi-sensor fusion, especially for underwater scenarios. Camera-IMU extrinsic calibration for underwater environments was studied in \cite{gu2019camera_imu_cali_underwater} and \cite{yang2020camera_sonar_extrinsic} when sonar data was available. However, camera-DVL calibration is rarely explored, with \cite{westman2018camera_dvl_extrinsic} being one of the only few existing works. However, its reliance on a marinized panel of fiducial markers and pre-setup made it impractical or time-consuming in offshore underwater settings. Our proposed calibration algorithm, in contrast, simply utilizes natural scene features for automatic online sensor calibration, being seamlessly integrated with our proposed SLAM algorithm.

\subsection{DVL Calibration}
Regarding DVL calibration, existing research predominantly focuses on SINS-DVL systems rather than vision-DVL systems. Xu et al. \cite{xu2022dvl_calibration} introduced an EKF-based method for calibrating DVL installation and scale factor using the Special Orthogonal Group. Li et al. \cite{li2022dvl_calibration} presented a DVL calibration algorithm employing particle swarm optimization, transforming the calibration issue into a Wahba problem. Luo et al. \cite{Luo2023dvl_calibration} proposed a SINS-DVL calibration system capable of calibrating various errors and misalignments, utilizing observability-based trajectory design for parameter observability. However, none of these addressed DVL transducer misalignment calibration which can have considerable impacts on its velocity measurements and its fusion with other sensors. % Our proposed algorithm, based on a tightly-coupled Acoustic-Visual-Inertial model, can calibrate DVL transducer misalignment.

%    \Sen{Add some literature on this as it is one of the contributions}

\section{Preliminaries} \label{sec:preliminaries}

\begin{figure}
    \centering
    \includegraphics[width=0.9\columnwidth]{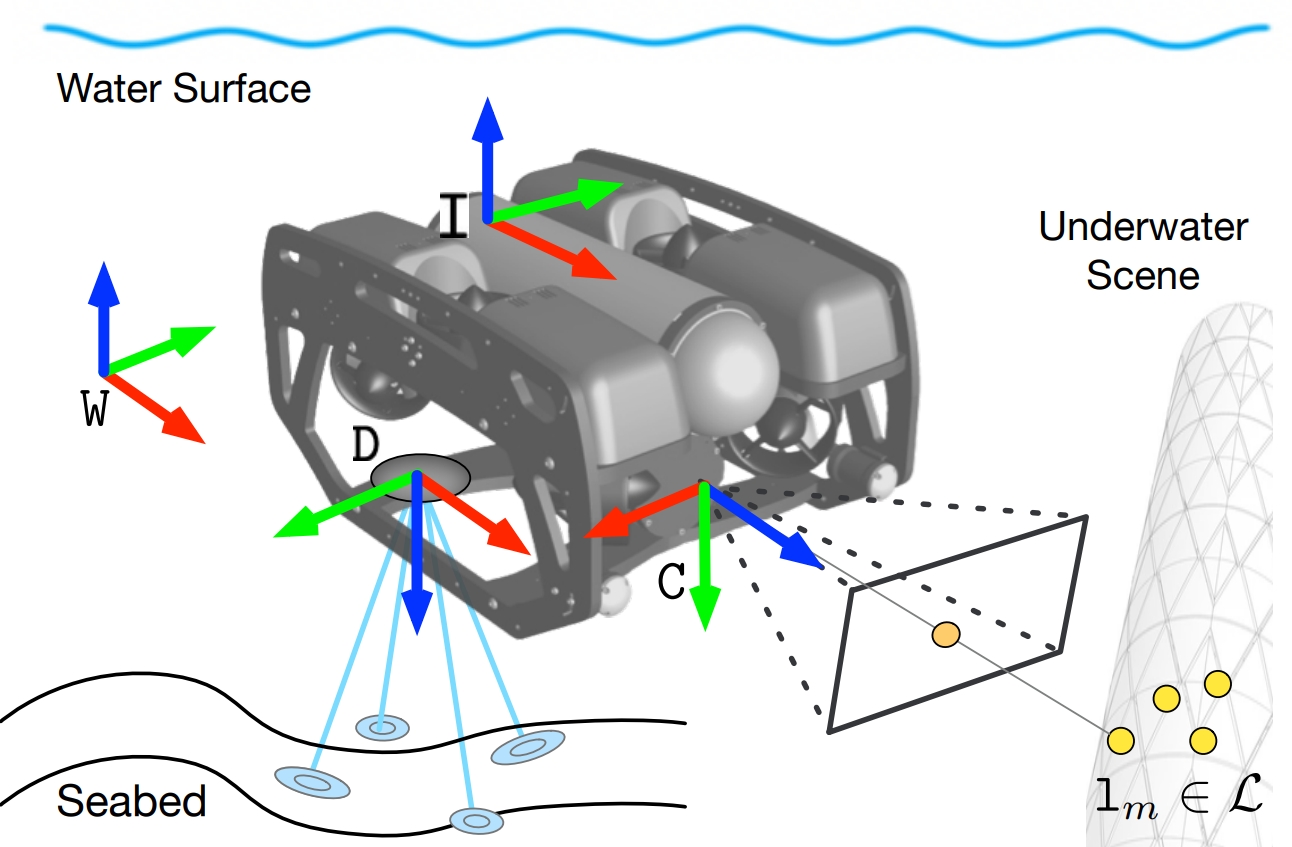}
    \caption{Coordinate frames. The world frame $\mathtt{W}$'s z-axis is aligned with the gravity vector. The DVL frame $\mathtt{D}$, the IMU frame $\mathtt{I}$ and the camera frame $\mathtt{C}$ are rigidly fixed on the robot. The DVL sensor measures linear velocity with respect to the seabed. 3D visual landmarks $\mathcal{L}$ are estimated from scenes. }
%     \includegraphics[width=0.97\columnwidth]{images/sensor_frame.pdf}
%     \caption{Coordinate frames. The world frame, denoted as $\mathtt{W}$, its z-axis is oriented in alignment with the gravitational vector. The initial camera frame, represented by $\mathtt{C}_0$, signifies the point of initialization for the SLAM system. Concurrently, $\mathtt{C}_i$ designates the current camera frame. The notations $\mathtt{D}_i$ and $\mathtt{I}_i$ respectively correspond to the current frames of the DVL and the IMU. The extrinsic parameters between the IMU, DVL and camera are represented by $\mathbf{T}_{\mathtt{I}\mathtt{D}}$ and $\mathbf{T}_{\mathtt{D}\mathtt{C}}$. $_{\mathtt{C}_0}\mathbf{l}_{\mathtt{C}_0\mathtt{l}_m}$ stands for the 3D location of landmark $m$ in frame $\mathtt{C}_0$.}
%     \Shida{remove background, make figure more tight, add background to text}
    \label{fig:sensor_frames}
\end{figure}

We use the following notation throughout this paper. A scalar is lowercase italics $a$, a vector is lowercase bold Roman  $\mathbf{a}$, a matrix is uppercase bold Roman $\mathbf{A}$, a coordinate frame is typewriter $\mathtt{A}$, and a set is calligraphic typeface $\mathcal{A}$.
% a time-dependent quantities are expanded with subscripts $()_i$.

\subsection{Riemannian Geometry Representation}
% 3D geometry is represented using Special Orthogonal Group $\text{SO}(3)$ and the Special Euclidean Group $\text{SE}(3)$, collectively referred to as manifolds. During the process of pose optimization, we choose to work on their local tangent space Lie algebra, specifically $\mathfrak{so}(3)$ and $\mathfrak{se}(3)$. This approach makes the mathematical handling of rotations and translations more convenient and effective, leading to improved results.\cite{sola2018microLie}

The group of 3D rotation matrices is described by Special Orthogonal Group $\text{SO}(3)$ as
$\text{SO}(3) \doteq \{\mathbf{R} \in \mathbb{R}^{3\times3} \mid \mathbf{R} \mathbf{R}^T= \mathbf{I},det(\mathbf{R})=1 \}$.
Its tangent space is denoted as $\mathfrak{so}(3)$,
%  is the  of $\text{SO}(3)$, and for each Lie group $\text{SO}(3)$ we can find an associated $\mathfrak{so}(3)$.
which can be represented as a $3\times3$ skew-symmetric matrix from a hat operator $(\cdot)^{\land}$ on a $3 \times 1$ vector, i.e.,
% \begin{equation}
    $\mathfrak{so}(3) \doteq \{ \bm{\phi}^{\land} \in \mathbb{R}^{3\times3} \big| \bm{\phi} \in \mathbb{R}^3 \}$.
% \end{equation}

The group of 3D rigid motion belongs to the Special Euclidean Group
$\text{SE}(3) \doteq
\left\{  \left( \mathbf{R}, \mathbf{t} \right) \big| \mathbf{R} \in \text{SO}(3), \mathbf{t} \in \mathbb{R}^3   \right\} $
whose corresponding transformation matrix is defined as $\mathbf{T}$.
% tangent space is $\mathfrak{se}(3) = \{ \bm{\xi}^{\land} \in \mathbb{R}^{4\times4} | \bm{\xi} \in \mathbb{R}^{6} \}$.
%  is defined as
% \begin{equation}
%     \mathfrak{se}(3) = \{ \mathbf{\xi}^{\land} \in \mathbb{R}^{4\times4} | \mathbf{\xi} \in \mathbb{R}^{6} \}
% \end{equation}
% which is a $4\times4$ matrix whose up-left corner is a skew-symmetric matrix.
% We define the hat operator $(\cdot)^{\land}$ for $\mathfrak{se}(3)$ as:
% \begin{equation}
%     \bm{\xi}^{\land} = \left[ \begin{matrix}
%        \bm{\rho} \\  \bm{\phi}
%     \end{matrix}  \right]^{\land} =
%     \left[ \begin{matrix}
%          \bm{\phi}^{\land} & \bm{\rho} \\
%          \mathbf{0}&0
%     \end{matrix}  \right] \quad \bm{\rho}, \bm{\phi} \in \mathbb{R}^3
% \end{equation}
%\Sen{remove Lie algebra definiation if not used}
Following the notations suggested in \cite{furgale2014}, a 3D transformation from a coordinate frame $\mathtt{B}$ to a coordinate frame $\mathtt{A}$ is defined as
\begin{equation}
    \mathbf{T}_{\mathtt{A}\mathtt{B}} \doteq \left[\begin{matrix} \mathbf{R}_{\mathtt{A}\mathtt{B}} & _\mathtt{A}\mathbf{p}_{\mathtt{A}\mathtt{B}}  \\
                                                                       \mathbf{0} & 1
                             \end{matrix}\right] \in \text{SE}(3)
    \label{equa:1}
\end{equation}
    where $\mathbf{R}_{\mathtt{A}\mathtt{B}} \in \text{SO}(3)$ describes its 3D rotation matrix and $_\mathtt{A}\mathbf{p}_{\mathtt{A}\mathtt{B}} \in \mathbb{R}^3$ is its 3D translation expressed in frame $\mathtt{A}$.

\subsection{Definition of Coordinate Frames}

Fig. \ref{fig:sensor_frames} specifies the coordinate frames relevant to this work, including the static world frame $\mathtt{W}$, the IMU frame $\mathtt{I}$, the camera frame $\mathtt{C}$ and the DVL frame $\mathtt{D}$. The z-axis of the world frame is aligned with the gravity direction.
% with its z-axis pointing upwards, aligned with the direction of gravity.
% The yaw and position of the world frame are aligned with the first
    A time-dependent moving coordinate frame is specified with a subscript, e.g., $\mathtt{C}_i$ means the camera coordinate frame $\mathtt{C}$ of keyframe $i$ .
% This consistent frame notation will be used throughout the remaining sections of this paper for clarity and conciseness.

% The current camera pose can be represented as $^{c_0}\mathbf{T}_{c_i}$. Considering the states $^{c_0}\mathbf{T}_{c_{i-1}}$ and $^{c_0}\mathbf{T}_{c_i}$ of previous and current camera frames $c_{i-1}$ and $c_i$ with respect to the initial camera frame $c_{0}$, their relative transformation can be calculated as follows:
% \begin{equation}
%     ^{c_{i-1}}{\mathbf{T}}_{c_i} = ^{c_0}\mathbf{T}_{c_{i-1}}^{-1}\
%     ^{c_0}\mathbf{T}_{c_i}
%     \label{equa:2}
% \end{equation}

% The fixed transformations between different sensors, a.k.a extrinsic parameters, are represented as follows:
% \begin{itemize}
%     \item From the DVL to the camera: $\mathbf{T}_{\mathtt{C}\mathtt{D}}$.
%     \item From the IMU to the camera: $\mathbf{T}_{\mathtt{C}\mathtt{B}}$.
%     \item From the DVL to the IMU: $\mathbf{T}_{\mathtt{I}\mathtt{D}}$.
% \end{itemize}

% \Shida{add Exp Log map}

\subsection{State Estimation}

\subsubsection{State Definition}

    The state of keyframe $i$ is defined as
\begin{equation}
\mathbf{x}_{i} \doteq [ \mathbf{R}_{i}, \mathbf{p}_{i}, \mathbf{v}_{i}, \mathbf{b}^g_{i}, \mathbf{b}^a_{i} ] \in \text{SO}(3) \times \mathbb{R}^{12}
\label{equa:state_timei}
\end{equation}
%    \Shida{not sure about $\text{SO}(3) \times \mathbb{R}^{12}$ maybe use a $\mathbb{R}^{15}$ instead}
    where $(\mathbf{R}_{i}, \mathbf{p}_{i}) \in \text{SE}(3)$ denotes the 3D camera pose $\mathbf{T}_{i}$ in $\mathtt{C}_0$ which stands for the initial camera frame, i.e., $(\mathbf{R}_{\mathtt{C}_0\mathtt{C}_i}, {_{\mathtt{C}_0}\mathbf{p}_{\mathtt{C}_0\mathtt{C}_i}})$, while $\mathbf{v}_{i} \doteq {_{\mathtt{D}_i}\mathbf{v}} \in \mathbb{R}^3$ represents the linear velocity in $\mathtt{D}$. $\mathbf{b}^g_{i}$ and $\mathbf{b}^a_{i}$ are the IMU gyroscope and accelerometer biases. %in $\mathtt{I}$.

Considering all keyframes $\mathcal{K}_n$ up to $n$, the set of historical keyframe states, landmarks and calibration parameters is defined as
\begin{equation}
\mathcal{X}_n \doteq \{\mathbf{x}_{i}, \mathcal{L}_i, \mathcal{E}, \mathbf{R}_{\mathtt{W}\mathtt{I}_0} \}, \quad i  \in \mathcal{K}_n
\label{equa:state_def}
\end{equation}
    where $\mathcal{L}_i \doteq \{ _{\mathtt{C}_0}\mathbf{l}_{\mathtt{C}_0\mathtt{l}_m} \}, m \in \mathcal{M}_i$ is the set of 3D locations of landmarks visible in keyframe $i$, $\mathcal{E} \doteq \{\mathbf{T}_{\mathtt{I}\mathtt{D}}, \mathbf{T}_{\mathtt{D}\mathtt{C}}\}$ includes the fixed extrinsic parameters between the sensors (IMU, DVL and camera), and $\mathbf{R}_{\mathtt{W}\mathtt{I}_0}$ is the initial orientation of the IMU (accelerometer) in $\mathtt{W}$ standing for the gravity direction.
    Notably, the extrinsic parameter $\mathbf{T}_{\mathtt{I}\mathtt{C}} $ can be derived by the transformation composition using $\mathcal{E}$:
    \begin{equation}
        \mathbf{T}_{\mathtt{I}\mathtt{C}} \doteq \mathbf{T}_{\mathtt{I}\mathtt{D}} \mathbf{T}_{\mathtt{D}\mathtt{C}}
        \label{equa:extrinsic_IC}
    \end{equation}
    Therefore, we do not define it explicitly in the state.

%    \SenDraft{transform composition?}

% where $\mathcal{X} \doteq \{^{c_0}\mathbf{R}_{c_n},^{c_0}\mathbf{p}_{c_0\_c_n} \}_{n \in \mathcal{K}}$ is camera poses under the first camera frame $c_0$ at all the given timestamps $\mathcal{K}$.

% For simplicity, a quantity with respect to $\mathtt{W}$ at time $i$ will be shortened when no ambiguity, e.g., $\mathbf{T}_{\mathtt{W}\mathtt{C}_i} \doteq $

% $\mathcal{L} \doteq \{^{c_0}\mathbf{l}_{k}\}_{k\in\mathcal{M}} $ is a set of 3D location of all the observable landmarks $\mathcal{M}$ under the first camera frame $c_0$.
% $}$ and $^{d}{\mathbf{T}}_{c}$ are the transformation matrix from DVL frame to the IMU local frame and the transformation from camera frame to the DVL frame respectively. Hence, the transformation from the camera to the IMU frame can be calculated by
% \begin{equation}
%     \begin{aligned}
%     &^{b}{\mathbf{T}}_{c} = ^{b}{\mathbf{T}}_{d} \  ^{d}{\mathbf{T}}_{c}\\
%     &^{c}{\mathbf{T}}_{b} = ^{b}{\mathbf{T}}_{c}^{-1}
%     \end{aligned}
%     \label{equa:state_extrinsic2}
% \end{equation}

\subsubsection{Measurement Definitions}
    Measurements provided by a DVL, an IMU and a camera between keyframes $i$ and $j$ are defined as follows.

    \paragraph{DVL} The raw measurements from the DVL are obtained by decoding acoustic signals emanated from its transducers. By utilizing the Doppler shift principle, these measurements facilitate the calculation of velocity. The collection of the DVL measurements is denoted as $ \mathcal{D}_{i,j}$.
% linear velocities \doteq \{ {\hat{\mathbf{v}}}_t \}_{t \in (i,j)}$

    \paragraph{IMU} The IMU measurements $\mathcal{I}_{i,j}$ are a set of rotational velocity ${\tilde{\bm{\omega}}}$ and linear acceleration ${\tilde{\mathbf{a}}}$.
% \doteq \{ {\tilde{\mathbf{\omega}}}_t, {\tilde{\mathbf{a}}}_t \}_{t \in [i,j]}

    \paragraph{Camera} The camera yields a pair of stereo images for $i$th keyframe $\mathcal{C}_{i}$, from which landmarks are extracted.
%    \SenDraft{The camera provides a pair of stereo images $\mathcal{C}_{i}$ ????? extracting landmarks.}

    To summarise, the set of all sensor measurements up to keyframe $n$ is
% \begin{equation}
    $\mathcal{Z}_{n} \doteq \{\mathcal{D}_{i,j},\mathcal{I}_{i,j}, \mathcal{C}_{i}\}, i,j \in \mathcal{K}_n$.
% \end{equation}
The detailed DVL, IMU and camera measurement models will be discussed in Section \ref{section:dvl_measurement_model}, \ref{section:imu_measurement_model} and \ref{section:camera_measurement_model}, respectively.
%

% A camera is a pair of stereo images. In practice, we can ascertain the two-dimensional
% are pixel locations of detected landmarks on camera images: % from the camera at time $i$ can be represented as all the pixel positions of all observable landmarks:
% $\mathcal{C}_{i} \doteq \{ ^{i}\mathbf{u}_{k} \}_{k \in \mathcal{M}_i}$.

% Given a time interval from $i$ to $j$, the DVL measurements can be represented as a set of velocities, denoted as: .

\subsubsection{Maximum a Posteriori (MAP) Estimation}
% Given a initial state $\mathcal{X}_0$ and all the measurement $\mathcal{Z}_{0:k}$ until time $k$, our goal is to estimate the state $\mathcal{X}$. We consider all the measurements and states as random variables. The goal is to maximum the posterior $p(\mathcal{X}| \mathcal{Z}_{0:k})$. According to Bayes Rule, instead of maximum the posterior, we can maximum the likelihood, and we assume the measurement are independent:

% an initial state set $\mathcal{X}_0$ and
Given the full set of measurements $\mathcal{Z}_n$, our goal is to estimate the optimal $\mathcal{X}_n$ by maximizing the posterior probability:
\begin{gather}
    \small
\begin{aligned}
    &\mathcal{X}_n^*  = \mathop{\mathrm{argmax}}_{\mathcal{X}_n} \ p(\mathcal{X}_n | \mathcal{Z}_{n})\\
    & = \mathop{\mathrm{argmax}}_{\mathcal{X}_n} \ p(\mathcal{X}_{0}) \prod_{i,j \in \mathcal{K}_n} p( \mathcal{D}_{i,j} | \mathcal{X}_{i},\mathcal{X}_{j}) p( \mathcal{I}_{i,j} | \mathcal{X}_{i},\mathcal{X}_{j}) \prod_{i \in \mathcal{K}_n} p(\mathcal{C}_{i} | \mathcal{X}_{i})
\end{aligned}
% \raisetag{40pt}
\end{gather}
Since the measurements are usually assumed  with Gaussian noises, this can be re-formulated as a non-linear least-squares problem:
\begin{gather}
    \small
\begin{aligned}
    & \mathcal{X}_n^* = \mathop{\mathrm{argmin}}_{\mathcal{X}_n} \ \| \mathbf{r}_{x_0}\|_{\Sigma_{0}}^2 + \sum_{i,j \in \mathcal{K}_n} \bigg( \bigg. \|\mathbf{r}_{D}\left(h_{D}\left(\mathcal{X}_i,\mathcal{X}_j\right),\mathcal{D}_{i,j}\right)\|_{\mathbf{\Sigma}_{D_{}}}^2  \\
    &  + \|\mathbf{r}_{I}(h_{I}(\mathcal{X}_i,\mathcal{X}_j),\mathcal{I}_{i.j})\|_{\mathbf{\Sigma}_{I_{}}}^2 \bigg. \bigg)
 + \sum_{i \in \mathcal{K}_n} \| \mathbf{r}_{C}(h_{C}(\mathcal{X}_i),\mathcal{C}_i)\|_{\mathbf{\Sigma}_{C}}^2
\end{aligned}
\label{equa:opt_implicit}
\raisetag{20pt}
\end{gather}
The notation $\| \cdot \|^2_{\Sigma}$ represents the Mahalanobis distance, with $\Sigma$ denoting covariance.
The prior residual  $\mathbf{r}_{x_0}$ computes the discrepancy between the initial state and the prior information. The three residuals $\mathbf{r}_{D}$, $\mathbf{r}_{I}$, and $\mathbf{r}_{C}$ represent the DVL, IMU and camera residuals, respectively. Each of these residuals quantifies the errors between the sensor measurements and the predicted measurements from the corresponding measurement models $h_{D}(\cdot)$, $h_{I}(\cdot)$ and $h_{C}(\cdot)$, which are to be detailed now.
% We will provide an explicit form of these measurement models and definitions of residuals in section \ref{section:Camera DVL IMU Fusion}.
%    Residual $\mathbf{r}_{x_0}$ is the initial state prior.
% This non-linear least-squares problem can be solved utilizing either the Gauss-Newton method or the Levenberg-Marquardt method.

% Rather than maximizing the likelihood, we minimize the negative logarithm of the likelihood.

% \subsubsection{Optimization \textbf{add}}
% where Given the measurement model of camera $h_{C}(\mathcal{X}_i)$, DVL $h_{D}(\mathcal{X}_i,\mathcal{X}_j)$, and IMU $h_{I}(\mathcal{X}_i,\mathcal{X}_j)$,

\section{Acoustic-Visual-Inertial Underwater SLAM} \label{section:Camera DVL IMU Fusion}

\begin{figure}
    \centering
    \includegraphics[width=0.97\columnwidth]{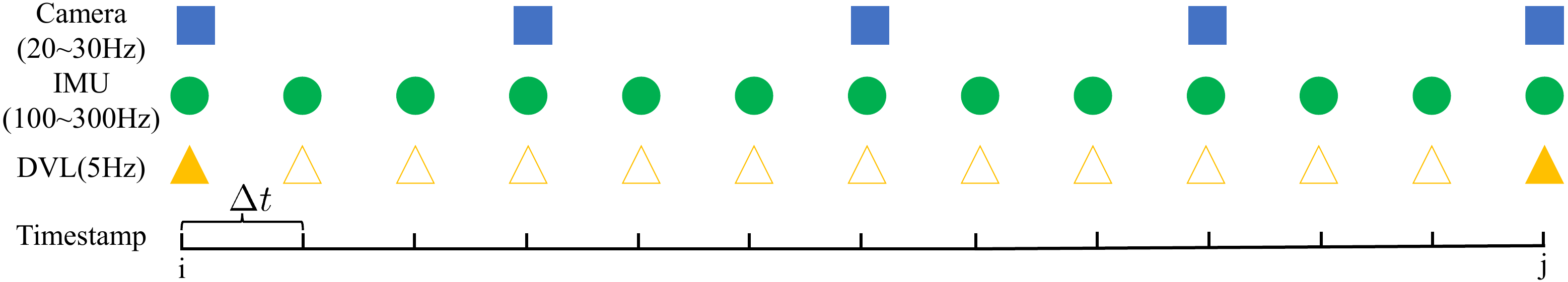}
    \caption{Sensor measurements from camera, IMU and DVL. Dash yellow triangles represent constant velocity taken from the last DVL measurement.}
    \label{fig:sensor_rate}
\end{figure}

    In this section, the sensor measurement models and the residual terms in \eqref{equa:opt_implicit} are derived for the proposed acoustic-visual-inertial underwater SLAM system, given the DVL, IMU and camera measurements between image keyframes $i$ and $j$ as shown in Fig. \ref{fig:sensor_rate}. The extrinsic parameters $\mathcal{E}$, which are estimated using the method proposed in Section \ref{sec:Extrinsic Calibration}, are assumed known in this section.
    %Given two image keyframes $i$ and $j$ with the set of IMU and DVL measurements inbetween, as shown in Fig. \ref{fig:sensor_rate}, we describe the method model and its residuals can be defined as follows.
    % Given two image keyframes, denoted as $i$ and $j$, accompanied by the corresponding set of IMU and DVL measurements between them, as illustrated in Fig. \ref{fig:sensor_rate}, the subsequent sections delineate the measurement model and expound on how the residuals can be formulated.
% We describe in a two-frame scenario, from which a multi-frame case can be directly extended to.

% the objective here is to estimate the state at times $i$ and $j$.
% In practical terms, key frames are selected based on the DVL rate to ensure that there is at least one DVL measurement present between every two adjacent key frames.

% The focus will then shift to model the state and the measurements. This relationship will subsequently be used to construct the residual definition and build a non-linear optimization problem, allowing for iterative optimization of the state.

\subsection{DVL Measurement Model and Its Residuals}
\label{section:dvl_measurement_model}

    A DVL sensor has the capability to measure linear velocity with respect to the seabed. Typically, it encompasses four transducers, as the example in Fig. \ref{fig:dvl_mount_example}. Each transducer is oriented towards the seabed and continuously emits acoustic signals. These signals, upon reflection from the seabed, are sampled to measure Doppler shifts and then the one-dimensional velocity along the direction of each transducer. % can be determined based on the corresponding Doppler shift.
    By aggregating each individual transducer's velocity, the overall velocity in the 3D space can be obtained \cite{Brokloff1994dvl_model}. It is a common practice to mount the DVL at the base of an underwater vehicle facing the seabed. % as shown in Fig. \ref{fig:dvl_mount_example}.

\subsubsection{Velocity Measurement from an Individual Transducer}
    Each transducer of the DVL operates as both a transmitter and a receiver, able to emit and receive acoustic signals. It is hypothesized that its transmitted signal possesses a frequency $f_t$, while the received signal has a frequency $f_r$. Therefore, the frequency shift is $\Delta f = f_r - f_t$.
% The transducer acts as both speaker and listener, which twice the frequency shift.
    According to the Doppler shift, the velocity along the radial direction of the $n$th transducer is $v_n = 2c_s\Delta f/f_t$ where $c_s$ is the sound speed in water \cite{Brokloff1994dvl_model}. 
    Therefore, the one-dimensional velocity measurement $\tilde{v}_n$ of an individual transducer along its radial direction is assumed below with Gaussian noise:
    \begin{equation}
        {\tilde{v}}_n = v_n + \eta^D, \quad n \in \{1, 2, 3, 4\}
        \label{equa:velocty_noise}
    \end{equation}
    where $\eta^D \sim \mathcal{N}(0,{\sigma}^D)$

\subsubsection{DVL Velocity Measurement and Its Model}
    \label{section:dvl_velocity_model}
\begin{figure}
        \centering    \includegraphics[width=1\columnwidth]{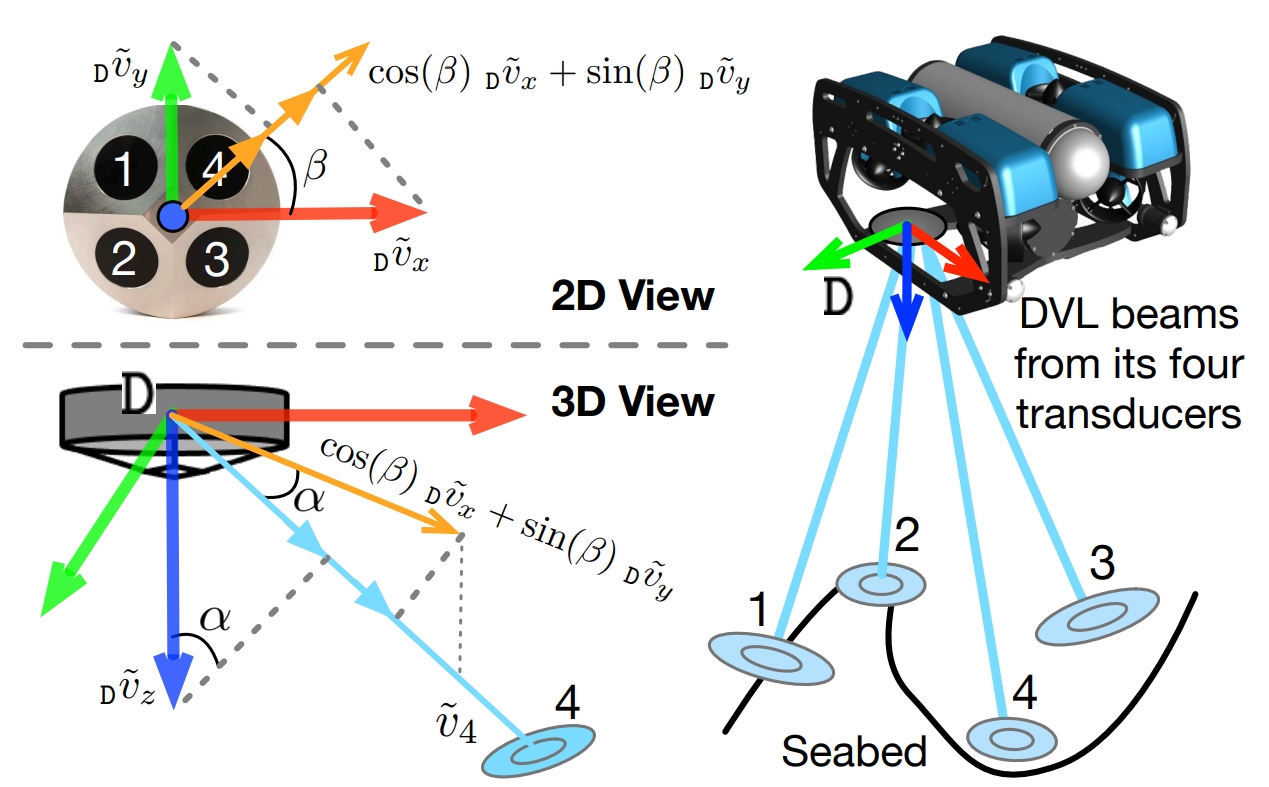}
        \caption{DVL transducer measurements with 2D and 3D views. The DVL has 4 transducers facing different directions. Transducer 4 is shown as an example.}
        \label{fig:dvl_mount_example}
    \end{figure}

    The DVL velocity measurement, associated with keyframe $i$ in the DVL frame $\mathtt{D}_i$, is denoted as ${_{\mathtt{D}_i}\tilde{\mathbf{v}}} \in \mathbb{R}^3$. It is correlated with the individual transducer's velocity $\tilde{v}_n$ through
\begin{equation}
        \tilde{v}_n = \mathbf{e}_n \cdot {_{\mathtt{D}_i}\tilde{\mathbf{v}}}, \quad n \in \{1, 2, 3, 4\}
\label{equa:v_d_v_i}
\end{equation}
    where $\mathbf{e}_n \in \mathbb{R}^{1 \times 3}$ signifies the orientation-dependent factor of the transducer $n$ and projects ${_{\mathtt{D}_i}\tilde{\mathbf{v}}}$ to the radial direction of transducer $n$, as shown in Fig. \ref{fig:dvl_mount_example}.
% As shown in Fig. \ref{fig:DVL_body} (b), the x,y,z axis represents the DVL local frame, $V_4$ is the velocity measured by transducer 4 which is a scalar. Then we project x-y-z component to the transducer direction.
% For the z-axis component in Fig. \ref{fig:DVL_body} (b), we can easily project it to the transducer direction which is ${{_{\mathtt{D}}\mathbf{v}}}_z \,  \sin(\alpha)$. For the x-axis and y-axis components, we need to get the direction of the transducer projection on x-y plane which is the yellow arrow in Fig. \ref{fig:DVL_body}. First we project x y component to its x-y plane projection which is ${{_{\mathtt{D}}\mathbf{v}}}_x \, \cos(\beta)  + {{_{\mathtt{D}}\mathbf{v}}}_y \, \sin(\beta)$. Then project the x-y plane projection to the transducer direction. Finally, we can get the relationship between $\mathbf{V}_4$ and local frame velocity:
% the x, y, and z axes represent the DVL local frame, with $V_4$ being the scalar velocity measured by transducer 4. Following this, the x-y-z components are projected onto the direction of the transducer.
    % Now we derive $\mathbf{e}_n$ based on the coordinate frames and velocities in Fig. \ref{fig:dvl_mount_example}.
    Assume the radial direction of each transducer is rotated by an angle $\alpha$ from horizontal plane and a yaw $\beta$ with respect to the DVL frame $\mathtt{D}_i$. Then, the z velocity ${_{\mathtt{D}}\tilde{v}_z}$ of ${_{\mathtt{D}_i}\tilde{\mathbf{v}}}$ can be projected onto the transducer direction by $\sin(\alpha)$. For the x and y velocities, ${_{\mathtt{D}}\tilde{v}_x}$ and ${_{\mathtt{D}}\tilde{v}_y}$ of ${_{\mathtt{D}_i}\tilde{\mathbf{v}}}$, it is necessary to establish their components along the transducer's x-y projection (see 2D view in Fig. \ref{fig:dvl_mount_example}). % Initially, the x-y components are projected onto their x-y plane projection.
    This aggregated x-y plane velocity is further projected to the transducer direction. Taking transducer 4 as an example, we can determine % the relationship between the transducer's velocity $\tilde{v}_4$ and ${_{\mathtt{D}_i}\tilde{\mathbf{v}}}$:
\begin{equation}
        \tilde{v}_4 =\mathbf{e}_4\cdot {_{\mathtt{D}_i}\tilde{\mathbf{v}}} =[\cos(\beta) \cos(\alpha) \  \sin(\beta) \cos(\alpha) \  \sin(\alpha)] \ {_{\mathtt{D}_i}\tilde{\mathbf{v}}}
 \label{equa:v_4_vd}
\end{equation}
% Each transducer is rotated the same degrees, but is facing different directions by default. Similarly we can get all the other transducer projection vector:
Since the transducer often shares the same amount of rotation by design, we can similarly obtain the projection vectors $\mathbf{e}_n$ for other transducers as
%    \SenDraft{or after DVL misalignment calibration at section VI}
\begin{equation}
\begin{aligned}
 \mathbf{e}_1
 &=[-\cos(\beta) \cos(\alpha) \  &\sin(\beta) \cos(\alpha) \  &\sin(\alpha)] \\
  \mathbf{e}_2
 &=[-\cos(\beta) \cos(\alpha) \  &-\sin(\beta) \cos(\alpha) \  &\sin(\alpha)] \\
  \mathbf{e}_3
 &=[\cos(\beta) \cos(\alpha) \  &-\sin(\beta) \cos(\alpha) \  &\sin(\alpha)] \\
  \mathbf{e}_4
 &=[\cos(\beta) \cos(\alpha) \  &\sin(\beta) \cos(\alpha) \  &\sin(\alpha)]
 \label{equa:e_vector1}
 \end{aligned}
\end{equation}
After vectorizing the velocity measurements from all the transducers, we can be obtain
\begin{equation}
\begin{aligned}
        [\tilde{v}_1 \ \tilde{v}_2 \ \tilde{v}_3 \ \tilde{v}_4]^T = \mathbf{E} \cdot {_{\mathtt{D}_i}\tilde{\mathbf{v}}}
\label{equa:e_full}
\end{aligned}
\end{equation}
where $\mathbf{E} \doteq \left[\begin{matrix}
    \mathbf{e}_1 ;\ \mathbf{e}_2; \ \mathbf{e}_3;\ \mathbf{e}_4
    \end{matrix}\right] \in \mathbb{R}^{4\times3}$.
% \begin{equation}
%     \begin{aligned}
%     \mathbf{E} \doteq \left[\begin{matrix}
%     \mathbf{e}_1\ & \mathbf{e}_2\ & \mathbf{e}_3 & \mathbf{e}_4
%     \end{matrix}\right]^T \in \mathbb{R}^{4\times3}
%     \label{equa:e_vector2}
%     \end{aligned}
% \end{equation}
% The local frame velocity can get from Equa (\ref{equa:e_full}) which has 4 equations and 3 unknow. It is a over-estimated problem. And it's just a linear equation, so it can be easily solved. The closed form solution is:
    Therefore, ${_{\mathtt{D}_i}\tilde{\mathbf{v}}}$ can be derived from \eqref{equa:e_full} which contains four equations and three unknowns. It presents an over-determined problem with linear equations. Therefore the closed-form solution of ${_{\mathtt{D}_i}\tilde{\mathbf{v}}}$ from the individual transducer velocity measurements is
\begin{equation}
\begin{aligned}
        {_{\mathtt{D}_i}\tilde{\mathbf{v}}} = (\mathbf{E}^T \, \mathbf{E})^{-1} \, \mathbf{E}^T [\tilde{v}_1 \ \tilde{v}_2 \ \tilde{v}_3 \ \tilde{v}_4]^T
\label{equa:dvl_velocity}
\end{aligned}
\end{equation}
    Since we assume each individual $\tilde{v}_n$ is corrupt by Gaussian noise, after a linear transformation ${_{\mathtt{D}_i}\tilde{\mathbf{v}}}$ still follow Gaussian distribution:
    \begin{equation}
        {_{\mathtt{D}_i}\mathbf{v}} = {_{\mathtt{D}_i}\tilde{\mathbf{v}}} - \bm{\eta}^D
    \end{equation}
    where $\bm{\eta}^D \sim \mathcal{N}(0,\mathbf{\Sigma}^D)$.

    The DVL velocity measurement model for keyframes $i$ and $j$ can be expressed as
    \begin{equation}
        % \begin{aligned}
            h_{D_v}(\mathcal{X}_i) \doteq {_{\mathtt{D}_i}\tilde{\mathbf{v}}} - \bm{\eta}^D, \quad
            h_{D_v}(\mathcal{X}_j) \doteq {_{\mathtt{D}_j}\tilde{\mathbf{v}}} - \bm{\eta}^D
        \label{equa:dvl_velocity2}
        % \end{aligned}
    \end{equation}

%    Under this assumption, the measurement model for the DVL velocity, corresponding to keyframe $i$, can be expressed as:
%%    \Shida{In IMU model sention, we use measuremnt minus noise, but in DVL section we use state plus noise}
%\begin{equation}
%    \begin{aligned}
%            & h_{D_v}(\mathcal{X}_i) \doteq \mathbf{v}_{i} + \bm{\eta}^D
%    \label{equa:dvl_velocity2}
%    \end{aligned}
%\end{equation}
%
%    Similarly the the measurement model for the DVL velocity, corresponding to keyframe $j$, can be expressed as:
%\begin{equation}
%    \begin{aligned}
%        & h_{D_v}(\mathcal{X}_j) \doteq \mathbf{v}_{j} + \bm{\eta}^D
%        \label{equa:dvl_velocity3}
%    \end{aligned}
%\end{equation}
% where  is the and ${{_{\mathtt{D}}\hat{\mathbf{v}}}} \doteq  [\mathbf{E}^T \, \mathbf{E}]^{-1} \, \mathbf{E}^T [V_1 \ V_2 \ V_3 \ V_4]^T$is the measured DVL local frame velocity.
% The DVL can measure the velocity in DVL local frame via the Doppler Shift. We will introduce more details \textbf{in the section}. We just assume the the velocity measurement from DVL is affected by a Gaussain noise here:

% \begin{equation}
%     \prescript{d}{}{\mathbf{V}} = \prescript{d}{}{\hat{\mathbf{V}}} - \mathbf{\eta}^d
% \end{equation}

\subsubsection{DVL Translation Measurement Model}

    The DVL measurements, in conjunction with the orientation estimates using the gyroscope measurements, can constrain the translation estimate between keyframes $i$ and $j$.
% It is assumed that the DVL and gyroscope are synchronized.

    %The velocity is assumed to be constant between consecutive DVL measurements, and we integrate the DVL velocity into the translation when a gyroscope measurement is received. Therefore, as illustrated in Fig. \ref{fig:DVL_vs_DVLgyro}, we calculate the DVL translation between every consecutive IMU measurements, illustrated by the short green vector. The translation between keyframe $i$ to keyframe $j$, represented by ${_\mathtt{W}\mathbf{p}_{\mathtt{D}_i\mathtt{D}_j}}$ (the long green vector in Fig. \ref{fig:DVL_vs_DVLgyro}), can be calculated by summing all the DVL translations between the two keyframes. Hence, we have the following relation:

    It is posited that the velocity remains consistent between two consecutive DVL measurements. Upon receiving a gyroscope measurement, the DVL velocity is integrated into the translation. When determining the translation from keyframe $i$ to keyframe $j$, denoted as ${_{\mathtt{C}_0}\mathbf{p}_{\mathtt{D}_i\mathtt{D}_j}}$ it is essential to aggregate all the DVL translations occurring between these two keyframes. This leads to the subsequent relationship:
% We integrate the DVL velocity into translation at the rate of IMU. As shown in the Fig. \ref{fig:DVL_vs_DVLgyro}, we calculate the translation $^{c_{0}}{\mathbf{p}}_{d_k\_d_{k+\Delta t}}$ between every two IMU measurements, which is represented by the short green vector. Then the translation from time $i$ to time $j$ $^{c_{0}}\mathbf{p}_{d_i\_d_j}$, which is the long green vector in the Fig. \ref{fig:DVL_vs_DVLgyro}, can be calculated by adding all the ${c_{0}}{\mathbf{p}}_{d_k\_d_{k+\Delta t}}$ together during this period.  So we have:
\begin{equation}
    % \begin{aligned}
    % ^{c_0}{\mathbf{p}}_{d_j}&=^{c_0}{\mathbf{p}}_{d_i} + {}^{c_0}{\mathbf{p}}_{d_i\_d_j}
    {_{\mathtt{C}_0}\mathbf{p}_{{\mathtt{C}_0}\mathtt{D}_j}} = {_{\mathtt{C}_0}\mathbf{p}_{{\mathtt{C}_0}\mathtt{D}_i}} + {_{\mathtt{C}_0}\mathbf{p}_{\mathtt{D}_i\mathtt{D}_j}}
    % &=^{c_0}{\mathbf{p}}_{d_i} + \sum_{k=i}^{j-1} {^{c_0}{\mathbf{p}}_{d_k\_d_{k+\Delta t}}}\\
    % \end{aligned}
    \label{equa:dvl_intg1}
\end{equation}
    where the terms ${_{\mathtt{C}_0}\mathbf{p}_{{\mathtt{C}_0}\mathtt{D}_i}}$ and ${_{\mathtt{C}_0}\mathbf{p}_{{\mathtt{C}_0}\mathtt{D}_j}}$ are defined as the DVL positions at keyframes $i$ and $j$ respectively. They can be represented as: ${_{\mathtt{C}_0}\mathbf{p}_{{\mathtt{C}_0}\mathtt{D}_i}} \doteq \mathbf{p}_{i} - \mathbf{R}_{i} \mathbf{R}_{\mathtt{D}\mathtt{C}}^T {_\mathtt{D}\mathbf{p}_{\mathtt{D}\mathtt{C}}} $ and ${_{\mathtt{C}_0}\mathbf{p}_{{\mathtt{C}_0}\mathtt{D}_j}} \doteq \mathbf{p}_{j} - \mathbf{R}_{j} \mathbf{R}_{\mathtt{D}\mathtt{C}}^T {_\mathtt{D}\mathbf{p}_{\mathtt{D}\mathtt{C}}}$ respectively.
    Additionally, the term ${_{\mathtt{C}_0}\mathbf{p}_{\mathtt{D}_i\mathtt{D}_j}}$ denotes the integration of DVL translations.
    This is formally expressed as ${_{\mathtt{C}_0}\mathbf{p}_{\mathtt{D}_i\mathtt{D}_j}} \doteq \sum_{k=i}^{j-1} \mathbf{R}_{i} \mathbf{R}_{\mathtt{I}\mathtt{C}}^T {\Delta\mathbf{R}_{\mathtt{I}_i\mathtt{I}_k}} \mathbf{R}_{\mathtt{I}\mathtt{D}}\ {(_{\mathtt{D}_i}{\tilde{\mathbf{v}}}} - \bm{\eta}^D) \Delta t$
    where $\Delta t$ is the time interval between the gyroscope (IMU) measurements, and the variable ${\Delta\mathbf{R}_{\mathtt{I}_i\mathtt{I}_k}}$ represents the gyroscope relative incremental defined in \eqref{equa:gyros_integration2}.
    %A comprehensive elaboration on gyroscope integration is provided in section \ref{section:imu_measurement_model}.
% Given Equa (\ref{equa:dvl_velocity2}) and (\ref{equa:dvl_intg1}), we already can define the DVL residual. But we want avoid keeping doing the integration during the optimization. So we need to define a relative translation term to store the result of Equa (\ref{equa:dvl_intg1}). In addition, we also need to decouple the state from the relative translation term. Therefore we reformulate Equa (\ref{equa:dvl_intg1}) and Equa (\ref{equa:dvl_def}) as:

%Given \eqref{equa:dvl_intg1}, we are ready to define the DVL translation residual. However, ${_\mathtt{W}\mathbf{p}_{\mathtt{D}_i\mathtt{D}_j}}$ is coupled with state $\mathbf{R}_{i}$ which may vary during optimization. In order to avoid repeated integrations during the optimization process, we define a DVL pre-integration translation term to decouple the state from the relative translation term. To this end, we reformulate \eqref{equa:dvl_intg1} as follows:

%Considering \eqref{equa:dvl_intg1}, we introduce the concept of the DVL translation residual.
Notably, the term ${_{\mathtt{C}_0}\mathbf{p}_{\mathtt{D}_i\mathtt{D}_j}}$ is interrelated with the state $\mathbf{R}_{i}$, which can cause repeated DVL translation integration during optimization iterations. To address this problem, we propose a DVL pre-integration term decoupling the state from the DVL translation integration. Hence, we reformulate \eqref{equa:dvl_intg1} as
\begin{gather}
     \begin{aligned}
% &\mathbf{R}_{\mathtt{I}\mathtt{D}}({_\mathtt{D}\mathbf{p}_{\mathtt{D}\mathtt{C}}}-\mathbf{R}_{\mathtt{D}\mathtt{C}}\mathbf{R}_{i}^T\mathbf{R}_{j}\mathbf{R}_{\mathtt{D}\mathtt{C}}^T{_\mathtt{D}\mathbf{p}_{\mathtt{D}\mathtt{C}}}+ \\ & \mathbf{R}_{\mathtt{D}\mathtt{C}}(\mathbf{R}_{i}^T\mathbf{p}_{j}-\mathbf{R}_{i}^T\mathbf{p}_{i}))
%h_{D_t}(\mathcal{X}_i,\mathcal{X}_j)
    & \underbrace{\mathbf{R}_{\mathtt{I}\mathtt{D}}\big({_\mathtt{D}\mathbf{p}_{\mathtt{D}\mathtt{C}}}-\mathbf{R}_{\mathtt{D}\mathtt{C}}\mathbf{R}_{i}^T\mathbf{R}_{j}\mathbf{R}_{\mathtt{D}\mathtt{C}}^T{_\mathtt{D}\mathbf{p}_{\mathtt{D}\mathtt{C}}} + \mathbf{R}_{\mathtt{D}\mathtt{C}} (\mathbf{R}_{i}^T\mathbf{p}_{j}-\mathbf{R}_{i}^T\mathbf{p}_{i})\big)}_\text{$h_{D_t}(\mathcal{X}_i,\mathcal{X}_j)$} \\
    & = \Delta {_{\mathtt{D}_i}}{\bar{\mathbf{p}}}_{\mathtt{D}_i\mathtt{D}_j} - \delta {_{\mathtt{D}_i}}{\bar{\mathbf{p}}}_{\mathtt{D}_i\mathtt{D}_j}
     %    \\& = \underbrace{{\sum_{k=i}^{j-1}} {\Delta\hat{\mathbf{R}}_{\mathtt{I}_i\mathtt{I}_k}} \mathbf{R}_{\mathtt{I}\mathtt{D}}\ {_{\mathtt{D}_i}{\tilde{\mathbf{v}}}} \Delta t}_\text{$\Delta {_{\mathtt{D}_i}}{\bar{\mathbf{p}}}_{\mathtt{D}_i\mathtt{D}_j}$} - \underbrace{{\sum_{k=i}^{j-1}} {\Delta\hat{\mathbf{R}}_{\mathtt{I}_i\mathtt{I}_k}} \mathbf{R}_{\mathtt{I}\mathtt{D}}\ \bm{\eta}^D \Delta t}_\text{$\delta {_{\mathtt{D}_i}}{\bar{\mathbf{p}}}_{\mathtt{D}_i\mathtt{D}_j}$}
     \end{aligned}
    \label{equa:dvl_decoupled}
    % \raisetag{-10pt}
\end{gather}
    where $h_{D_t}(\mathcal{X}_i,\mathcal{X}_j)$ stands for the DVL translation measurement model. $\Delta {_{\mathtt{D}_i}}{\bar{\mathbf{p}}}_{\mathtt{D}_i\mathtt{D}_j} \doteq {\sum_{k=i}^{j-1}} {\Delta\hat{\mathbf{R}}_{\mathtt{I}_i\mathtt{I}_k}} \mathbf{R}_{\mathtt{I}\mathtt{D}}\ {_{\mathtt{D}_i}{\tilde{\mathbf{v}}}} \Delta t$ represents the DVL translation pre-integration and $\Delta{\hat{\mathbf{R}}}_{\mathtt{I}_i\mathtt{I}_k}$ stands for the gyroscope pre-integration from keyframe $i$ to $k$ defined at \eqref{equa:gyros_integration2}. $\delta {_{\mathtt{D}_i}}{\bar{\mathbf{p}}}_{\mathtt{D}_i\mathtt{D}_j} \doteq \sum_{k=i}^{j-1}  -{\Delta{\hat{\mathbf{R}}}_{\mathtt{I}_i\mathtt{I}_k}} (\mathbf{R}_{\mathtt{I}\mathtt{D}}\ {_{\mathtt{D}_i}{\tilde{\mathbf{v}}}})^\land \delta{\bm{\hat{\phi}}}_{\mathtt{I}_i\mathtt{I}_k} \Delta t + {\Delta{\hat{\mathbf{R}}}_{\mathtt{I}_i\mathtt{I}_k}} \mathbf{R}_{\mathtt{I}\mathtt{D}}\ {\bm{\eta}^D} \Delta t$ stands for the Gaussian noise.
%where the right side defined as ${_\mathtt{W}}{\hat{\mathbf{p}}}_{\mathtt{D}_j\mathtt{D}_i}$
%  \doteq \sum_{k=i}^{j-1} \prescript{b_i}{}{\mathbf{R}}_{b_k} \mathbf{R}_{\mathtt{I}\mathtt{D}} \prescript{d_k}{}{\hat{\mathbf{V}}} \Delta t_{k\_k+1}$
 %is the relative translation measured by DVL, and
%\begin{equation}
%    \begin{aligned}
%h_{D_t}(\mathcal{X}_i,\mathcal{X}_j) \doteq & \mathbf{R}_{\mathtt{I}\mathtt{D}}\big({_\mathtt{D}\mathbf{p}_{\mathtt{D}\mathtt{C}}}-\mathbf{R}_{\mathtt{D}\mathtt{C}}\mathbf{R}_{i}^T\mathbf{R}_{j}\mathbf{R}_{\mathtt{D}\mathtt{C}}^T{_\mathtt{D}\mathbf{p}_{\mathtt{D}\mathtt{C}}}\\
%& + \mathbf{R}_{\mathtt{D}\mathtt{C}}(\mathbf{R}_{i}^T\mathbf{p}_{j}-\mathbf{R}_{i}^T\mathbf{p}_{i})\big)
%\end{aligned}
%\end{equation}
%is the DVL translation measurement model.
Please see Appendix \ref{sec:appendix DVL Translation Measurement Model Derivation} and \ref{sec:appendix DVL Preintegration Derivation} for detailed derivation of this DVL pre-integration.
% Assume the measured velocity is corrupted by a Gaussian noise $\mathbf{\eta}_d$, then we have the  DVL velocity measurement model
%\Sen{Add derivation in Appendix, double check equations here}

\subsubsection{Residual Derivation}
%We can now derive the explicit formulation of the DVL residual in \eqref{equa:opt_implicit}, which is defined as the errors between the DVL measurement and the
The explicit formulation of the DVL residual in \eqref{equa:opt_implicit} can be derived as
\begin{equation}
    % \begin{aligned}
    \mathbf{r}_{D} \doteq
    \left[ \begin{matrix}
        \mathbf{r}_{v}(h_{D_v}(\mathcal{X}_i),\mathcal{D}_{i}) \\
        \mathbf{r}_{v}(h_{D_v}(\mathcal{X}_j),\mathcal{D}_{j}) \\
        \mathbf{r}_{t}(h_{D_t}(\mathcal{X}_i,\mathcal{X}_j),\mathcal{D}_{i,j})
    \end{matrix} \right]
    % \end{aligned}
    \label{equa:dvl_measurement_model}
\end{equation}
where $\mathbf{r}_{v}(\cdot)$ constrains the velocities at keyframe $i$ and $j$, while $\mathbf{r}_{t}(\cdot)$ constrains the relative translation between the keyframes.
In accordance with \eqref{equa:dvl_velocity} and \eqref{equa:dvl_velocity2}, the DVL velocity residual can be defined as follows:
\begin{equation}
    \begin{aligned}
    \mathbf{r}_{v}(h_{D_v}(\mathcal{X}_i),\mathcal{D}_{i}) & \doteq {_{\mathtt{D}_i}\tilde{\mathbf{v}}} - \mathbf{v}_{i}\\
    \mathbf{r}_{v}(h_{D_v}(\mathcal{X}_j),\mathcal{D}_{j}) & \doteq {_{\mathtt{D}_j}\tilde{\mathbf{v}}} - \mathbf{v}_{j}
    \end{aligned}
    \label{equa:dvl_velocity_residual1}
\end{equation}
According to \eqref{equa:dvl_decoupled}, the DVL translation residual can be obtained as
\begin{equation}
    \mathbf{r}_{t}(h_{D_t}(\mathcal{X}_i,\mathcal{X}_j),\mathcal{D}_{i,j})
     \doteq  {\Delta {_{\mathtt{D}_i}}{\bar{\mathbf{p}}}_{\mathtt{D}_i\mathtt{D}_j}} - h_{D_t}(\mathcal{X}_i,\mathcal{X}_j)
    %  \mathbf{R}_{\mathtt{I}\mathtt{D}}\bigl({_\mathtt{D}\mathbf{p}_{\mathtt{D}\mathtt{C}}} \\
    % & -\mathbf{R}_{\mathtt{D}\mathtt{C}} \mathbf{R}_{i}^T\mathbf{R}_{j}\mathbf{R}_{\mathtt{D}\mathtt{C}}^T {_\mathtt{D}\mathbf{p}_{\mathtt{D}\mathtt{C}}} +  \mathbf{R}_{\mathtt{D}\mathtt{C}}(\mathbf{R}_{i}^T\mathbf{p}_{j}-\mathbf{R}_{i}^T\mathbf{p}_{i})\bigr)
    \label{equa:dvl_residual}
\end{equation}

\subsection{IMU Measurement Model and Its Residuals}\label{section:imu_measurement_model}
\subsubsection{Gyroscope Model}
    The gyroscope measures the instantaneous angular velocity of $\mathtt{I}$ relative to $\mathtt{W}$ expressed in the IMU coordinate frame $\mathtt{I}$. We assume the measurement ${_\mathtt{I}{\tilde{\bm{\omega}}_{\mathtt{W}\mathtt{I}}}} \in \mathbb{R}^3$ is affected by a zero-mean Gaussian white noise $\bm{\eta}^g$ and random walk noise $\mathbf{b}^g$:
\begin{equation}
    {_\mathtt{I}{\tilde{\bm{\omega}}_{\mathtt{W}\mathtt{I}}}}= {_\mathtt{I}\bm{\omega}_{\mathtt{W}\mathtt{I}}}  +\mathbf{b}^g + \bm{\eta}^g
\end{equation}
where $\bm{\eta}^g \sim \mathcal{N}(0,\mathbf{\Sigma}^g)$

Given $\mathcal{X}_i$ at keyframe $i$ and the extrinsic parameters  $\mathcal{E}$, along with the series of the gyroscope measurements between keyframes $i$ and $j$, %we assume the angular velocity remains constant between time intervals. Hence,
the gyroscope orientation at keyframe $j$, ${\mathbf{R}}_{\mathtt{W}\mathtt{I}_j}$, is computed as follows:
\begin{equation}
    \begin{aligned}
    &{\mathbf{R}}_{\mathtt{W}\mathtt{I}_j} = \underbrace{ {\mathbf{R}}_{\mathtt{W}\mathtt{I}_i} \prod_{k=i}^{j-1} \text{Exp}\big(( _{\mathtt{I}_{k}}{\tilde{\bm{\omega}}_{\mathtt{W}\mathtt{I}_k}} -\mathbf{b}^g_i-\bm{\eta}^g) \Delta t\big)}_\text{${\mathbf{R}}_{\mathtt{W}\mathtt{I}_k}$}
    \end{aligned}
    \label{equa:gyros_integration}
\end{equation}
% \begin{equation}
%     \begin{aligned}
%     &\mathbf{R}_{j} [\mathbf{R}_{\mathtt{I}\mathtt{D}} \mathbf{R}_{\mathtt{D}\mathtt{C}}]^T \\ &={\mathbf{R}}_{i} [\mathbf{R}_{\mathtt{I}\mathtt{D}} \mathbf{R}_{\mathtt{D}\mathtt{C}}]^T \prescript{b_i}{}{\mathbf{R}}_{b_j}\\
%     % &=^{g0}{}\mathbf{R}_{b_i} \prod_{k=i}^{j-1} Exp(^{b}{}{\mathbf{\omega}}_k \Delta \mathbf{t})\\
%     &={\mathbf{R}}_{i} [\mathbf{R}_{\mathtt{I}\mathtt{D}} \mathbf{R}_{\mathtt{D}\mathtt{C}}]^T \prod_{k=i}^{j-1} Exp((_\mathtt{I}{\hat{\mathbf{\omega}}_k}-\mathbf{b}^g_i) \Delta t)
%     \end{aligned}
%     \label{equa:gyros_integration}
% \end{equation}
where ${\mathbf{R}}_{\mathtt{W}\mathtt{I}_j} \doteq \mathbf{R}_{\mathtt{W}\mathtt{I}_0} \mathbf{R}_{\mathtt{I}\mathtt{C}} \mathbf{R}_{j} {\mathbf{R}_{\mathtt{I}\mathtt{C}}^T}$,  ${\mathbf{R}}_{\mathtt{W}\mathtt{I}_i} \doteq \mathbf{R}_{\mathtt{W}\mathtt{I}_0} \mathbf{R}_{\mathtt{I}\mathtt{C}} {\mathbf{R}}_{i} {\mathbf{R}_{\mathtt{I}\mathtt{C}}^T}$ are the IMU orientations at keyframe $j$ and $i$ respectively, $\text{Exp}(\cdot): \mathbb{R}^3 \rightarrow \text{SO}(3)$ stands for the exponential map from a vectorized $\mathfrak{so}(3)$ to $\text{SO}(3)$, and $\mathbf{b}^g_i$ stands for the gyroscope bias at $i$.
% $\Delta{\hat{\mathbf{R}}}_{\mathtt{I}_i\mathtt{I}_k}$ is gyroscope pre-integration between keyframes $i$ and $j$.
%and $\prod_{k=i}^{j-1} Exp((_\mathtt{I}{\hat{\mathbf{\omega}}_k}-\mathbf{b}^g_i-\mathbf{\eta}^g) \Delta t)$ is the result of  which we can considere as the rotation measured by the gyroscope from time i to time j.

\subsubsection{Accelerometer Model}
The accelerometer measures linear acceleration with respect to the IMU frame $\mathtt{I}$, where the measurement $_\mathtt{I}{\tilde{\mathbf{a}}} \in \mathbb{R}^3$ is consistently influenced by gravity. We assume that it is also affected by zero-mean Gaussian white noise $\bm{\eta}^a$ and a random walk noise $\mathbf{b}^a$:
\begin{equation}
    _\mathtt{I}{\tilde{\mathbf{a}}} ={\mathbf{R}}_{\mathtt{W}\mathtt{I}}^T({_\mathtt{W}\mathbf{a}}-{_\mathtt{W}\mathbf{g}})+\mathbf{b}^a + \bm{\eta}^a
 \end{equation}
where ${_\mathtt{W}\mathbf{g}} \doteq [0,0,-g]^T$ is the gravity vector in the world frame, and $g$ is its magnitude.

Similarly given the state $\mathcal{X}_i$ and the extrinsic parameters $\mathcal{E}$ and a set of accelerometer measurements between keyframe $i$ and $j$, the linear velocity in the world frame $\mathtt{W}$ at $j$ is:
\begin{equation}
    \begin{aligned}
    {_{\mathtt{W}_j}\mathbf{v}} = &\underbrace{{_{\mathtt{W}_i}\mathbf{v}} + \sum_{k=i}^{j-1} \big( {_\mathtt{W}\mathbf{g}} \Delta t + {{\mathbf{R}}_{\mathtt{W}\mathtt{I}_k}} {({_{\mathtt{I}_k}\tilde{\mathbf{a}}}-\mathbf{b}^a_i - \bm{\eta}^a) \Delta t } \big)}_\text{${_{\mathtt{W}_k}\mathbf{v}}$} \\
    \end{aligned}
    \label{equa:acc_v_integration}
\end{equation}
where ${_{\mathtt{W}_j}\mathbf{v}} \doteq \mathbf{R}_{\mathtt{W}\mathtt{I}_0} \mathbf{R}_{\mathtt{I}\mathtt{C}} \mathbf{R}_{j} \mathbf{R}_{\mathtt{D}\mathtt{C}}^T {{\mathbf{v}_j}}$ and ${_{\mathtt{W}_i}\mathbf{v}} \doteq \mathbf{R}_{\mathtt{W}\mathtt{I}_0} \mathbf{R}_{\mathtt{I}\mathtt{C}} \mathbf{R}_{i} \mathbf{R}_{\mathtt{D}\mathtt{C}}^T {{\mathbf{v}}_i}$ are the linear velocities in the frame $\mathtt{W}$ at keyframes $j$ and $i$ respectively, and $\mathbf{b}^a_i$ stands for the accelerometer bias at $i$.
% $\mathbf{R}_{\mathtt{C}\mathtt{I}}$ is derived from the extrinsic parameter according to \eqref{equa:state_extrinsic2}.
In addition, the position at keyframe $j$ is:
\begin{equation}
    \begin{aligned}
    {_\mathtt{W}}{{\mathbf{p}}}_{\mathtt{W}\mathtt{I}_j}   = {_\mathtt{W}}{{\mathbf{p}}}_{\mathtt{W}\mathtt{I}_i}  + \sum_{k=i}^{j-1}{}\big( &  {_{\mathtt{W}_k}\mathbf{v}} \Delta t + \frac{1}{2}{_\mathtt{W}\mathbf{g}}\Delta t^2 \\ & + \frac{1}{2}\mathbf{R}_{\mathtt{W}\mathtt{I}_k}({_{\mathtt{I}_k}\tilde{\mathbf{a}}}-\mathbf{b}^a_i - \bm{\eta}^a)\Delta t^2\big)\\
    \end{aligned}
    \label{equa:acc_p_integration}
\end{equation}
where ${_\mathtt{W}}{{\mathbf{p}}}_{\mathtt{W}\mathtt{I}_j} \doteq \mathbf{R}_{\mathtt{W}\mathtt{I}_0} {\mathbf{R}_{\mathtt{I}\mathtt{C}}} \mathbf{R}_{j} {_\mathtt{C}\mathbf{p}_{\mathtt{C}\mathtt{I}}} + \mathbf{R}_{\mathtt{W}\mathtt{I}_0} \mathbf{R}_{\mathtt{I}\mathtt{C}} \mathbf{p}_{j} $, and ${_\mathtt{W}}{{\mathbf{p}}}_{\mathtt{W}\mathtt{I}_i} \doteq \mathbf{R}_{\mathtt{W}\mathtt{I}_0} \mathbf{R}_{\mathtt{I}\mathtt{C}} \mathbf{R}_{i} {_\mathtt{C}\mathbf{p}_{\mathtt{C}\mathtt{I}}} + \mathbf{R}_{\mathtt{W}\mathtt{I}_0} \mathbf{R}_{\mathtt{I}\mathtt{C}} \mathbf{p}_{i} $ are the positions of the IMU in $\mathtt{W}$ at keyframes $j$ and $i$, respectively. % And $\mathbf{R}_{\mathtt{C}\mathtt{I}}$ and ${_\mathtt{C}\mathbf{p}_{\mathtt{C}\mathtt{I}}}$ are derived from the extrinsic parameter according to \eqref{equa:state_extrinsic2}.

As thoroughly investigated in \cite{forster2016manifold}, defining the IMU residual in accordance with \eqref{equa:gyros_integration}, \eqref{equa:acc_v_integration}, and \eqref{equa:acc_p_integration} would necessitate multiple times of integration during the optimization iterations. This can lead to considerable time expenditure. Therefore, a further reformulation of \eqref{equa:gyros_integration}, \eqref{equa:acc_v_integration}, and \eqref{equa:acc_p_integration} is applied by following the IMU pre-integration proposed in \cite{forster2016manifold}.
First, \eqref{equa:gyros_integration} can be reformulated as:
\begin{equation}
    \begin{aligned}
        \underbrace{ \mathbf{R}_{\mathtt{I}\mathtt{C}}  \mathbf{R}_{i}^T \mathbf{R}_{j} {\mathbf{R}_{\mathtt{I}\mathtt{C}}^T} }_\text{$h_{I_r}(\mathcal{X}_i,\mathcal{X}_j)$} = &
        \underbrace{ {\Delta{\hat{\mathbf{R}}}_{\mathtt{I}_i\mathtt{I}_j}} \text{Exp}( -\delta{\hat{\bm{\phi}}}_{\mathtt{I}_i\mathtt{I}_j}) }_\text{${\Delta{\mathbf{R}}_{\mathtt{I}_i\mathtt{I}_j}}$}
%        \\ = &\underbrace{\prod_{k=i}^{j-1} Exp\big(( _{\mathtt{I}_k}{\tilde{\bm{\omega}}} -\mathbf{b}^g_i) \Delta t\big) }_{\text{$\Delta{\hat{\mathbf{R}}}_{\mathtt{I}_i\mathtt{I}_j}$}} \\ & \underbrace{\prod_{k=i}^{j-1} Exp(-\Delta{\hat{\mathbf{R}}^T}_{\mathtt{I}_{k-1}\mathtt{I}_j} J_r(_{\mathtt{I}_k}{\tilde{\bm{\omega}}} -\mathbf{b}^g_i) \bm{\eta}^g \Delta t)}_{\text{$Exp( -\delta{\hat{\bm{\phi}}}_{\mathtt{I}_i\mathtt{I}_j})$}}
    \end{aligned}
    \label{equa:gyros_integration2}
\end{equation}
where $h_{I_r}(\mathcal{X}_i,\mathcal{X}_j)$ is the gyroscope measurement model and $\Delta{\mathbf{R}}_{\mathtt{I}_i\mathtt{I}_j}$  represents the incremental gyroscope relative motion. $\Delta{\hat{\mathbf{R}}}_{\mathtt{I}_i\mathtt{I}_j} \doteq \prod_{k=i}^{j-1} \text{Exp}\big(( _{\mathtt{I}_k}{\tilde{\bm{\omega}}}_{\mathtt{W}\mathtt{I}_k} -\mathbf{b}^g_i) \Delta t\big)$ stands for the gyroscope pre-integration from keyframe $i$ to $j$ and $\text{Exp}( -\delta{\hat{\bm{\phi}}}_{\mathtt{I}_i\mathtt{I}_j}) \doteq \prod_{k=i}^{j-1} \text{Exp}(-\Delta{\hat{\mathbf{R}}^T}_{\mathtt{I}_{k+1}\mathtt{I}_j} J_r(_{\mathtt{I}_k}{\tilde{\bm{\omega}}}_{\mathtt{W}\mathtt{I}_k} -\mathbf{b}^g_i) \bm{\eta}^g \Delta t)$ is the noise term and $ J_r(\bm{\phi}) \doteq \mathbf{I}-\frac{1-\\cos(\|\bm{\phi}\|)}{\|\bm{\phi}\|^2}\bm{\phi}^\land+\frac{\|\bm{\phi}\|-\\sin(\|\bm{\phi}\|)}{\|\bm{\phi}^3\|}(\bm{\phi}^\land)^2 $ is the right Jacobian of $\text{SO}(3)$ \cite{forster2016manifold}. % $\Delta {}^{b_i}{\hat{\mathbf{R}}}_{b_k} \doteq \prod_{k=i}^{j-1} \text{Exp}\big((_\mathtt{I}{\hat{\mathbf{\omega}}_k}-\mathbf{b}^g_i-\mathbf{\eta}^g) \Delta t\big)$ is the .
Moreover, \eqref{equa:acc_v_integration} can be reformulated as:
\begin{gather}
    \begin{aligned}
     & \underbrace{
     \mathbf{R}_{\mathtt{I}\mathtt{C}} {{\mathbf{R}_{i}^T}} \big(\mathbf{R}_{j} {\mathbf{R}_{\mathtt{D}\mathtt{C}}^T} \mathbf{v}_j - \mathbf{R}_{i} {\mathbf{R}_{\mathtt{D}\mathtt{C}}^T} \mathbf{v}_i - (\mathbf{R}_{\mathtt{W}\mathtt{I}_0} \mathbf{R}_{\mathtt{I}\mathtt{C}})^T {_\mathtt{W}\mathbf{g}} \Delta t_{ij} \big) }_\text{$h_{I_{v}}(\mathcal{X}_i,\mathcal{X}_j)$} \\
     & = \underbrace{\Delta {\hat{\mathbf{v}}}_{\mathtt{I}_i\mathtt{I}_j} - \delta {\hat{\mathbf{v}}}_{\mathtt{I}_i\mathtt{I}_j}}_\text{$\Delta {\mathbf{v}}_{\mathtt{I}_i\mathtt{I}_j}$}
%     \\ & = \underbrace{\sum_{k=i}^{j-1}{\Delta {\mathbf{R}}_{\mathtt{I}_i\mathtt{I}_k}({_{\mathtt{I}_k}\tilde{\mathbf{a}}}-\mathbf{b}^a_i ) \Delta t }}_\text{$\Delta {\hat{\mathbf{v}}}_{\mathtt{I}_i\mathtt{I}_j}$} - \underbrace{\sum_{k=i}^{j-1}{\Delta {\mathbf{R}}_{\mathtt{I}_i\mathtt{I}_k}\bm{\eta}^a \Delta t }}_\text{$\delta {\hat{\mathbf{v}}}_{\mathtt{I}_i\mathtt{I}_j}$}
    \end{aligned}
    \label{equa:acc_v_integration2}
    % \raisetag{25pt}
\end{gather}
where $h_{I_{v}}(\mathcal{X}_i,\mathcal{X}_j)$ is the accelerometer velocity measurement model and $\Delta {\mathbf{v}}_{\mathtt{I}_i\mathtt{I}_j}$ is the relative linear velocity incremental. $\Delta {\hat{\mathbf{v}}}_{\mathtt{I}_i\mathtt{I}_j} \doteq \sum_{k=i}^{j-1}{\Delta {\hat{\mathbf{R}}}_{\mathtt{I}_i\mathtt{I}_k}({_{\mathtt{I}_k}\tilde{\mathbf{a}}}-\mathbf{b}^a_i)\Delta t }$ is the velocity pre-integration of the accelerometer measurements from $i$ to $j$ and $\delta {\hat{\mathbf{v}}}_{\mathtt{I}_i\mathtt{I}_j} \doteq \sum_{k=i}^{j-1}\left[-\Delta {\hat{\mathbf{R}}}_{\mathtt{I}_i\mathtt{I}_k}({_{\mathtt{I}_k}\tilde{\mathbf{a}}}-\mathbf{b}^a_i )^{\land} \delta{\hat{\bm{\phi}}}_{\mathtt{I}_i\mathtt{I}_k} \Delta t + \Delta {\hat{\mathbf{R}}}_{\mathtt{I}_i\mathtt{I}_k} \bm{\eta}^{a}\Delta t\right]$ is the Gaussian noise \cite{forster2016manifold}. % $\Delta {\hat{\mathbf{R}}}_{\mathtt{I}_i\mathtt{I}_k}$ stands for the gyroscope pre-integration from $i$ to $k$ refering to the definition of $\Delta {\hat{\mathbf{R}}}_{\mathtt{I}_i\mathtt{I}_j}$ at \eqref{equa:gyros_integration2}.
In addition, \eqref{equa:acc_p_integration} can be reformulated as:
\begin{equation}
%    \footnotesize
    \begin{aligned}
    &\mathbf{R}_{\mathtt{I}\mathtt{C}} \mathbf{R}_{i}^T \big(\mathbf{R}_{j} {_\mathtt{C}\mathbf{p}_{\mathtt{C}\mathtt{I}}} + {\mathbf{p}_j} - \mathbf{R}_{i} {_\mathtt{C}\mathbf{p}_{\mathtt{C}\mathtt{I}}} -  {\mathbf{p}_i} \\& \underbrace{ - \mathbf{R}_{i} {\mathbf{R}_{\mathtt{I}\mathtt{C}}^T} {\mathbf{v}}_i \Delta t_{ij}  - \frac{1}{2} (\mathbf{R}_{\mathtt{W}\mathtt{I}_0} \mathbf{R}_{\mathtt{I}\mathtt{C}})^T {_\mathtt{W}\mathbf{g}} \Delta t_{ij}^2 \big)}_\text{$h_{I_{t}}(\mathcal{X}_i,\mathcal{X}_j)$} \\
     &= \underbrace{\Delta {_{\mathtt{I}_i}{\hat{\mathbf{p}}}_{\mathtt{I}_i{\mathtt{I}_j}}} - \delta {_{\mathtt{I}_i}{\hat{\mathbf{p}}}_{\mathtt{I}_i{\mathtt{I}_j}}}}_\text{$\Delta {_{\mathtt{I}_i}{\mathbf{p}}_{\mathtt{I}_i{\mathtt{I}_j}}}$}
%    \underbrace{\sum_{k=i}^{j-1}{}[ \Delta {\hat{\mathbf{v}}}_{\mathtt{I}_i\mathtt{I}_k} \Delta t + \frac{1}{2} \Delta{\hat{\mathbf{R}}}_{\mathtt{I}_i\mathtt{I}_k} ({_{\mathtt{I}_k}\tilde{\mathbf{a}}}-\mathbf{b}^a_i - \bm{\eta}^a)\Delta t^2]}_\text{$\Delta {_{\mathtt{I}_i}{\hat{\mathbf{p}}}_{\mathtt{I}_i{\mathtt{I}_j}}}$}\\ &=
    \end{aligned}
    \label{equa:acc_t_integration2}
\end{equation}
where $h_{I_{t}}(\mathcal{X}_i,\mathcal{X}_j) $ is the accelerometer translation model, $\Delta {_{\mathtt{I}_i}{\hat{\mathbf{p}}}_{\mathtt{I}_i{\mathtt{I}_j}}} \doteq \sum_{k=i}^{j-1}{}[ \Delta {\hat{\mathbf{v}}}_{\mathtt{I}_i\mathtt{I}_k} \Delta t + \frac{1}{2} \Delta{\hat{\mathbf{R}}}_{\mathtt{I}_i\mathtt{I}_k} ({_{\mathtt{I}_k}\tilde{\mathbf{a}}}-\mathbf{b}^a_i)\Delta t^2]$ is the translation pre-integration of accelerometer measurement, and $\delta {_{\mathtt{I}_i}{\hat{\mathbf{p}}}_{\mathtt{I}_i{\mathtt{I}_j}}} \doteq \sum_{k=i}^{j-1}\Big[\delta {\hat{\mathbf{v}}}_{\mathtt{I}_i\mathtt{I}_k} \Delta t - \frac12 {\Delta {\hat{\mathbf{R}}}_{\mathtt{I}_i\mathtt{I}_k}} ({_{\mathtt{I}_k}\tilde{\mathbf{a}}}-\mathbf{b}^a_i)^\land \delta{\hat{\bm{\phi}}}_{\mathtt{I}_i\mathtt{I}_k} \Delta t^2 +  \frac12 {\Delta {\hat{\mathbf{R}}}_{\mathtt{I}_i\mathtt{I}_k}} \bm{\eta}^{a}\Delta t^2]$ is the noise. Refer to \cite{forster2016manifold} for the detailed derivation.
%\Sen{is there any specific difference to the IMU pre-integration in [13]. If no, we may be able to cut some of this part}
%    \Shida{Compared to the approach in reference [13], the underlying principles remain the same. However, we've reformulated all the equations based on our state definition. Simply referring to [13] without delving into these specifics might make it challenging to understand, even for those familiar with VIO.}

\subsubsection{Residual Definition}
Then we can define the formula of the IMU residual in  \eqref{equa:opt_implicit} as follows:
\begin{equation}
    % & \prescript{b_i}{}{\mathbf{R}}_{b_j} \doteq  \prescript{g}{}{\mathbf{R}}_{d} \mathbf{R}_{\mathtt{D}\mathtt{C}} \mathbf{R}_{i}^T \mathbf{R}_{j} [\prescript{g}{}{\mathbf{R}}_{d} \mathbf{R}_{\mathtt{D}\mathtt{C}}]^T \\
    \mathbf{r}_{I}(h_{I}(\mathcal{X}_i,\mathcal{X}_j),\mathcal{I}_{i,j}) \doteq
    \left[ \begin{matrix}
        \mathbf{r}_{r}(h_{I_r}(\mathcal{X}_i,\mathcal{X}_j),\mathcal{I}_{i,j})\\
        \mathbf{r}_{v}(h_{I_{v}}(\mathcal{X}_i,\mathcal{X}_j),\mathcal{I}_{i,j}) \\
        \mathbf{r}_{t}(h_{I_{t}}(\mathcal{X}_i,\mathcal{X}_j),\mathcal{I}_{i,j})
    \end{matrix} \right]
    \label{equa:gyros_integration3}
\end{equation}
According to  \eqref{equa:gyros_integration2}, the IMU rotation residual is defined as
\begin{equation}
    \begin{aligned}
    &\mathbf{r}_{r}(h_{I_r}(\mathcal{X}_i,\mathcal{X}_j),\mathcal{I}_{i,j}) \doteq \text{Log} \big(h_{I_r}(\mathcal{X}_i,\mathcal{X}_j) \Delta {\hat{\mathbf{R}}}_{\mathtt{I}_i\mathtt{I}_k}^T\big)
    \end{aligned}
    \label{equa:imu_rotation_residual}
\end{equation}
where $\text{Log}(\cdot): \text{SO}(3) \rightarrow \mathbb{R}^3$ stands for the logarithm map from $\text{SO}(3)$ to a vectorized $\mathfrak{so}(3)$.

The IMU velocity residual, based on \eqref{equa:acc_v_integration2}, is
\begin{equation}
    \begin{aligned}
    &\mathbf{r}_{v}(h_{I_{v}}(\mathcal{X}_i,\mathcal{X}_j),\mathcal{I}_{i,j}) \doteq \Delta {\hat{\mathbf{v}}}_{\mathtt{I}_i\mathtt{I}_j} - h_{I_{v}}(\mathcal{X}_i,\mathcal{X}_j)
    \end{aligned}
    \label{equa:imu_velocity_residual}
\end{equation}
Similarly, from  \eqref{equa:acc_t_integration2} the IMU translation residual can be derived as
\begin{equation}
    \begin{aligned}
    &\mathbf{r}_{t}(h_{I_t}(\mathcal{X}_i,\mathcal{X}_j),\mathcal{I}_{i,j}) \doteq \Delta {_{\mathtt{I}_i}{\hat{\mathbf{p}}}_{\mathtt{I}_i{\mathtt{I}_j}}} - h_{I_{t}}(\mathcal{X}_i,\mathcal{X}_j)
    \end{aligned}
    \label{equa:imu_translation_residual}
\end{equation}

\subsection{Camera Measurement Model and Its Residuals}\label{section:camera_measurement_model}
The camera measurement is a set of stereo image pairs that are used to extract image features and estimate 3D landmarks. % More specifically each measurement is a pixel on the image.
Given the landmarks $\mathcal{L}_i$ visible in keyframe $i$ and
% In practice, the initial map points are obtained via stereo matching of a stereo image pair. % The data association can be obtained by feature extraction and matching, or KLT tracker \textbf{cite}.
the state $\mathcal{X}_i$, %, and a corresponding landmark ${_{\mathtt{C}_0}\mathbf{l}_{\mathtt{C}_0\mathtt{l}_m}}$ and its pixel position ${\mathbf{u}_m}$ on the image
the camera projection model of the landmark ${_{\mathtt{C}_0}\mathbf{l}_{\mathtt{C}_0\mathtt{l}_m}}$ is
\begin{equation}
    \Pi(\mathbf{T}_{i}^{-1} \otimes {_{\mathtt{C}_0}\mathbf{l}_{\mathtt{C}_0\mathtt{l}_m}}) \doteq {\mathbf{u}_m}
    \label{equa:camera_motion_i}
\end{equation}
% where $\mathcal{L}_i$. % and $^{c_0}\mathbf{T}_{c_i}^{-1}\otimes {^{c_0}\mathbf{p}_{k}}$ represents transform the map point from world($c_0$) frame to camera frame.
where $\Pi(\cdot)$ is the camera model that projects a 3D map point from the local camera coordinate frame to the pixel coordinate frame,
$\otimes$ is the transformation operation of SE(3) group over $\mathbb{R}^3$ elements,
and ${\mathbf{u}_m}$ is the observed pixel position of the landmark on the image. % In practice, we will have hundreds measurement as  \eqref{equa:camera_motion_i}) to constrain the camera pose.
Therefore, the camera measurement model in  \eqref{equa:opt_implicit} is
\begin{equation}
    h_C(\mathcal{X}_{i}) = \Pi(\mathbf{T}_{i}^{-1}\otimes {_{\mathtt{C}_0}\mathbf{l}_{\mathtt{C}_0\mathtt{l}_m}})
\end{equation}
We assume the camera measurement is affected by a Gaussian noise. Therefore, considering all the landmarks of keyframe $i$, the camera residual is formulated as
\begin{equation}
    \mathbf{r}_{C}(h_{C}(\mathcal{X}_i),\mathcal{C}_i)) = \sum^{}_{m \in \mathcal{M}_i} \Pi(\mathbf{T}_{i}^{-1}\otimes {_{\mathtt{C}_0}\mathbf{l}_{\mathtt{C}_0\mathtt{l}_m}}) - \mathbf{u}_m
    \label{equa:camera_residual}
\end{equation}
Refer to \cite{mur2017orb2} for more details on the camera measurement model.

\section{Online Sensor Calibration}
\label{sec:Extrinsic Calibration}
%The performance of the SLAM system may be influenced by two potential sources of errors related to sensor extrinsic calibration. The first pertains to the extrinsic parameters, encompassing the transformations among the camera, the DVL and the IMU. These parameters may experience slight variations following extended use and pose challenges in terms of accurate direct measurement. The second source of error is associated with the misalignment of the DVL transducers. These transducers could be misaligned during the manufacturing process, and further changed over time. In this section, we propose a methodology for calibrating both the extrinsic parameters among sensors, and focus on  the DVL transducer misalignment calibration in next section.
    The performance of the acoustic-visual-inertial SLAM system proposed in the previous section can be affected by two primary sources of errors related to sensor calibration. The first source concerns the extrinsic sensor calibration, i.e., the transformations between the camera, the DVL and the IMU. It is often challenging to manually measure these extrinsic parameters $\mathcal{E}$, particularly rotations, with a high accuracy. Since these parameters might also undergo minor variations over extended usage, the extrinsic calibration process needs to be carried out regularly. The second source of error arises from the misalignment of the DVL transducers. This misalignment can originate during the manufacturing process or be developed as the DVL sensor ages. Therefore, we propose online sensor calibration methods to alleviate or remove these two types of errors.
    % the first subsection section, we introduce a method for calibrating the extrinsic parameters among the sensors. The calibration of the DVL transducer misalignment will be addressed in the subsequent subsection.

\subsection{Extrinsic Calibration of DVL, Camera and IMU Sensors}
    \label{sec:extrinsic_calibration_algorithm}

%In the preceding section, we assumed the extrinsic parameters, denoted as $\mathcal{E}$, to be known. However, within the realm of extrinsic calibration, these parameters are now treated as variables requiring estimation. The problem formulation remains as described by Equation (\ref{equa:opt_implicit}). However, it's essential to note that the convergence of this system is highly sensitive to the initial values of the unknown extrinsic parameters. A suboptimal initial value can easily lead the system to converge to a local minimum. Therefore, we propose a meticulously designed crease-to-fine calibration procedure to robustly estimate these parameters. The mitivation of the calibration procedure is first initially calibrate a rough parameter with a simple model and as least as possible variables as a good initial value. And gradually increase the complexity of the model and the number of variables, and start the calibration from the intialial value from previous steps and gradually refine the parameters.
In the previous section, the extrinsic parameters $\mathcal{E}$ are presumed known. Now they are considered unknown variables to be estimated. Therefore, we cannot directly leverage the SLAM methodology proposed in Section \ref{section:Camera DVL IMU Fusion}. % because the extrinsic parameters $\mathcal{E}$ are unknown.
A straightforward solution is to jointly estimate the SLAM states and the extrinsic parameters within the MAP framework in \eqref{equa:opt_implicit}. However, this is impracticable since the optimization would be acutely sensitive to initial values and suffer from considerable local minima.
% It is pivotal to recognize that the system's convergence is acutely sensitive to the initial conditions of the unknown extrinsic parameters. If initialized suboptimally, the system may converge to a local minimum.
To address these issues, we propose a systematic coarse-to-fine calibration methodology that has the following steps to progressively refine the estimates of the extrinsic parameters $\mathcal{E}$: % to reliably estimate these parameters. The underlying motivation for this calibration procedure is to initially approximate a rudimentary parameter using a simplified model with minimal variables, serving as a favorable initial condition. Subsequently, the model's intricacy and the variable count are progressively augmented, utilizing the prior step's initial value as a starting point, thereby refining the parameters iteratively.
%Given a set of camera images, DVL measurements and IMU measurements, our goal to find the optimal extrinsic parameters $\mathcal{E}$ by following the steps below:
% The proposed calibration methodology is delineated through the subsequent phases:
\begin{enumerate}
    \item Vision-only Bundle Adjustment.
    \item Initial estimation of extrinsic parameters.
    \item Refined estimation of extrinsic parameters with gyroscope bias.
    \item Initialization of gravity direction.
    \item Full refinement.
\end{enumerate}

\subsubsection{Vision-only Bundle Adjustment}
    \label{sec:Vision_Only_Bundle_Adjustment}
    %The first step is to estimate the camera pose and 3D position of all landmarks from the given camera images. This problem can be treated as a pure visual SLAM problem which has been well studied at \cite{mur2017orb2}. We formulate this probelm as a Bundle Adjustment probelm as follow:
    The initial step is to estimate the camera poses and the 3D landmark positions purely based on a short sequence of  images (empirically 10-100 keyframes) for sensor calibration. The standard vision based SLAM and local Bundle Adjustment method \cite{mur2017orb2,triggs2000bundle} is used:
%We assume, during the calibration process, good landmarks can be extracted. Then they are used to build a Bundle-Adjustment problem for estimating camera poses. % And we also assume the DVL and gyroscope measurement are accurate during the calibration.
\begin{equation}
    \{\mathbf{T}_{i}^*,\mathcal{L}_i^*\} = \mathop{\mathrm{argmin}}_{\mathbf{T}_{i},\mathcal{L}_i} \ \sum^{}_{m \in \mathcal{M}_i} \| \Pi(\mathbf{T}_{i}^{-1}\otimes {_{\mathtt{C}_0}\mathbf{l}_{\mathtt{C}_0\mathtt{l}_m}}) - \mathbf{u}_m \|^2_{\Sigma_C}
    \label{equa:calibration_vision_only}
\end{equation}
Since a stereo camera is used in our system, the absolute metric scale can be recovered.
% First we need to estimate the camera pose and the landmark position from a set of images, which is well studied in visual SLAM \textbf{cite}. We can build a Bundle-Adjustment problem as \eqref{equa:opt_visual}. % We assume the image contains enough clear feature and the data association is given properly and we also assume
After solving the optimization, a set of refined camera poses and landmarks $\{\mathbf{T}_{i}^*,\mathcal{L}_i^*\}$ are obtained. They are treated as known in the following calibration procedure to facilitate convergence and reliability.
Notably, to ensure the success of this step, the short image sequence utilized needs to have reasonable image quality with sufficient visual textures for feature extraction and matching, and preferably the observability of the parameters is ensured from the robot motion \cite{xu2023observability}.
% Consequently, the presence of distinct and abundant features on the image is imperative for optimal calibration.
\subsubsection{Initial Estimation of Extrinsic Parameters}
    %The goal of the second step is to estimate a roughly accurate extrinsic parameters without the impact of IMU bias and gravity direction estimation. In the second step, we initially estimate the extrinsic parameters using the camera pose and landmark we obtained at the first step and formulate the problem as:
    In the second phase, an initial estimation of $\mathcal{E}$ is computed by mitigating the influence of the IMU biases and the gravity. By leveraging the optimal camera poses $\{\mathbf{T}_{i}^*\}$ obtained in the last step, the optimization problem that employs the DVL translation residual in \eqref{equa:dvl_residual} and the IMU rotation residual in \eqref{equa:imu_rotation_residual} is formulated as
\begin{equation}
    \begin{aligned}
        \mathop{\mathrm{argmin}}_{\mathcal{E}} \ \sum_{i,j \in \mathcal{K}_n} & \| \mathbf{r}_{t}(h_{D_t}(\mathcal{X}_i,\mathcal{X}_j),\mathcal{D}_{i,j}) \|_{\Sigma_{D}}^2 \\& + \sum_{i,j \in \mathcal{K}_n} \| \mathbf{r}_{r}(h_{I_r}(\mathcal{X}_i,\mathcal{X}_j),\mathcal{I}_{i,j}) \|_{\Sigma_{I}}^2
    \end{aligned}
    \label{equa:calibration_second}
\end{equation}
    %where the DVl translation residual defined at Equation (\ref{equa:dvl_residual}) and IMU rotation residual defined at Equation (\ref{equa:imu_rotation_residual}) are used. Notably, we assume zero gyroscope bias at current step.
    % In this formulation,  are adopted. It's pivotal to highlight that,
    For this step, the gyroscope bias is assumed zero.

\subsubsection{Refined Estimation of Extrinsic Parameters with Gyroscope Bias}
%After the previous step, the extrinsic parameters are already in a optimal initial value and we include the gyroscope bias to optimization in this step to refine the extrinsic parameters. Similar with the last step we still use the DVL translation residual and IMU rotation residual and the problem is formulated as:
% Following the prior procedure, $\mathcal{E}$ have been estimated to an optimal initial value. At this juncture,
This step incorporates the gyroscope bias into the optimization process to refine the $\mathcal{E}$ estimate. The same residual terms to the preceding stage are employed for the optimization problem but with the gyroscope biases included as variables to optimize:
    \begin{equation}
        \begin{aligned}
            \mathcal{E}^*,{\mathcal{B}^g}^* = \mathop{\mathrm{argmin}}_{\mathcal{E},{\mathcal{B}}^g} \ \sum_{i,j \in \mathcal{K}_n} &\| \mathbf{r}_{t}(h_{D_t}(\mathcal{X}_i,\mathcal{X}_j),\mathcal{D}_{i,j}) \|_{\Sigma_{D}}^2 \\& + \sum_{i,j \in \mathcal{K}_n} \| \mathbf{r}_{r}(h_{I_r}(\mathcal{X}_i,\mathcal{X}_j),\mathcal{I}_{i,j}) \|_{\Sigma_{I}}^2
        \end{aligned}
        \label{equa:calibration_third}
    \end{equation}
where ${\mathcal{B}}^g \doteq \{\mathbf{b}^a_{i}\}, i \in \mathcal{K}_n$ denotes the collection of gyroscope biases associated with the keyframes in the image sequence.

    \subsubsection{Initialization of Gravity Direction}
    %In the following procedure, we aim to incorporate IMU velocity and translation residual to further refine $\mathcal{E}$. To do so we need to initialize $\mathbf{R}_{\mathtt{W}\mathtt{I}_0}$ first. In this step, we fix bias and extrinsic parameters and only optimize  $\mathbf{R}_{\mathtt{W}\mathtt{I}_0}$. The problem can be formulated as:
    To further enhance the accuracy of $\mathcal{E}$, the IMU velocity and translation residuals in \eqref{equa:imu_velocity_residual} and \eqref{equa:imu_translation_residual} should be considered. For this purpose, the initialization of $\mathbf{R}_{\mathtt{W}\mathtt{I}_0}$ is essential. Therefore, based on the optimal $\{\mathbf{T}_{i}^*, \mathcal{E}^*, {\mathcal{B}^g}^*\}$, $\mathbf{R}_{\mathtt{W}\mathtt{I}_0}$ is optimized by
    \begin{equation}
        \begin{aligned}
            \mathop{\mathrm{argmin}}_{ {\mathbf{R}_{\mathtt{W}\mathtt{I}_0}} } \ \sum_{i,j \in \mathcal{K}_n} & \| \mathbf{r}_{r}(h_{I_r}(\mathcal{X}_i,\mathcal{X}_j),\mathcal{I}_{i,j}) \|_{\Sigma_{I}}^2 \\& + \sum_{i,j \in \mathcal{K}_n} \| \mathbf{r}_{v}(h_{I_{v}}(\mathcal{X}_i,\mathcal{X}_j),\mathcal{I}_{i,j}) \|_{\Sigma_{I}}^2
        \end{aligned}
        \label{equa:calibration_forth}
    \end{equation}

    \subsubsection{Full Refinement}
    %After the previous steps, we can already obtain a good initial value for all the bias, gravity direction, and extrinsic parameters. Therefore we are ready to use the full Visual Acoustic Inertial model to refine the extrinsic parameters to a more accurate result.
    % encompassing IMU bias, gravity direction, and extrinsic parameters
    % Following the aforementioned procedures, we have successfully determined preliminary values for the bias, $\mathbf{R}_{\mathtt{W}\mathtt{I}_0}$, and $\mathcal{E}$. With this foundation established,
    Now we are ready to perform full refinement of the extrinsic parameters by using the DVL translation residual and all the three IMU residual terms:
    \begin{equation}
        \small
        \begin{aligned}
            \mathcal{E}^*  = \mathop{\mathrm{argmin}}_{\mathcal{E},\mathbf{R}_{\mathtt{W}\mathtt{I}_0},{\mathcal{B}}^g,{\mathcal{B}}^a} \ \sum_{i,j \in \mathcal{K}_n} & \| \mathbf{r}_{t}(h_{D_t}(\mathcal{X}_i,\mathcal{X}_j),\mathcal{D}_{i,j}) \|_{\Sigma_{D}}^2 \\& + \sum_{i,j \in \mathcal{K}_n} \| \mathbf{r}_{I}(h_{I}(\mathcal{X}_i,\mathcal{X}_j),\mathcal{I}_{i,j})\|_{\Sigma_{I}}^2
        \end{aligned}
        \label{equa:calibration_fifth}
    \end{equation}
    whose initialization takes the previous optimal values. ${\mathcal{B}}^a \doteq \{\mathbf{b}^a_{i}\}, i \in \mathcal{K}_n $ is the set of accelerometer biases.

Since the DVL translation residual % as presented in \eqref{equa:dvl_decoupled}, intrinsically
incorporates the extrinsic parameter $\mathbf{R}_{\mathtt{I}\mathtt{D}}$, any alteration in $\mathbf{R}_{\mathtt{I}\mathtt{D}}$ necessitates the re-integration of $\Delta {_{\mathtt{D}_i}}{\bar{\mathbf{p}}}_{\mathtt{D}_i\mathtt{D}_j}$ during the optimization process. This recurrent computation is computationally intensive and hinders online optimization. A solution to this problem is elaborated upon in Section \ref{sec:rapid_iteration}.
    %In theory, the extrinsic parameters can be optimized with the state  online. However, according to our practical results,  the calibration is performed before the full state estimation and in scenarios where good visual features are detachable since the calibration results are heavily determined by the quality of pose estimation which is not always reliable in underwater. In addition, the calibration is coupled with robot motions during the calibration process. %We will discuss this in Section \ref{sec:Observability Analysis}. % So it cannot give a good result if we always keep optimizing the extrinsic parameters with state.

% \subsection{Observability Analysis}\label{sec:Observability Analysis}
% The calibration need the vehicle have enough motion in different axis to get a good calibration result. In this section, we will discuss what motion does it require.

%\section{DVL Misalignment Calibration}\label{sec:DVL Calibration}
%The misalignment of the DVL transducers also affects the system performance. Such misalignment can occur during manufacturing and may evolve over time. In this section, we introduce a method to calibrate the misalignment of each DVL transducer based on the DVL we introduced in  \ref{section:dvl_measurement_model}.

%\subsection{Misalignment of DVL Transducers}
\subsection{DVL Misalignment Calibration}
    \label{sec:DVL_calibration}
In Section \ref{section:dvl_velocity_model}, it is assumed that each transducer is tilted by fixed angles $\alpha$ and $\beta$. For the DVL misalignment calibration, the $\alpha$ and $\beta$ of each transducer are treated as unknown instead. Therefore, \eqref{equa:e_vector1} can be reformulated as
\begin{gather}
\small
    \begin{aligned}
     \mathbf{\hat{e}}_1
     &=[-\cos(\beta_1) \cos(\alpha_1) \  &\sin(\beta_1) \cos(\alpha_1) \  &\sin(\alpha_1)] \\
      \mathbf{\hat{e}}_2
     &=[-\cos(\beta_2) \cos(\alpha_2) \  &-\sin(\beta_2) \cos(\alpha_2) \  &\sin(\alpha_2)] \\
      \mathbf{\hat{e}}_3
     &=[\cos(\beta_3) \cos(\alpha_3) \  &-\sin(\beta_3) \cos(\alpha_3) \  &\sin(\alpha_3)] \\
      \mathbf{\hat{e}}_4
     &=[\cos(\beta_4) \cos(\alpha_4) \  &\sin(\beta_4) \cos(\alpha_4) \  &\sin(\alpha_4)]
     \label{equa:e_vector_mis}
     \end{aligned}
\end{gather}
where $\mathcal{O} \doteq \{ \alpha_n, \beta_n\}, n \in \{1, 2, 3, 4\}$ is the DVL transducer alignment parameters to be estimated. The extrinsic parameters $\mathcal{E}$ are known during the DVL misalignment calibration.

    %The idea is we first estimate a set a camera pose and landmark locations at given keyframes based on Vision-Only Bundle Adjustment. Then we treat the camera pose and landmark locations as knows and initialize velocity, IMU bias, and gravity direction. Finally, we treat the velocities as known and estimate the orientation of transducer upon them.
    Similar to the extrinsic calibration introduced in Section \ref{sec:extrinsic_calibration_algorithm}, a coarse-to-fine strategy is adopted for the robust calibration of $\mathcal{O}$.
    % In a manner analogous to the extrinsic calibration detailed in Section \ref{sec:extrinsic_calibration_algorithm}, we employ a hierarchical coarse-to-fine strategy for the robust calibration of $\mathcal{O}$. And we assume extrinsic parameters are known during DVL misalignment calibration. Initially, we derive an estimation of both the camera pose and landmark positions at specific keyframes, utilizing Vision-Only Bundle Adjustment which is described at Section \ref{sec:Vision_Only_Bundle_Adjustment}. Upon obtaining these estimates, we then consider the camera pose and landmark positions as known parameters. Subsequently, we initialize the parameters of velocity, IMU bias, and the direction of gravity. In the final step, with the velocities treated as known values, we estimate the orientation of the transducer based on these velocities.
    It includes the following steps:
    \begin{enumerate}
        \item Vision-only Bundle Adjustment.
        \item Optimization of DVL body velocity.
        \item Optimization of DVL transducer alignment parameters.
    \end{enumerate}

    \subsubsection{Vision-only Bundle Adjustment}
    This initial phase is identical to the vision-only Bundle Adjustment in Section \ref{sec:Vision_Only_Bundle_Adjustment}. The optimal camera poses and landmark locations are also fixed in the following procedure.
    \subsubsection{Optimization of DVL Body Velocity}
    We then optimize the velocities based on the DVL translation residual in \eqref{equa:dvl_residual} with $h_{D_t}(\mathcal{X}_i,\mathcal{X}_j)$ being considered as a constant measurement. Therefore, the problem is reformulated as
\begin{equation}
    \small
    \mathcal{V}^* = \mathop{\mathrm{argmin}}_{ \mathcal{V} } \| h_{D_t}(\mathcal{X}_i,\mathcal{X}_j) - {\sum_{k=i}^{j-1}} {\Delta\hat{\mathbf{R}}_{\mathtt{I}_i\mathtt{I}_k}} \mathbf{R}_{\mathtt{I}\mathtt{D}}\ {_{\mathtt{D}_i}{\mathbf{v}}} \Delta t \|^2_{\Sigma_D}
    \label{eq:DVLBodyVOpt}
\end{equation}
    where $\mathcal{V} \doteq \{ _{\mathtt{D}_i}\mathbf{v} \}, i \in \mathcal{K}_n$ is the set of the DVL body velocity variables to optimize.

    \subsubsection{Optimization of DVL Transducer Alignment Parameters}
    Given the calculated DVL body velocities $\mathcal{V}^*$, the DVL transducer alignment parameters $\mathcal{O}$ is estimated by
    \begin{equation}
        \small
            \mathcal{O}^* = \mathop{\mathrm{argmin}}_{\mathcal{O}} \sum_{i \in \mathcal{K}_n}  \sum_{n=1}^{4} \ \| _{i}\tilde{v}_n - \mathbf{\hat{e}}_n \cdot {_{\mathtt{D}_i}\mathbf{v}^*} \|^2_{\Sigma_d}
    \end{equation}
    where $n$ denotes the transducer index, $ _{i}\tilde{v}_n $ refers to the velocity measurement of the $n$th individual transducer at keyframe $i$, and $\mathbf{\hat{e}}_n$ is defined in \eqref{equa:e_vector_mis}.

    % Same to the extrinsic calibration, the term $\Delta {_{\mathtt{D}_i}}{\bar{\mathbf{p}}}_{\mathtt{D}_i\mathtt{D}_j}$, as delineated in \eqref{equa:dvl_decoupled}, is intrinsically associated with the DVL body velocity, denoted as $_{\mathtt{D}_i}\mathbf{v}$. Throughout the optimization procedure, any modifications to $_{\mathtt{D}_i}\mathbf{v}$ demand the recalibration of $\Delta {_{\mathtt{D}_i}}{\bar{\mathbf{p}}}_{\mathtt{D}_i\mathtt{D}_j}$. The repeated calculations involved in this process are computationally demanding, making real-time optimization challenging. A detailed resolution to this issue will be discussed in Section \ref{sec:rapid_iteration}.

    \subsection{Rapid Linear Approximation Iteration}
    \label{sec:rapid_iteration}
    %In order to avoid the repeated integration of $\Delta {_{\mathtt{D}_i}}{\bar{\mathbf{p}}}_{\mathtt{D}_i\mathtt{D}_j}$ during both extrinsic calibration and transducer calibration, we proposed the rapid linear approximation iteration method. The idea is to linearize the DVL translation pre-integration with respect to the variable we are going to estimate, then use its linear approximation instead of the repeated integration during iteration of the optimization when the incremental update is lower than a threshold compared with the linearization point.
To enhance the computational efficiency of minimizing the DVL translation residual, a linear approximation method is introduced in the optimization iteration to rapidly approximate the full integration of $\Delta {_{\mathtt{D}_i}}{\bar{\mathbf{p}}}_{\mathtt{D}_i\mathtt{D}_j}$ with linearization. When incremental updates are relatively minor, this approximation is used instead.

\subsubsection{Approximation Iteration for Extrinsic Calibration}
For the extrinsic calibration \eqref{equa:calibration_fifth}, the DVL translation pre-integration $\Delta {_{\mathtt{D}_i}}{\bar{\mathbf{p}}}_{\mathtt{D}_i\mathtt{D}_j}$ is linearized with respect to $ \mathbf{R}_{\mathtt{I}\mathtt{D}}$ as
\begin{equation}
    \frac{\partial {\Delta {_{\mathtt{D}_i}}{\bar{\mathbf{p}}}_{\mathtt{D}_i\mathtt{D}_j}} }{\partial {\bm{\phi}_{\mathtt{I}\mathtt{D}}}} = \sum_{k=i}^{j-1} \ - {\Delta\hat{\mathbf{R}}_{\mathtt{I}_i\mathtt{I}_k}} ( {{\mathbf{R}}_{\mathtt{I}\mathtt{D}}} \ {_{\mathtt{D}_i}\mathbf{v}} )^\land \Delta t
    \label{equa:extrinsic_linearization}
\end{equation}
where $ {\bm{\phi}_{\mathtt{I}\mathtt{D}}} \in \mathbb{R}^3 $ represents the vectorized Lie algebra $\mathfrak{so}(3)$ of ${{\mathbf{R}}_{\mathtt{I}\mathtt{D}}}$. See its full derivation in Appendix \ref{sec:Jacobian of Extrinsic Calibration Approximation Iteration Derivation}. Therefore, the approximation of the DVL translation pre-integration with an incremental update $\Delta {\bm{\phi}_{\mathtt{I}\mathtt{D}}}$ can be computed by
\begin{equation}
{{\bar{\mathbf{P}}}_{\bm{\phi}}}(\Delta {\bm{\phi}_{\mathtt{I}\mathtt{D}}}) \doteq \Delta {_{\mathtt{D}_i}}{\bar{\mathbf{p}}}_{\mathtt{D}_i\mathtt{D}_j} + \frac{\partial {\Delta {_{\mathtt{D}_i}}{\bar{\mathbf{p}}}_{\mathtt{D}_i\mathtt{D}_j}} }{\partial {\bm{\phi}_{\mathtt{I}\mathtt{D}}}} \Delta {\bm{\phi}_{\mathtt{I}\mathtt{D}}}
\end{equation}
${{\bar{\mathbf{P}}}_{\bm{\phi}}}(\cdot)$ maps the incremental update of $ {{\mathbf{R}}_{\mathtt{I}\mathtt{D}}} $ to the updated DVL translation integration. To balance accuracy and efficiency, the linearization is employed when the magnitude of $ \Delta {\bm{\phi}_{\mathtt{I}\mathtt{D}}} $ is smaller than a threshold $\sigma_{\phi}$.

\subsubsection{Approximation Iteration for Misalignment Calibration}
For the optimization \eqref{eq:DVLBodyVOpt} in the DVL misalignment calibration, $ {\Delta {_{\mathtt{D}_i}}{\bar{\mathbf{p}}}_{\mathtt{D}_i\mathtt{D}_j}} $ is linearized with respect to $ {_{\mathtt{D}_i}\mathbf{v}} $ by using
\begin{equation}
    \frac{\partial {\Delta {_{\mathtt{D}_i}}{\bar{\mathbf{p}}}_{\mathtt{D}_i\mathtt{D}_j}} }{\partial\ {{_{\mathtt{D}_i}\mathbf{v}}} } = \sum_{k=i}^{j-1} \ {\Delta\hat{\mathbf{R}}_{\mathtt{I}_i\mathtt{I}_k}} {{\mathbf{R}}_{\mathtt{I}\mathtt{D}}} \Delta t
\end{equation}
See its derivation in Appendix \ref{sec:Jacobian of Misalignment Calibration Approximation Iteration Derivation}. Therefore, the DVL translation pre-integration can be approximated as with an incremental update of $ {_{\mathtt{D}_i}\mathbf{v}} $:
\begin{equation}
{{\bar{\mathbf{P}}}_{\mathbf{v}}}(\Delta {{_{\mathtt{D}_i}\mathbf{v}}}) \doteq {\Delta {_{\mathtt{D}_i}}{\bar{\mathbf{p}}}_{\mathtt{D}_i\mathtt{D}_j}} + \frac{\partial {\Delta {_{\mathtt{D}_i}}{\bar{\mathbf{p}}}_{\mathtt{D}_i\mathtt{D}_j}} }{\partial {{_{\mathtt{D}_i}\mathbf{v}}} } \Delta {{_{\mathtt{D}_i}\mathbf{v}}}
\label{equa:misalignment_linearization}
\end{equation}
where $ {{\bar{\mathbf{P}}}_{\mathbf{v}}}(\cdot)$ defines the mapping from the incremental update of $ {_{\mathtt{D}_i}\mathbf{v}} $ to the updated DVL translation integration. Similarly, this linear approximation is used only when the norm of $ \Delta {{_{\mathtt{D}_i}\mathbf{v}}} $ is below a threshold $\sigma_{\mathbf{v}}$.

% The theoretical framework suggests that both the extrinsic parameters and the tranducer alignment can be optimized online.
It is worth mentioning that the calibration accuracy is largely contingent upon the quality of poses estimated through the vision-only bundle adjustment, whose reliability can be compromised in underwater environments due to the challenges aforementioned. Therefore, it is recommended that calibration is conducted in areas with discernible visual features, ideally with sufficient motion to ensure the observability of the parameters to calibrate \cite{xu2023observability}.

    \section{System Implementation}\label{sec:system_imp}

    We integrate the proposed acoustic-visual-inertial approach and the online sensor calibration method into the ORB-SLAM3 system \cite{orbslam3}, with a particular emphasis on incorporating DVL measurements to enhance accuracy and robustness. While the system retains the classic three-thread design (tracking, local mapping, and loop closure) and the multi-map strategy Atlas, all of these modules have been upgraded. The system overview is shown in Fig. \ref{fig:system_overview}. Specifically, we have made the following upgrades to ORB-SLAM3:

    \begin{figure}
        \centering
        \includegraphics[width=1.0\columnwidth,height=5.3cm]{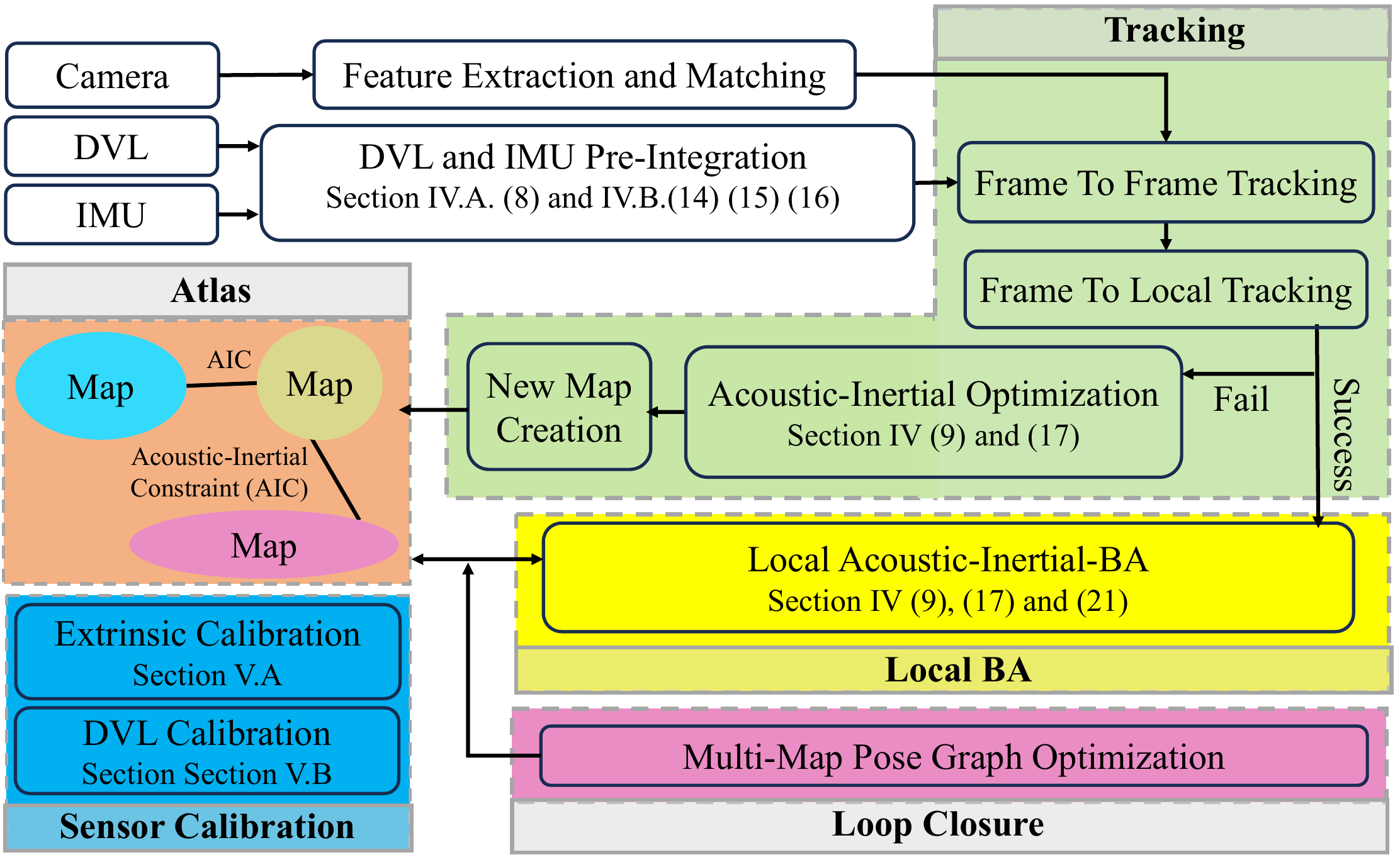}
        \caption{Overview of system implementation.}
        \label{fig:system_overview}
    \end{figure}

    \subsection{Sensor Data Processing}
    The system takes stereo image pairs from a camera and performs stereo matching to provide depth with an absolute scale.
    % from cameras and extract ORB feature and compute descriptor, then a . Then feature matching is performed to build data association properly.
    The DVL measurement is pre-integrated using \eqref{equa:dvl_decoupled}, and the IMU data is processed using \eqref{equa:gyros_integration2}, \eqref{equa:acc_v_integration2} and \eqref{equa:acc_t_integration2}. % which calculate and store the pre-integration of the DVL and IMU measurement to avoid the repeat computation during optimization and speed up the computation.

    \subsection{Tracking}
    The tracking thread is mainly to track image frames using the predicted poses from the DVL and IMU measurements. % Furthermore the pose prediction is adopted as the initial value  of frame-to-local track when frame-to-frame tracking is failed.
    When the frame-to-local tracking fails, e.g., due to poor image quality, acoustic-inertial optimization is performed for pose estimation based on the DVL and IMU residuals in \eqref{equa:dvl_measurement_model} and  \eqref{equa:gyros_integration3}. %and  the current map to Atlas and create a map when the visual tracking is recovered.
    Meanwhile, an acoustic-inertial constraint is derived and added between consecutive sub-maps in the Atlas module. Thanks to the acoustic-inertial optimization, pose tracking is more accurate and reliable in poor visual conditions.

    \subsection{Local BA}
    The local BA module maintains a sliding window of up to 10 keyframes. When a new keyframe is inserted, the acoustic-inertial BA is performed to fuse the data from DVL, IMU, and camera using the residuals in \eqref{equa:dvl_measurement_model}, \eqref{equa:gyros_integration3} and \eqref{equa:camera_residual}.

    \subsection{Atlas}
    The Atlas module manages sub-maps. Different from ORB-SLAM3 which divides sub-maps into active and inactive ones, the proposed system maintains a consistent map which interlinks all sub-maps through the acoustic-inertial constraints.

    \subsection{Loop Closure}
    The loop closure thread handles loop fusion within the same map and sub-map merging. % We only improved the map merging part.
    ORB-SLAM3 \cite{orbslam3} only corrects the active map and a sub-map to be merged. Instead, the proposed method corrects all sub-maps via the acoustic-inertial constraints. Therefore, the overall map and pose errors can be corrected for each time when a sub-map merging is performed.

    \subsection{Sensor Calibration}
    % Subsequent to image processing, the system evaluates the need for
    When the online sensor calibration module is activated, either the extrinsic calibration or the DVL calibration is conducted using the sensor data between 10 keyframes. % Un-calibrated systems proceed to an  calibration phase.

% \fi
% \newpage
% \iffalse

\section{Experimental Results}\label{sec:experiments}

\begin{figure}
    \centering
    \subfigure[BlueROV2 for WaveTank test]{
    \includegraphics[width=0.45\columnwidth,height=3.65cm]{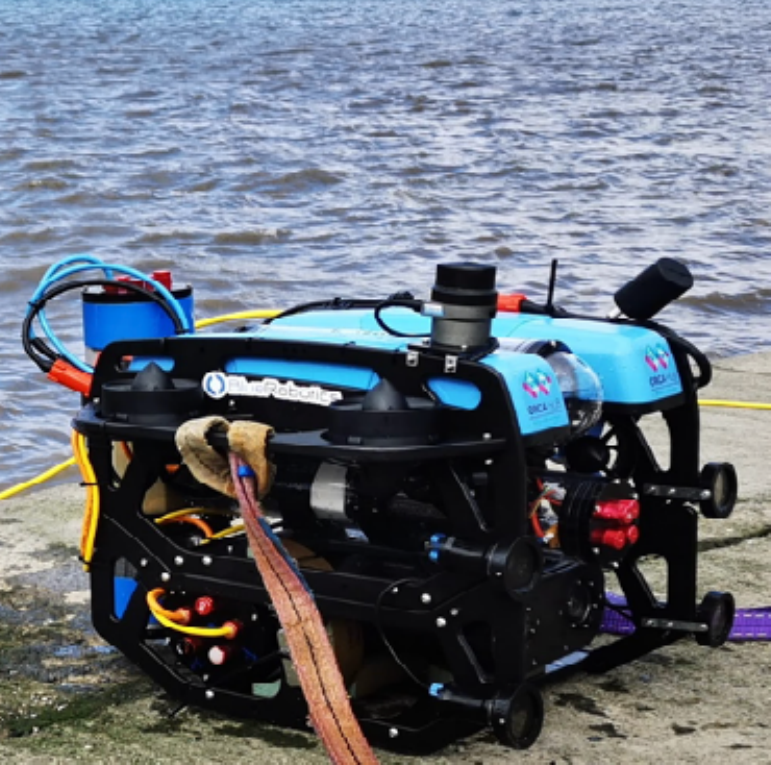}}
    \subfigure[Falcon vehicle for Offshore test]{
    \includegraphics[width=0.45\columnwidth]{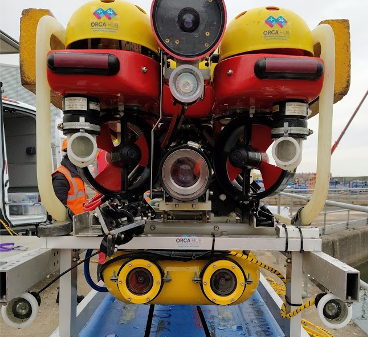}}
    \caption{Underwater vehicles used for experiments.}
    \label{fig:vehicles}
\end{figure}

% on a self-collected dataset named WaveTank dataset with Ground Truth (GT)
The performances of the proposed SLAM and calibration methods are evaluated in this section.    % by comparing our method with other state-of-the-art visual-inertial SLAM/odometry. An extensive number of repeat experiments are performed to provide detailed quantitative and qualitative analyses.
% We also validate the proposed SLAM algorithm and its robustness in real-world offshore environments which represent extremely challenging visual scenarios. Only qualitative results are provided due to the lack of GT at the offshore environments.
\begin{table}
    \centering
    \caption{Statistics of the WaveTank sequences.}
    \label{table:dataset_config}
    {\setlength{\tabcolsep}{2pt}
    \begin{tabular}[h]{M{3.2cm}||M{2.8cm}|M{0.8cm}|M{1.2cm}}
        \hline
        \textbf{Sequence} & \textbf{Velocity Distribution (0 to 0.5 m/s)} & \textbf{Light} & \textbf{GT \%} \\
        \hline
        \hline
        Structure Easy(SE) & \includegraphics[width=2.2cm,height=2cm,keepaspectratio]{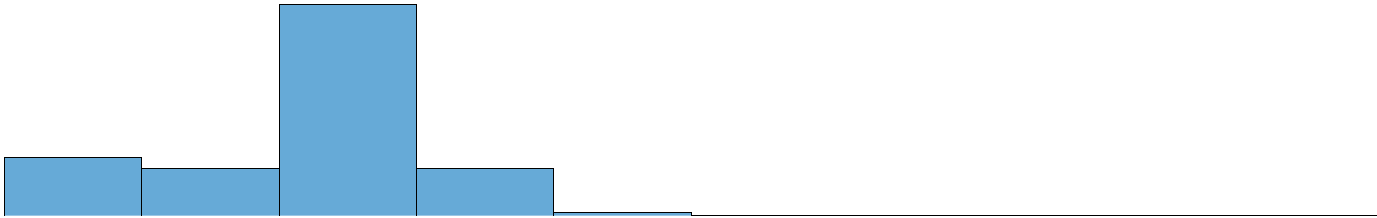} & on & 99.37\% \\
        Structure Medium(SM) & \includegraphics[width=2.2cm,height=2cm,keepaspectratio]{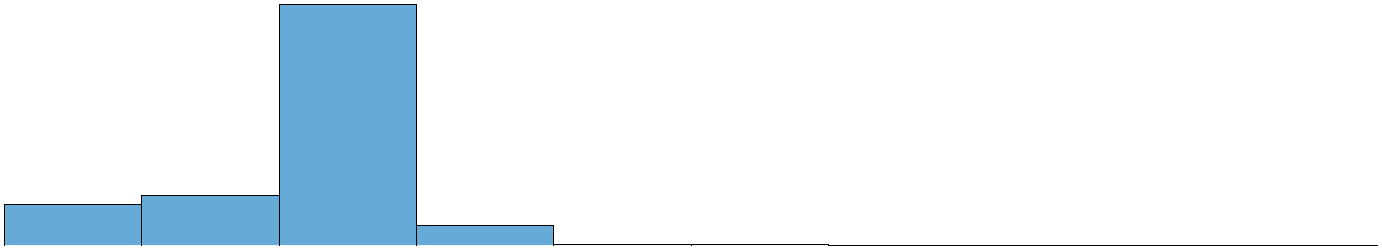} & off & 99.67\% \\
        Structure Hard(SH) & \includegraphics[width=2.2cm,height=2cm,keepaspectratio]{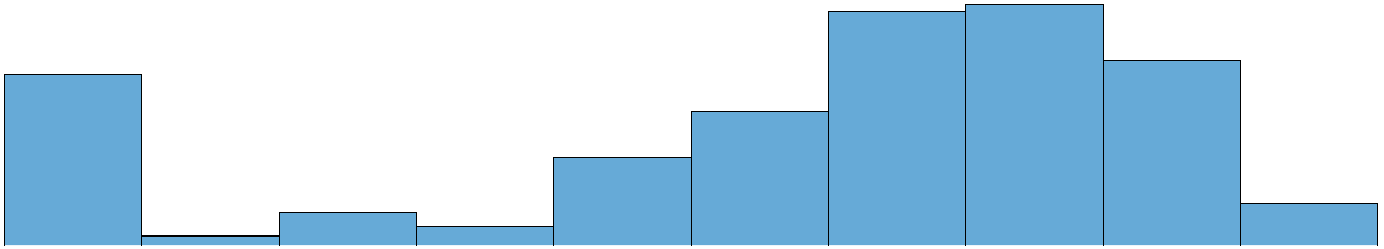} & off & 84.65\% \\
        HalfTank Easy(HE) & \includegraphics[width=2.2cm,height=2cm,keepaspectratio]{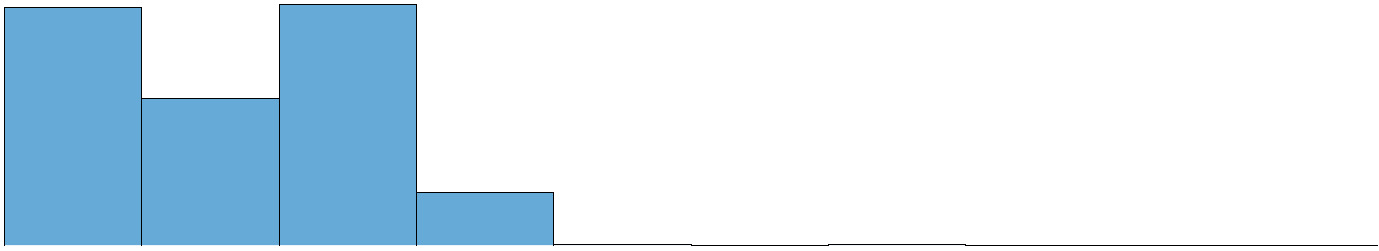} & on & 36.96\% \\
        HalfTank Medium(HM) & \includegraphics[width=2.2cm,height=2cm,keepaspectratio]{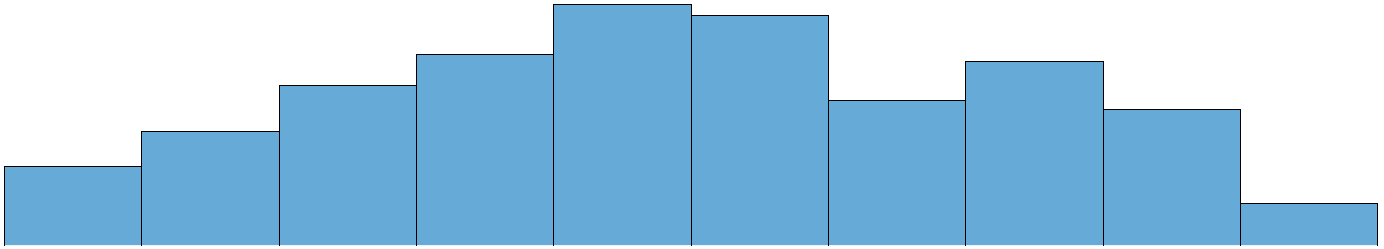} & on & 44.19\% \\
        HalfTank Hard(HH) & \includegraphics[width=2.2cm,height=2cm,keepaspectratio]{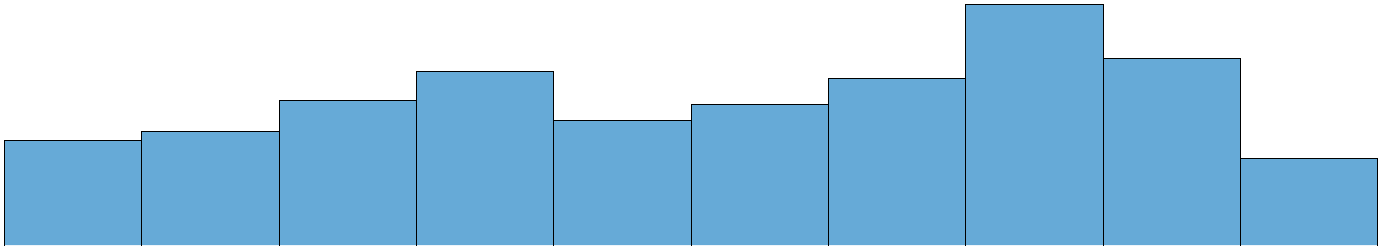} & off & 41.40\% \\
        WholeTank Medium(WM) & \includegraphics[width=2.2cm,height=2cm,keepaspectratio]{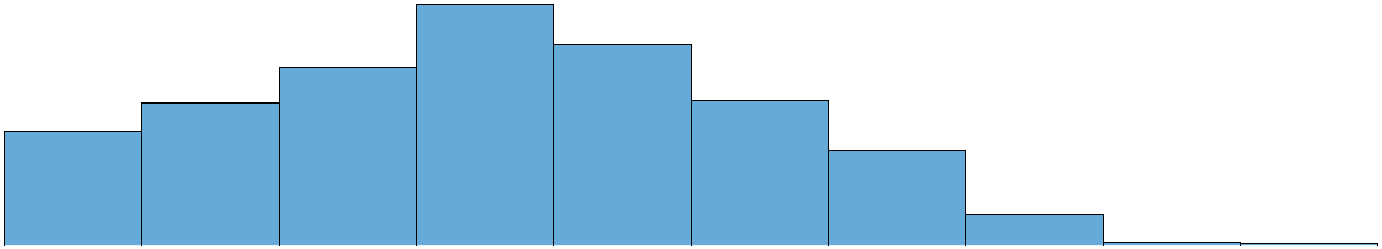} & on & 60.73\% \\
        WholeTank Hard(WH) & \includegraphics[width=2.2cm,height=2cm,keepaspectratio]{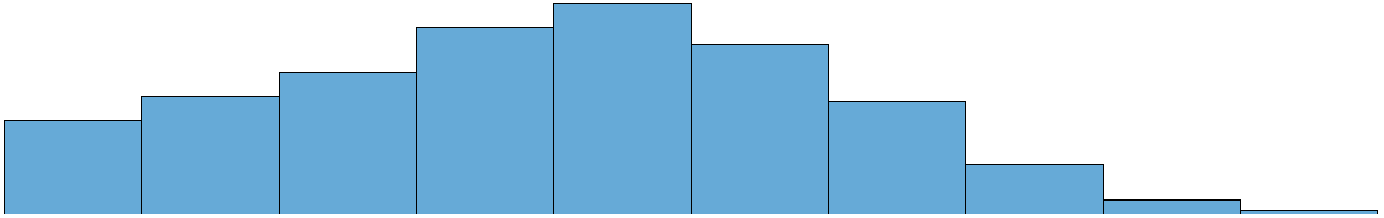} & off & 18.97\% \\
        \hline
    \end{tabular}
    }
\end{table}
%\Shida{need to update HM,WM,WH data}

\subsection{Experiment Settings and Datasets}

Due to the lack of a public underwater dataset with the required sensor suite (i.e., stereo camera, IMU and DVL), two datasets - WaveTank and Offshore datasets - are collected for the experimental evaluation using the two underwater remotely operated vehicles (ROVs) shown in Fig. \ref{fig:vehicles}.

\begin{figure}
    \centering
    \includegraphics[width=0.9\linewidth]{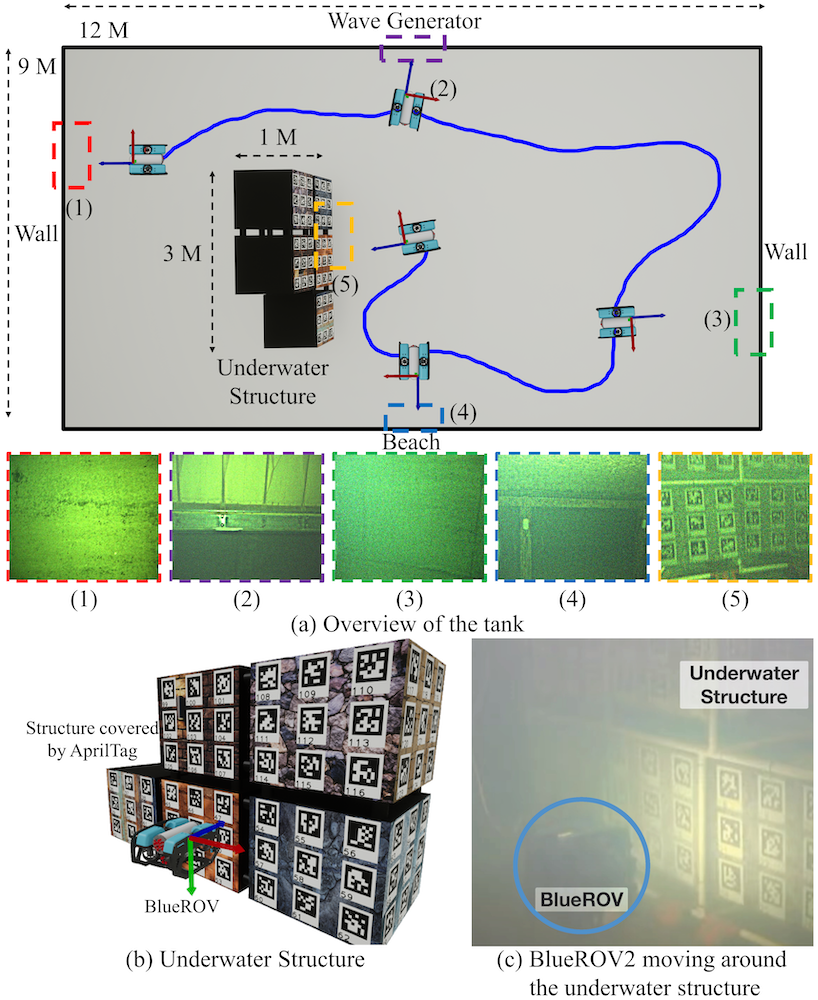}
    \caption{Experiment settings of the self-collected WaveTank dataset.}
    % \SenDraft{Change (b) to match (c)}
    \label{fig:underwater_structure}
\end{figure}

\subsubsection{WaveTank Dataset}
The WaveTank dataset is collected in a $9\times12$ meter tank with a structure deployed in the middle, as illustrated in Fig. \ref{fig:underwater_structure}. A BlueROV2 vehicle equipped with a WaterLinked A50 DVL running at 5 Hz, a MICROSTRAIN 3DM-GX5-AHRS IMU at 330 Hz and a custom underwater stereo camera at 20 Hz \cite{luczynski2019stereo} is used to collect DVL, IMU and stereo data.
In order to have ground truth (GT) for quantitative evaluation, the structure is covered with AprilTag markers \cite{wang2016iros_april}, as shown in Fig. \ref{fig:underwater_structure}. A fiducial-marker based SLAM is developed to provide GT robot poses.
Specifically, the marker poses are initially calibrated using a calibration sequence collected by allowing the vehicle to move slowly and smoothly, generating images where markers are clearly recognizable. Additionally, multiple loops are closed to ensure the precision of the landmark poses. This process is crucial to ensure the high accuracy of the marker poses. % In total, 125 markers are distributed and mapped on the structure. % Each marker is positioned under the marker frame, the origin of which is the location of the marker with ID 0. % , denoted as the $\mathtt{M}_0$ frame. % The x-axis of the $\mathtt{M}_0$ frame points to the right, the y-axis points downwards, and the z-axis points towards the interior of the marker.
Following the calibration, the location of each marker is treated as a priori information, and the camera pose is subsequently obtained by being localized in relation to the calibrated marker poses. %, $\mathbf{T}_{\mathtt{C}_0\mathtt{C}_i}$, on each individual sequence.
% This is facilitated by the application of fiducial-marker-based SLAM. As such,
Therefore, whenever a marker is recognized from an image, an accurate GT camera pose becomes available. Note the AprilTag markers are only used for generating GT, and they are not used in the SLAM systems to be evaluated. The dataset, along with its GT and evaluation toolset, will be introduced in detail and made publicly accessible in a separate dataset paper.

%\Sen{Add some Onboard camera images showing the challenging environments of the experiments}

Eight sequences are employed for the quantitative analysis. The sequences are obtained under varying scenarios and configurations, thereby resulting in assorted levels of difficulty, as summarized in Table \ref{table:dataset_config}. Factors, such as high speed or the lack of light, significantly augment the difficulty of the sequences. Moreover, the AprilTag coverage rate indicates the proportion of images with AprilTag markers visible to estimate the GT poses, which roughly indicates the visual conditions.
% This rate is influenced by aspects, such as motion blur and where the images are captured, as a rough metric for describing the visual conditions.

According to the trajectories, three categories exist: structure, half tank, and whole tank. Structure sequences are exclusively collected around the underwater structure and a majority of their images have visible markers. The half-tank sequences are collected along a loop traversing half of the tank. The whole tank sequences are gathered along a loop traversing the entire tank. Therefore, the half-tank and whole-tank sequences contain a larger number of images from textureless areas without GT.

\subsubsection{Offshore Dataset}
The Offshore dataset is collected at an offshore wind farm using a Saab Seaeye Falcon ROV as shown in Fig. \ref{fig:vehicles}.
% which has a MEMS IMU(MICROSTRAIN 3DM-GX5-AHRS, running at 100 HZ), DVL(Teledyne Explorer DVL) and our stereo camera\cite{luczynski2019stereo}.
Since there is no GT in the open sea, our results were compared to those generated by COLMAP  \cite{schoenberger2016sfm}, a widely-used offline Structure from Motion (SfM) and Multi-View Stereo (MVS) pipeline.
%\Shida{do we need images here? maybe images of we deploy the Falcon from boat, and images of the wind turbine?}

% \Sen{evaluation metrics: relative errors do not reflect reality for sequences whose GT of some path segments is unavailable, only reported for sequences with full GT. Absolute errors whenever GT is available }(done)

% \begin{figure}
%     \centering
%     \subfigure[WaveTank Dataset]{
%     \includegraphics[width=0.4\columnwidth]{image/WaveTank.jpg}}
%     \subfigure[Falcon]{
%     \includegraphics[width=0.4\columnwidth]{IROS/image/falcon.png}}
%     \caption{Experiment settings and vehicles of the self-collected WaveTank and Offshore datasets.}
%     \label{fig:datasets}
% \end{figure}

\begin{table}
    \centering
    \caption{Competing SLAM methods}
    \begin{tabular}[h]{M{2.3cm}||M{2cm}|M{1.5cm}|M{1.5cm}}
      \hline
      \textbf{Method} & \textbf{Sensor} & \textbf{Front-end} & \textbf{Back-end} \\
      \hline
      \hline
      Ours & Stereo cameras, IMU, DVL & Feature-based & Graph based \\
      \hline
      Previous Work \cite{xu2021underwater} & Stereo cameras, Gyroscope, DVL & Feature based & EKF, Graph based \\
      \hline
      SVIN2 \cite{rahman2022svin2} & Stereo cameras, IMU & Feature based & Graph based \\
      \hline
      ORB-SLAM3 \cite{orbslam3}& Stereo cameras, IMU& Feature based & Graph based \\
      \hline
      VINS-Fusion \cite{lin2018VINS-Mono}& Stereo cameras, IMU & Optical-Flow & Graph based \\
      \hline
      Basalt \cite{usenko2019basalt} & Stereo cameras, IMU & Optical-Flow & Graph based \\
      \hline
    \end{tabular}
    \label{table:method_config}
  \end{table}

\subsection{Competing Methods}
    In our study, we conduct a comparative analysis of our results with five state-of-the-art SLAM systems, encompassing two underwater SLAM works, our previous work  \cite{xu2021underwater} and SVIN2 \cite{rahman2022svin2}, and three visual-inertial SLAM works, VINS-Fusion \cite{lin2018VINS-Mono}, ORB-SLAM3 (ORB3) \cite{orbslam3} and Basalt \cite{usenko2019basalt}. Our previous work \cite{xu2021underwater} integrates a DVL, a gyroscope, and a stereo camera within a loosely-coupled framework. In the absence of a downward-looking sonar system on our ROVs, SVIN2 \cite{rahman2022svin2} operates in its stereo-inertial mode. Similarly, VINS-Fusion, ORB-SLAM3 and Basalt are configured to function in their stereo-inertial mode. The distinctive features of these systems are summarized in Table \ref{table:method_config}, providing a comprehensive overview of their respective configurations and capabilities.

To ensure fair comparisons, we used the same parameters, including camera, IMU, and extrinsic parameters, as much as possible for all methods. For unique parameters specific to each method, we used the default settings provided by the authors. Specifically, VINS-Fusion, ORB SLAM3, and Basalt used the EuRoc settings from their original GitHub repositories. SVIN2 used the ``svin\_stereorig\_v2'' settings applied to the cave sequence in the original repository. Our previous work used the same parameters as the proposed method.

\subsection{Evaluation Metrics and Preprocessing}
Given that GT is only accessible when the AprilTag markers are visible, i.e., when the robot faces towards the structure,
% the acquisition of the GT traveled distance becomes impossible when the vehicle traverses areas devoid of AprilTag coverage.
% This situation culminates in a falsely magnified relative error.
% Consequently,
only pose estimates with associated GT are used for quantitative evaluation.
For each sequence, we execute each method ten times to compute an average error of the root-mean-square errors (RMSE) and standard deviation (STD) of the absolute error. % and summarize the results in Table \ref{table:odom_rmse} and \ref{table:slam_rmse}.

\begin{table*}
    \fontsize{6.9}{9}\selectfont
    \caption{Odometry performance in WaveTank dataset averaging 10 runs. }
    \label{table:odom_rmse}
    \begin{center}
    \begin{tabular}{@{\hskip 1pt} l @{\hskip 1pt}||@{\hskip 2pt} c @{\hskip 2pt}|@{\hskip 2pt} c @{\hskip 2pt}|@{\hskip 2pt} c @{\hskip 2pt}|@{\hskip 2pt} c @{\hskip 2pt}|@{\hskip 2pt} c @{\hskip 2pt}|@{\hskip 2pt} c @{\hskip 2pt}||@{\hskip 2pt} c @{\hskip 2pt}|@{\hskip 2pt} c @{\hskip 2pt}|@{\hskip 2pt} c @{\hskip 2pt}|@{\hskip 2pt} c @{\hskip 2pt}|@{\hskip 2pt} c @{\hskip 2pt}|@{\hskip 2pt} c @{\hskip 2pt}}
    \hline
    & \multicolumn{6}{c||}{\makecell[c]{Translation Error \\ RMSE (in meter) / STD }} & \multicolumn{6}{c}{\makecell[c]{Rotation Error \\RMSE (in degree) / STD }} \\
    \cline{2-13}
       & \makecell[c]{Ours}  & \makecell[c]{ Previous\\ Work} & \makecell[c]{SVIN2} & \makecell[c]{ORB3} & \makecell[c]{VINS} & \makecell[c]{Basalt}  & \makecell[c]{Ours}  & \makecell[c]{Previous\\ Work} & \makecell[c]{SVIN2} & \makecell[c]{ORB3} & \makecell[c]{VINS} & \makecell[c]{Basalt} \\
       \hline 
       Structure Easy & \textbf{0.07} / \textbf{0.03} & 0.21 / 0.12 & 0.09 / 0.03 & 0.28 / 0.09 & 0.13 / 0.04 & 0.12 / 0.06 & \textbf{1.62} / 0.68 & 4.88 / 2.49 & 1.84 / \textbf{0.49} & 5.45 / 1.52 & 2.15 / 0.72 & 2.57 / 1.02\\ 
       \hline 
       Structure Medium & \textbf{0.18} / \textbf{0.08} & 0.54 / 0.30 & 2.94 / 1.64 & 3.30 / 1.08 & NaN / NaN & 3.66 / 1.81 & \textbf{4.17} / \textbf{1.57} & 11.19 / 5.05 & 74.56 / 47.51 & 93.87 / 52.48 & 66.55 / 22.17 & NaN / NaN\\ 
       \hline 
       Structure Hard & \textbf{0.50} / 0.24 & 0.50 / \textbf{0.23} & 3.26 / 1.43 & 2.73 / 1.45 & NaN / NaN & 4.48 / 2.10 & \textbf{5.63} / 2.76 & 5.81 / \textbf{2.35} & 16.57 / 6.83 & 98.37 / 60.33 & 35.61 / 10.83 & 21.28 / 12.26\\ 
       \hline 
       HalfTank Easy & \textbf{0.28} / \textbf{0.17} & 1.69 / 1.10 & 6.01 / 4.33 & 2.69 / 1.45 & 29.83 / 20.74 & 3.12 / 2.14 & \textbf{2.04} / \textbf{0.74} & 29.87 / 18.54 & 3.00 / 1.65 & 75.35 / 37.99 & 8.38 / 5.03 & 4.02 / 1.80\\ 
       \hline 
       HalfTank Medium & \textbf{0.29} / \textbf{0.14} & 0.44 / 0.22 & 3.40 / 1.76 & 0.74 / 0.38 & NaN / NaN & 1.87 / 0.63 & \textbf{4.29} / \textbf{1.95} & 8.49 / 3.75 & 32.13 / 15.37 & 16.24 / 7.42 & 64.33 / 29.31 & 35.52 / 17.42\\ 
       \hline 
       HalfTank Hard & \textbf{0.36} / \textbf{0.22} & 0.58 / 0.37 & 77.60 / 55.07 & 1.10 / 0.70 & NaN / NaN & 5.14 / 3.34 & \textbf{3.84} / \textbf{1.99} & 10.20 / 5.95 & 18.96 / 10.37 & 19.03 / 13.10 & 97.81 / 44.08 & 63.50 / 37.69\\ 
       \hline 
       WholeTank Medium & \textbf{0.52} / \textbf{0.28} & 1.34 / 0.73 & 0.72 / 0.41 & 1.18 / 0.71 & 13.08 / 7.22 & 2.18 / 1.25 & \textbf{3.34} / 1.38 & 12.49 / 5.51 & 5.29 / 2.12 & 27.36 / 14.34 & NaN / NaN & 3.38 / \textbf{1.35}\\ 
       \hline 
       WholeTank Hard & \textbf{0.22} / \textbf{0.12} & 1.11 / 0.83 & 0.83 / 0.65 & 2.96 / 2.49 & NaN / NaN & 0.95 / 0.76 & \textbf{3.99} / \textbf{1.37} & 8.14 / 4.17 & 4.69 / 2.35 & 30.91 / 24.97 & 29.24 / 11.20 & 4.61 / 2.56\\ 
       \hline 
    \end{tabular}
    \end{center}
\end{table*}

\subsubsection{Preprocessing}
The preprocessing of pose estimates from different methods is a crucial step before conducting evaluations, given that these pose estimates are represented in varying reference frames and originate from different timestamps based on the implementations of the methods.

    % \paragraph{Quantitative Evaluation}
% We present tracking performance as in our quantitative evaluation. The quantitative evaluation includes ATE Table \ref{table:odom_rmse} and \ref{table:slam_rmse}, error heat map in Fig. \ref{fig:odom_error_map} and \ref{fig:slam_error_map}, and error with timestamp at Fig. \ref{fig:error_time_odom_he}, \ref{fig:error_time_odom_wh}, \ref{fig:error_time_slam_he} and \ref{fig:error_time_slam_wh}. By tracking performance, we refer to the evaluation of the pose emitted from the tracking thread, devoid of any further corrections brought by local BA or sliding window optimization at the backend.

\begin{figure}
    \centering
    \subfigure[Errors on odometry translation.]{
    \includegraphics[width=1\columnwidth]{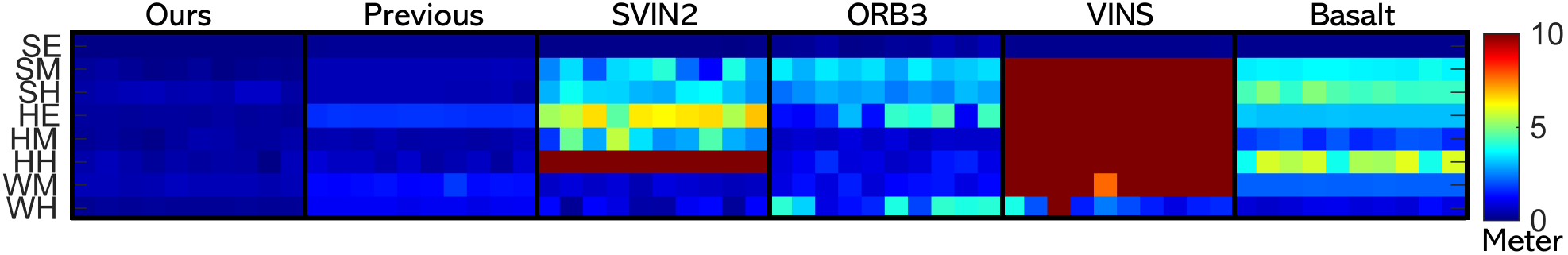}}
    \subfigure[Errors on odometry rotation.]{
    \includegraphics[width=1\columnwidth]{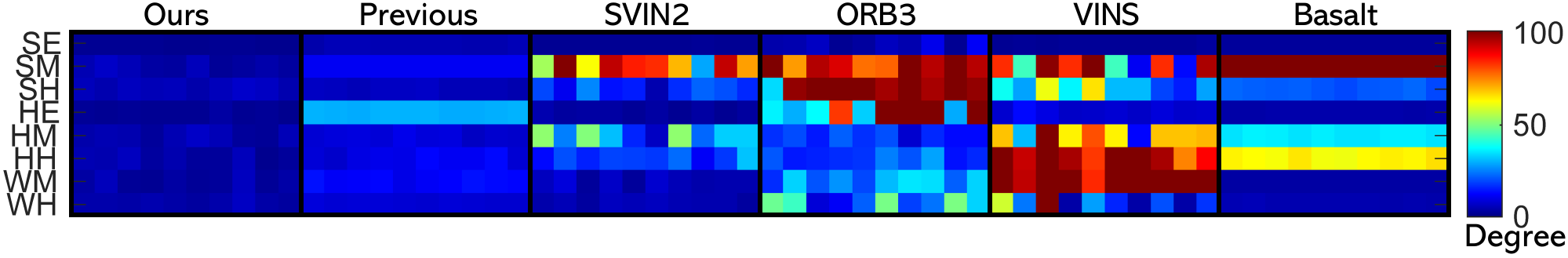}}
    \caption{ Odometry results of 10 runs on all WaveTank sequences.}
    \label{fig:odom_error_map}
\end{figure}

 The preprocessing can be divided into three parts: % Frame Transformation, Time Synchronization, and Trajectory Alignment.
\begin{itemize}
    \item Frame Transformation: % Initially, we use a ROS Python script to gather pose data from various sources.
    Since the implementations of the competing methods use different reference frames (some in camera frames, some in IMU frames) for their poses, we first convert all these poses into the same camera frame. % This step, known as frame transformation, changes the poses from their original frames to the camera frame. This makes data easier to compare and analyze.
    \item Time Synchronization: For a trajectory estimated by a method, its poses are matched with GT based on timestamps. % and eliminate all unmatched poses. This procedure is termed time synchronization.
    \item Trajectory Alignment: A SE3 transformation is executed to align the first pose of a trajectory with the GT trajectory. This step ensures all trajectory starts from the same origin.
\end{itemize}

\subsection{Evaluation on Odometry}

For the odometry evaluation, the loop-closure detection functions of all methods are disabled. % to evaluate the localisation performance without loop closure.

    \subsubsection{Quantitative Evaluation}

% - quantitative: table, heatmap, error-time plots (2 sequences(structure easy+medium))
% - qualitative: reconstructed 3D point clouds,

% \subsubsection{Quantitative Experiments}

\begin{figure*}
    \subfigure{\includegraphics[width=2.05\columnwidth]{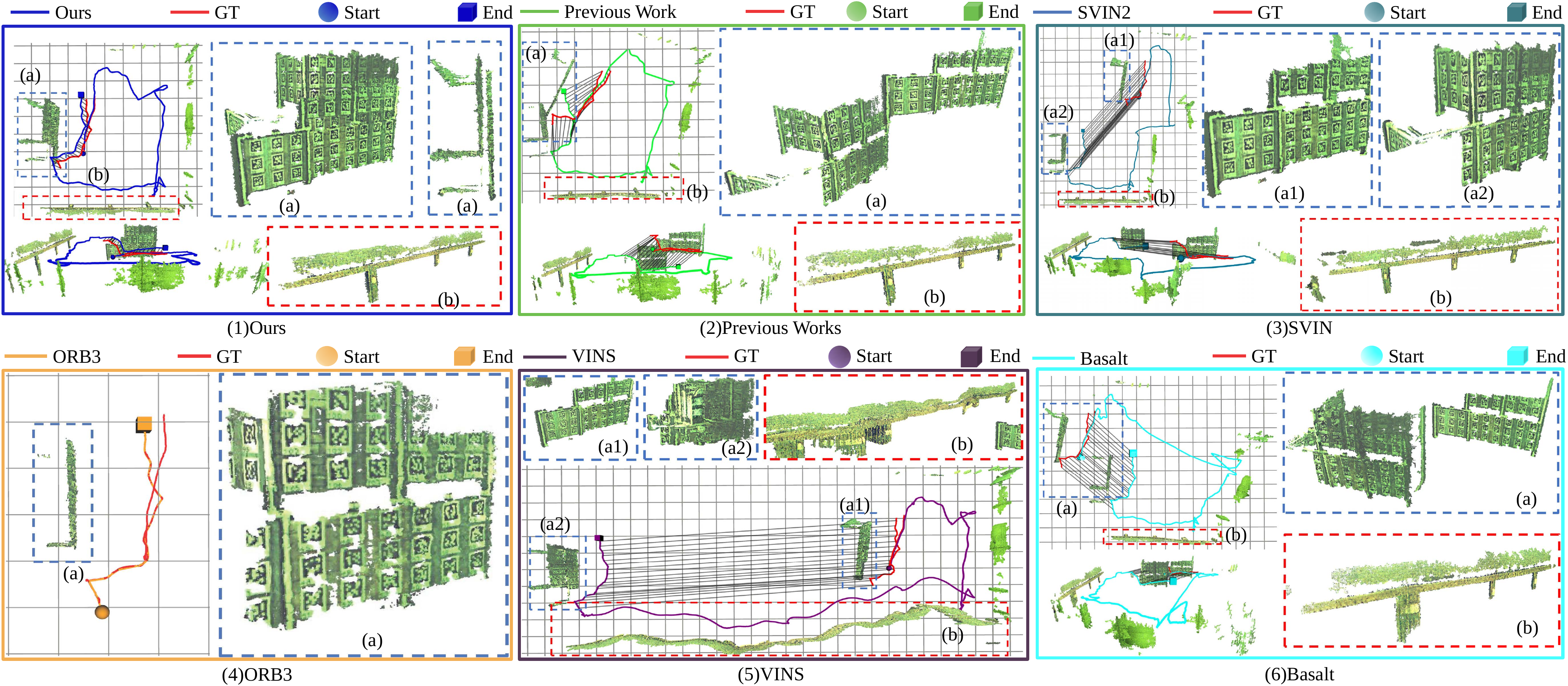}}
    \caption{Estimated trajectories and dense reconstruction of all methods on HalfTank Easy sequence in odometry mode.
    $\iscircle$ and $\Box$ mark the start and the end of a trajectory.
    For each subfigure, the left part (with a grid background) shows the dense map and the estimated trajectory with errors to the GT as $-$, and its map sections highlighted with dashed lines are shown correspondingly. A grid cell indicates 1 m$^2$.}
    \label{fig:dense_odom_he}
\end{figure*}

    The RMSE and STD are detailed in Table \ref{table:odom_rmse}, and the RMSE of the 10 runs are given in Fig. \ref{fig:odom_error_map} as error maps. These results clearly demonstrate the superiority of our proposed method across all WaveTank sequences in terms of both translational and rotational accuracy. This can be attributed to the tightly-coupled integration of acoustic, visual, and inertial sensing in the proposed SLAM methodology.
    %  coupled with a meticulously designed system implementation.

    In comparison, our previous work surpasses other visual-inertial SLAM systems in translational accuracy, largely due to the incorporation of the DVL sensor. However, its rotational performance is compromised. We hypothesize that this is primarily because its loosely-coupled framework does not facilitate real-time correction of gyroscope bias. Additionally, the absence of an accelerometer renders the absolute roll and pitch angles unobservable, further impacting its rotational accuracy. To further demonstrate this, a detailed ablation study is presented in Section \ref{sec:ablation_acc_bias}.

    \begin{figure*}
        \subfigure{\includegraphics[width=2.05\columnwidth]{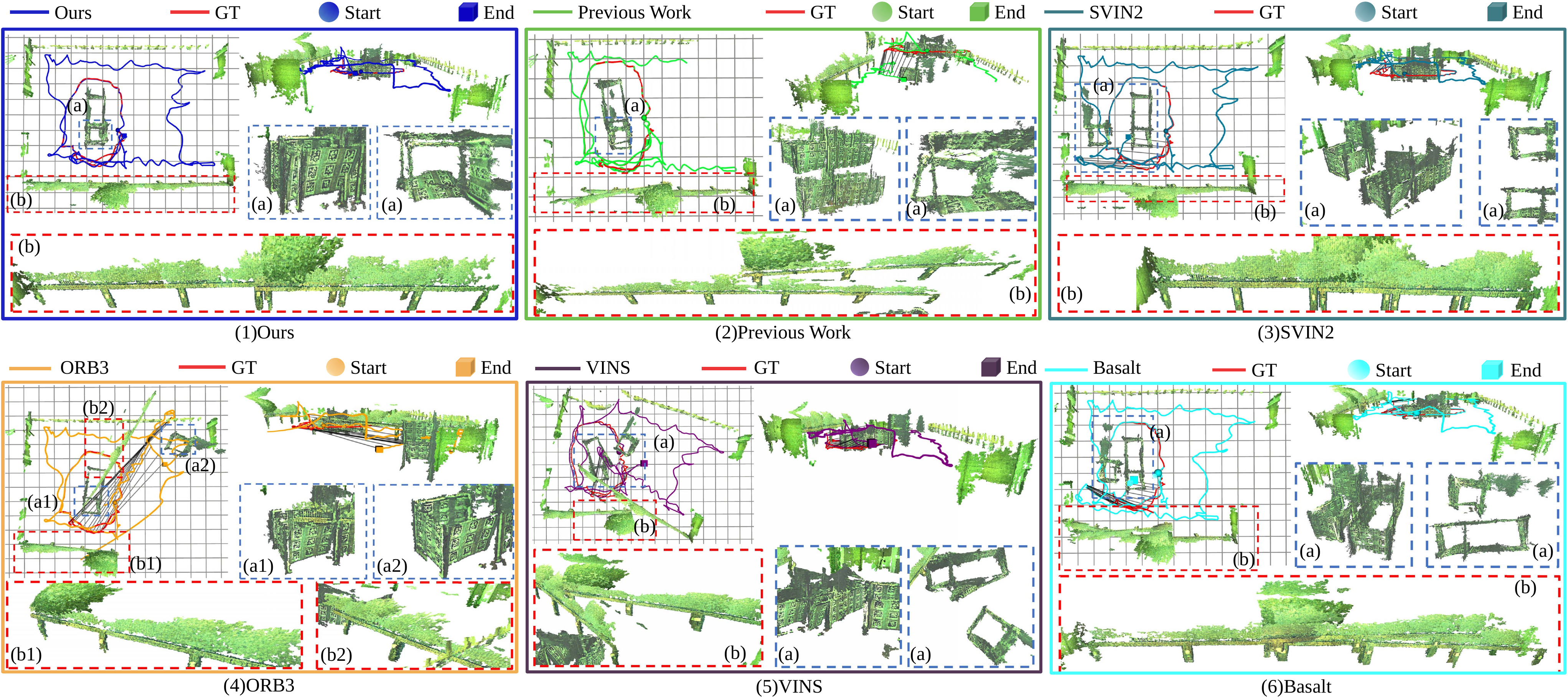}}
        \caption{Estimated trajectories and dense maps on WholeTank Hard sequence in odometry mode. The figure legends and grid size are the same as Fig. \ref{fig:dense_odom_he}.}
        \label{fig:dense_odom_wh}
    \end{figure*}

    \begin{figure}
        \centering
        \includegraphics[width=0.5\textwidth]{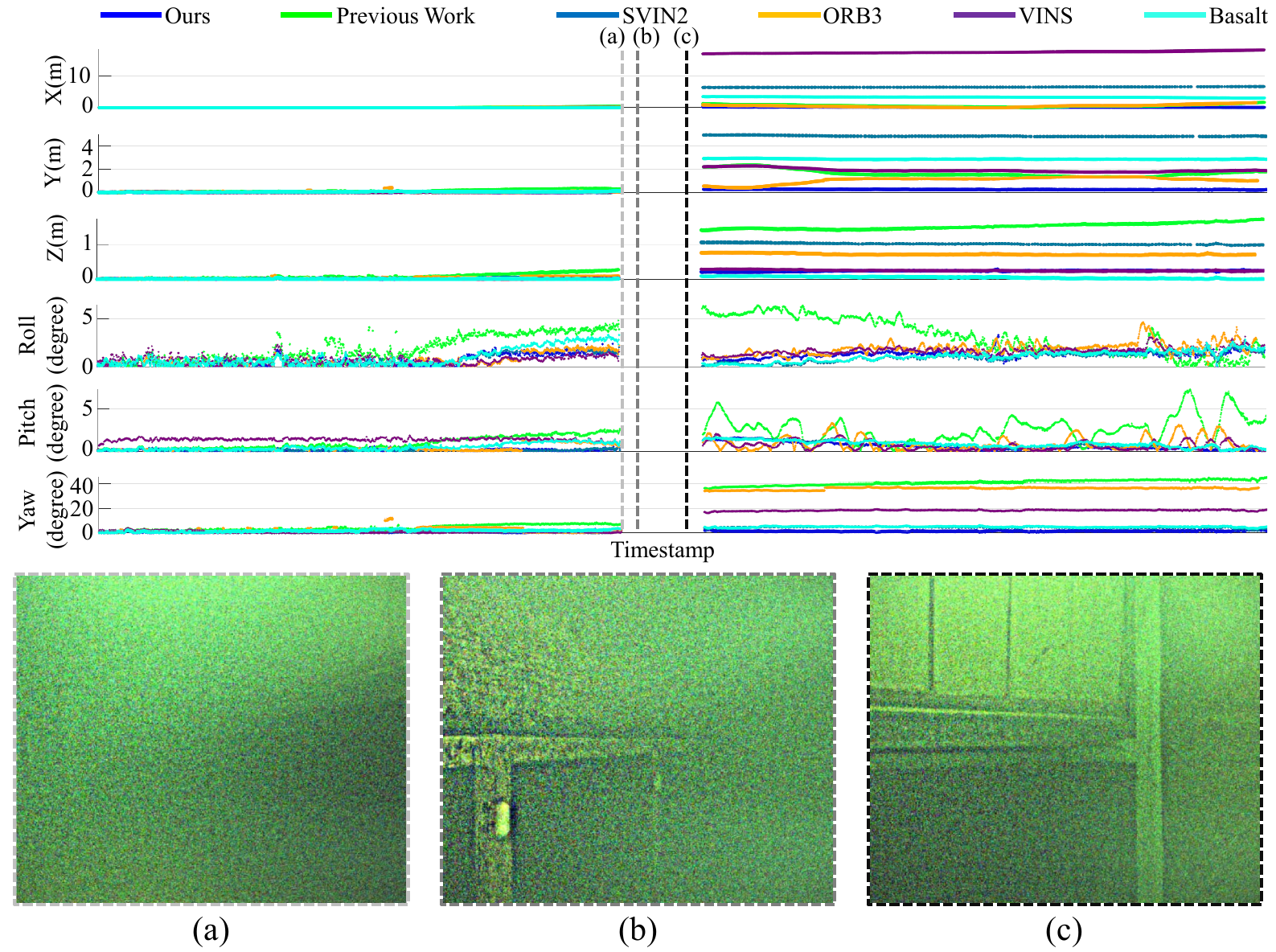}
        \caption{Odometry errors on HalfTank Easy sequence and visually degraded cases. The timestamp gaps indicate the times without GT poses. (a), (b) and (c) depict the challenging visual conditions in visually degraded cases.}
        \label{fig:error_time_odom_he}
    \end{figure}

    Other systems, SVIN2, ORB3, VINS and Basalt, exhibit significant errors in translation, particularly in sequences characterized by challenging visual conditions. These conditions include scenarios with insufficient features in the images or inadequate lighting, underlining the limitations of these systems in handling the challenges in underwater environments.
    They also show substantial variability in accuracy across different sequences, and this inconsistency is further evident when observing different runs of the same sequence.
    SVIN2 achieves results comparable to our proposed method on the SE and WH sequences, where the image quality is generally high. In the WH sequence specifically, SVIN2 surpasses other visual-inertial approaches in performance, though it still exhibits a noticeable increase in error compared to our method. However, in more challenging sequences, SVIN2's performance aligns with that of other systems, showing significant odometry drifts.
    ORB3 performs well on the SE, HM, HH and WM sequences but fails on other sequences. When ORB3 loses tracking, it resets its pose to the origin and ceases to publish poses. Hence, only a few poses, mainly in the beginning where  features are clearly visible, are available. Despite the seemingly acceptable error, in reality, ORB3 mostly fails on these sequences.
    VINS shows considerable errors in most sequences. Unlike ORB3, when confronted with poor-quality images that cause loss of tracking, VINS drifts rapidly resulting in high errors.
    Basalt only achieves comparable performance to our proposed method on the SE sequence, drifting on the other sequences with more challenging visual conditions.

    \begin{table*}
        \fontsize{6.8}{9}\selectfont
        \caption{SLAM performance in WaveTank dataset averaging 10 runs. }
        \label{table:slam_rmse}
        \begin{center}
        \begin{tabular}{@{\hskip 1pt} l @{\hskip 1pt}||@{\hskip 2pt} c @{\hskip 2pt}|@{\hskip 2pt} c @{\hskip 2pt}|@{\hskip 2pt} c @{\hskip 2pt}|@{\hskip 2pt} c @{\hskip 2pt}|@{\hskip 2pt} c @{\hskip 2pt}|@{\hskip 2pt} c @{\hskip 2pt}||@{\hskip 2pt} c @{\hskip 2pt}|@{\hskip 2pt} c @{\hskip 2pt}|@{\hskip 2pt} c @{\hskip 2pt}|@{\hskip 2pt} c @{\hskip 2pt}|@{\hskip 2pt} c @{\hskip 2pt}|@{\hskip 2pt} c @{\hskip 2pt}}
        \hline
        & \multicolumn{6}{c||}{\makecell[c]{Translation Error \\ RMSE (in meter) / STD }} & \multicolumn{6}{c}{\makecell[c]{Rotation Error \\RMSE (in degree) / STD }} \\
        \cline{2-13}
           & \makecell[c]{Ours}  & \makecell[c]{ Previous\\ Work} & \makecell[c]{SVIN2} & \makecell[c]{ORB3} & \makecell[c]{VINS} & \makecell[c]{Basalt}  & \makecell[c]{Ours}  & \makecell[c]{Previous\\ Work} & \makecell[c]{SVIN2} & \makecell[c]{ORB3} & \makecell[c]{VINS} & \makecell[c]{Basalt} \\
           \hline 
           Structure Easy & 0.06 / 0.03 & 0.18 / 0.11 & 0.09 / 0.04 & 0.20 / 0.09 & 0.22 / 0.07 & \textbf{0.04} / \textbf{0.01} & 1.51 / 0.67 & 4.11 / 2.23 & 1.76 / 0.45 & 4.18 / 1.82 & 2.32 / 0.67 & \textbf{0.89} / \textbf{0.38}\\ 
           \hline 
           Structure Medium & \textbf{0.16} / \textbf{0.10} & 0.49 / 0.29 & 2.11 / 1.36 & 3.49 / 1.17 & NaN / NaN & NaN / NaN & \textbf{3.39} / \textbf{1.48} & 10.98 / 5.41 & 57.51 / 39.42 & 90.01 / 49.18 & 65.96 / 23.40 & NaN / NaN\\ 
           \hline 
           Structure Hard & \textbf{0.31} / \textbf{0.17} & 0.43 / 0.23 & 3.59 / 1.54 & 2.93 / 1.30 & NaN / NaN & 4.17 / 2.15 & \textbf{4.10} / \textbf{2.06} & 5.92 / 2.76 & 23.44 / 7.56 & NaN / NaN & 38.24 / 14.15 & 14.42 / 7.86\\ 
           \hline 
           HalfTank Easy & \textbf{0.18} / \textbf{0.13} & 1.12 / 0.86 & 4.51 / 3.44 & 2.21 / 1.76 & 26.93 / 19.67 & 17.46 / 15.20 & \textbf{1.84} / \textbf{0.65} & 19.83 / 13.66 & 3.46 / 1.65 & 48.61 / 38.08 & 8.43 / 5.12 & 66.63 / 19.90\\ 
           \hline 
           HalfTank Medium & \textbf{0.24} / \textbf{0.16} & 0.26 / 0.17 & 2.90 / 2.07 & 0.71 / 0.42 & NaN / NaN & NaN / NaN & \textbf{3.17} / \textbf{1.64} & 5.93 / 3.51 & 24.15 / 13.89 & 15.16 / 7.77 & 45.99 / 21.16 & NaN / NaN\\ 
           \hline 
           HalfTank Hard & \textbf{0.24} / \textbf{0.16} & 0.37 / 0.26 & 68.70 / 47.50 & 1.15 / 0.75 & NaN / NaN & NaN / NaN & \textbf{3.45} / \textbf{1.94} & 9.85 / 6.36 & 15.65 / 8.41 & 22.24 / 15.57 & 92.07 / 48.77 & 65.16 / 37.67\\ 
           \hline 
           WholeTank Medium & \textbf{0.14} / \textbf{0.11} & 0.27 / 0.13 & 0.41 / 0.24 & 0.68 / 0.22 & 4.11 / 2.24 & 2.17 / 1.25 & \textbf{1.64} / \textbf{0.87} & 9.20 / 4.88 & 6.58 / 3.25 & 8.19 / 3.22 & 91.50 / 45.44 & 3.36 / 1.38\\ 
           \hline 
           WholeTank Hard & \textbf{0.12} / \textbf{0.07} & 0.30 / 0.17 & 0.27 / 0.21 & 2.37 / 1.95 & NaN / NaN & 0.95 / 0.76 & \textbf{2.83} / \textbf{1.21} & 10.70 / 6.87 & 3.80 / 1.90 & 30.50 / 23.99 & 28.58 / 11.13 & 4.54 / 2.57\\ 
           \hline   
        \end{tabular}
        \end{center}
    \end{table*}

% The proposed method exhibits the best performance on most sequences and shows the best robustness. The prior work\cite{xu2021underwater} exhibits slightly lower performance but demonstrates superior robustness compared to the proposed method. This is because the prior work does not consider bias initialization and merely assumes zero bias, which introduces additional error into the results.

% The performance of SVIN2 \cite{rahman2022svin2}, ORB3 \cite{orbslam3}, and VINS \cite{lin2018VINS-Mono} demonstrates substantial variability in accuracy across different sequences, and this inconsistency is further evident when observing different runs of the same sequence. Because visual-inertial SLAM systems face a challenge due to the low signal-to-noise ratio of the accelerometer. The acceleration measured by the accelerometer needs to be integrated twice to obtain translation information, which tends to magnify any existing noise. Over time, these errors can accumulate significantly. Consequently, visual-inertial systems often rely heavily on visual estimation to bound translational error.

% If the image quality degrades, even for a brief period, the accruing errors can rapidly compound, leading to substantial inaccuracies in translational estimation. This issue is particularly pronounced in underwater environments, where limited visibility often results in low-quality images, making these visual-inertial systems less reliable.

In contrast, our method incorporates the use of a DVL to enhance the accuracy of translational estimation. This approach significantly improves both performance and robustness, offering a notable advantage over the visual-inertial systems, especially in underwater environments where image quality is usually compromised.

\subsubsection{Qualitative Evaluation on HalfTank Easy Sequence}

Fig. \ref{fig:dense_odom_he} shows the trajectories estimated on the HalfTank Easy sequence by all the methods. This sequence encompasses challenging scenarios in which images lack distinct features. %, although the motion of the vehicle remains predominantly smooth.
As a consequence, this sequence presents a decent level of difficulty. In this sequence, the vehicle initially traverses the front side of the structure where GT is available. Subsequently, it moves through half of the tank along its boundary where GT is unavailable. The vehicle finally returns to the front side of the structure where GT becomes accessible again. % In this illustration, spheres are used to represent the starting points of trajectories, while boxes are employed to denote the endpoints of these trajectories.

The results indicate that our proposed method achieves the lowest level of drift. Our previous work exhibits the second-lowest drift. % Both these methods display only minor errors when compared to the GT trajectory, highlighting their relative accuracy and effectiveness.
SVIN2, VINS, and Basalt exhibit significantly higher drift than the proposed method, primarily due to the rapid accumulation of translation errors under degraded visual conditions. In contrast, ORB3 loses tracking until the robot traverses a significant portion of the trajectory and returns to the structure where it successfully initializes again.
% ORB3\cite{orbslam3} only save keyframes in current active map therefore the trajectory of ORB3 begins from the latter part of the GT segment, rather than from the start. These results highlight the distinct limitation of visual-inertial SLAM methods under these specific conditions.

In each sub-figure of Fig. \ref{fig:dense_odom_he}, parts (a) and (b) provide detailed views of the reconstructed map sections. The presence of misalignment or deformation in these reconstructions is indicative of the errors that have accumulated over the course of traversing the entire trajectory.
%  These visual discrepancies serve as a clear illustration of the extent and impact of error accumulation in the reconstruction process.
Our proposed method achieves the most accurate and comprehensive dense reconstruction of the structure and the tank's boundary.
% As shown in Fig. \ref{fig:dense_odom_he}(1)(a), the reconstruction of the proposed method shows slight misalignment, primarily in translation.
% In contrast, our previous work exhibits substantial misalignment throughout the entire structure as shown in Fig. \ref{fig:dense_odom_he} (2) (a).

% The reconstructed structure by SVIN2 \cite{rahman2022svin2} exhibits significant misalignment as evidenced in two distinct reconstructions depicted in Fig. \ref{fig:dense_odom_he} (3) (a1) and Fig. \ref{fig:dense_odom_he} (3) (a2).
% This divergence is predominantly attributed to considerable translational drift. Additionally, ORB3 \cite{orbslam3} experiences a rapid loss of tracking when the ROV navigates towards the wave-generator area, leading to an incomplete reconstruction. The reconstruction generated by VINS Fusion \cite{lin2018VINS-Mono} is notably inaccurate, particularly in the beach area as delineated in Fig. \ref{fig:dense_odom_he} (5) (b). This inaccuracy is further compounded by the drift, resulting in duplicate reconstructions of the same structure, as illustrated in Fig. \ref{fig:dense_odom_he} (5) (a1) and (a2), despite representing a singular entity in reality. These observed trajectories and reconstructions underscore the significant challenges visual-inertial systems encounter in preserving accuracy within complex underwater environments.

Fig. \ref{fig:error_time_odom_he} shows the evolution of errors over time across the 6-DoF poses. The proposed method consistently achieves high accuracy across all axes. 
Visual-inertial methods, including SVIN2, ORB-SLAM3, VINS, and Basalt, demonstrate comparable performance in roll and pitch but exhibit noticeable drift on other axes due to the challenging visual conditions.
Our previous work \cite{xu2021underwater} exhibited significant drift on orientation, contributing to overall trajectory drift. We hypothesize that this was primarily due to the exclusion of accelerometer integration and the assumption of zero gyroscope bias. To further demonstrate the effects of bias correction and accelerometer integration, a detailed ablation study is presented in Section \ref{sec:ablation_acc_bias}.

    % SVIN2\cite{rahman2022svin2}, ORB3 \cite{orbslam3} and VINS Fusion\cite{lin2018VINS-Mono} exhibit significant drift. This issue is largely attributed to the challenging image quality encountered during operation. The poor image quality impairs these systems' ability to effectively constrain the error. As a result, there is a rapid and significant drift, underscoring a critical limitation in these systems' performance in environments with suboptimal visual conditions.

    \subsubsection{Qualitative Evaluation on WholeTank Hard Sequence}

Fig. \ref{fig:dense_odom_wh} presents the trajectories on the WholeTank Hard sequence. This sequence involves a longer traversal distance and more aggressive vehicular motion, apart from challenging visual conditions. Initially, the vehicle circumnavigates the structure where GT is accessible, and it subsequently traverses the entire tank by following its boundary. The vehicle finally returns to the structure where GT becomes available again.

The trajectories depicted in Fig. \ref{fig:dense_odom_wh} demonstrate that our proposed method exhibits the least drift after navigating the entire tank. All other methods show noticeable drifts. The drift in our previous work is primarily along the z-axis, while SVIN2 and Basalt experience mainly translational drift. ORB3 and VINS demonstrate substantially higher overall drift compared to the other methods. % The performance of these methods can be also seen from the densely reconstructed maps in Fig. \ref{fig:dense_odom_wh}.

\begin{figure*}
    \addtocounter{figure}{+1}
    \subfigure{\includegraphics[width=2.05\columnwidth]{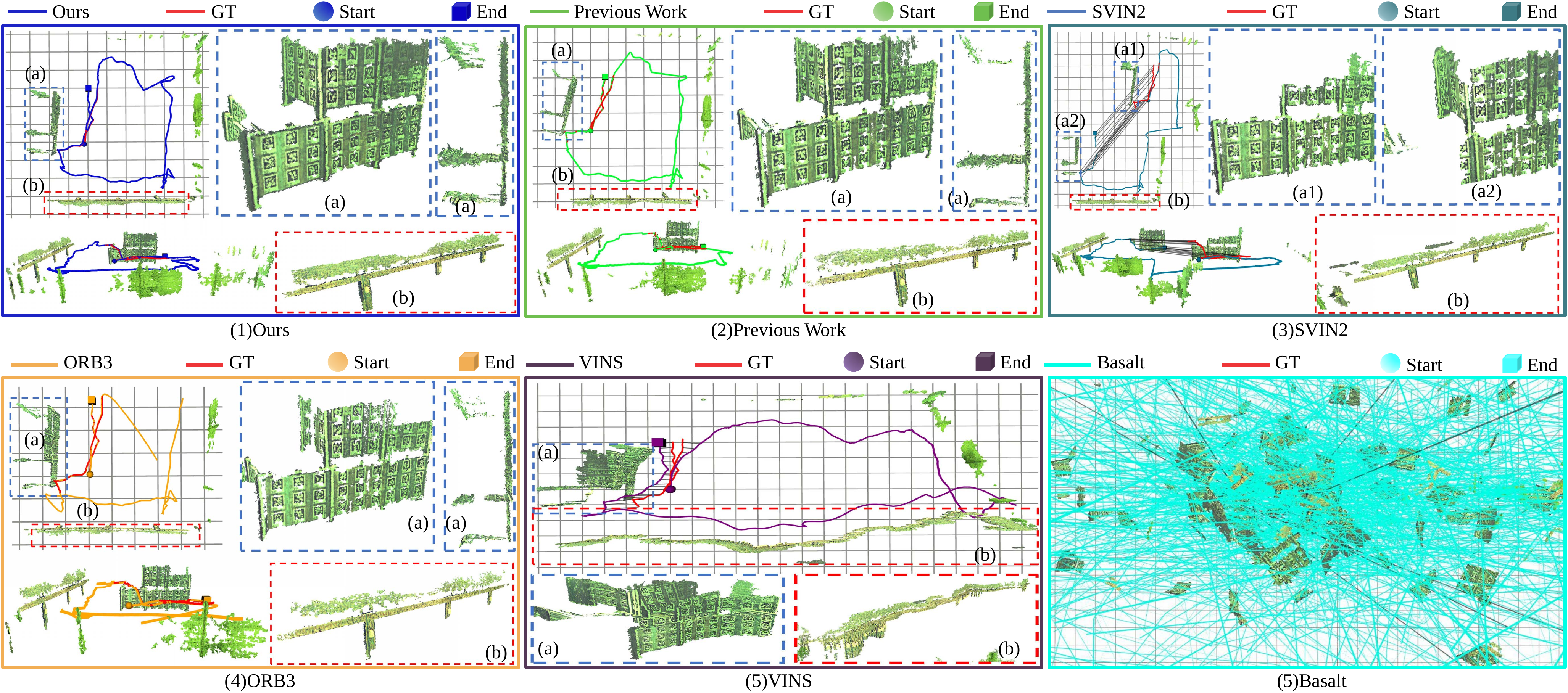}}
    \caption{Estimated trajectories and dense maps on HalfTank Easy sequence in SLAM mode. The figure legends and grid size are the same as Fig. \ref{fig:dense_odom_he}.}
    \label{fig:dense_slam_he}
\end{figure*}

\begin{figure}
    \centering
    \addtocounter{figure}{-2}
    \subfigure[SLAM translation error]{
    \includegraphics[width=1\columnwidth]{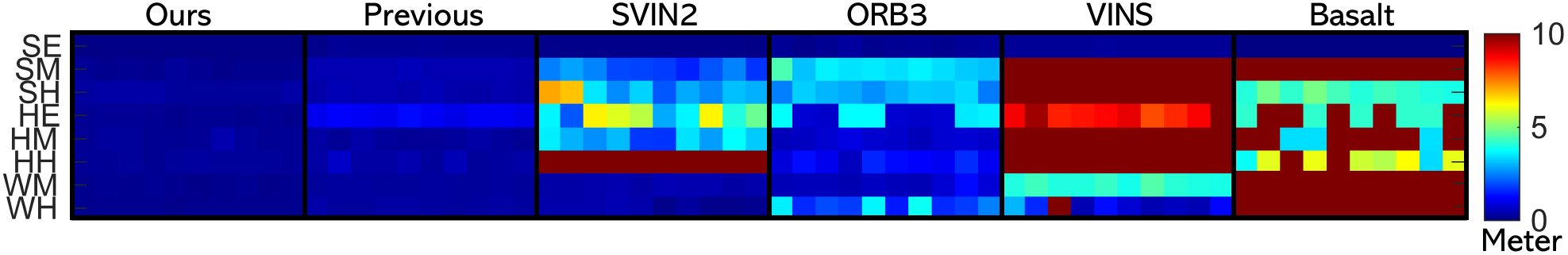}}
    \subfigure[SLAM rotation error]{
    \includegraphics[width=1\columnwidth]{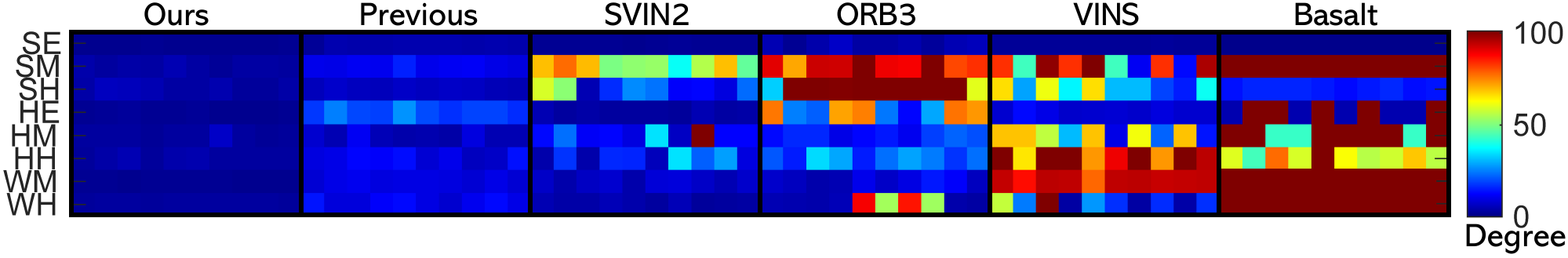}}
    \caption{SLAM results of 10 runs on all WaveTank sequences.}
    \label{fig:slam_error_map}
    \addtocounter{figure}{+1}
\end{figure}

\subsection{Evaluation on SLAM}
For the SLAM evaluation, the loop closure functionality of each method is activated.
% \Shida{*************I need to re-run the VINS results to get the loop results*************}
% \Shida{loop closure parameters of ORB3 are not tuned for wavetank dataset, so ORB3 sometimes cannot find loop or find loop more slowly than proposed method.}

    \subsubsection{Quantitative Evaluation}

\begin{figure*}
\addtocounter{figure}{+1}
    \subfigure{\includegraphics[width=2.05\columnwidth]{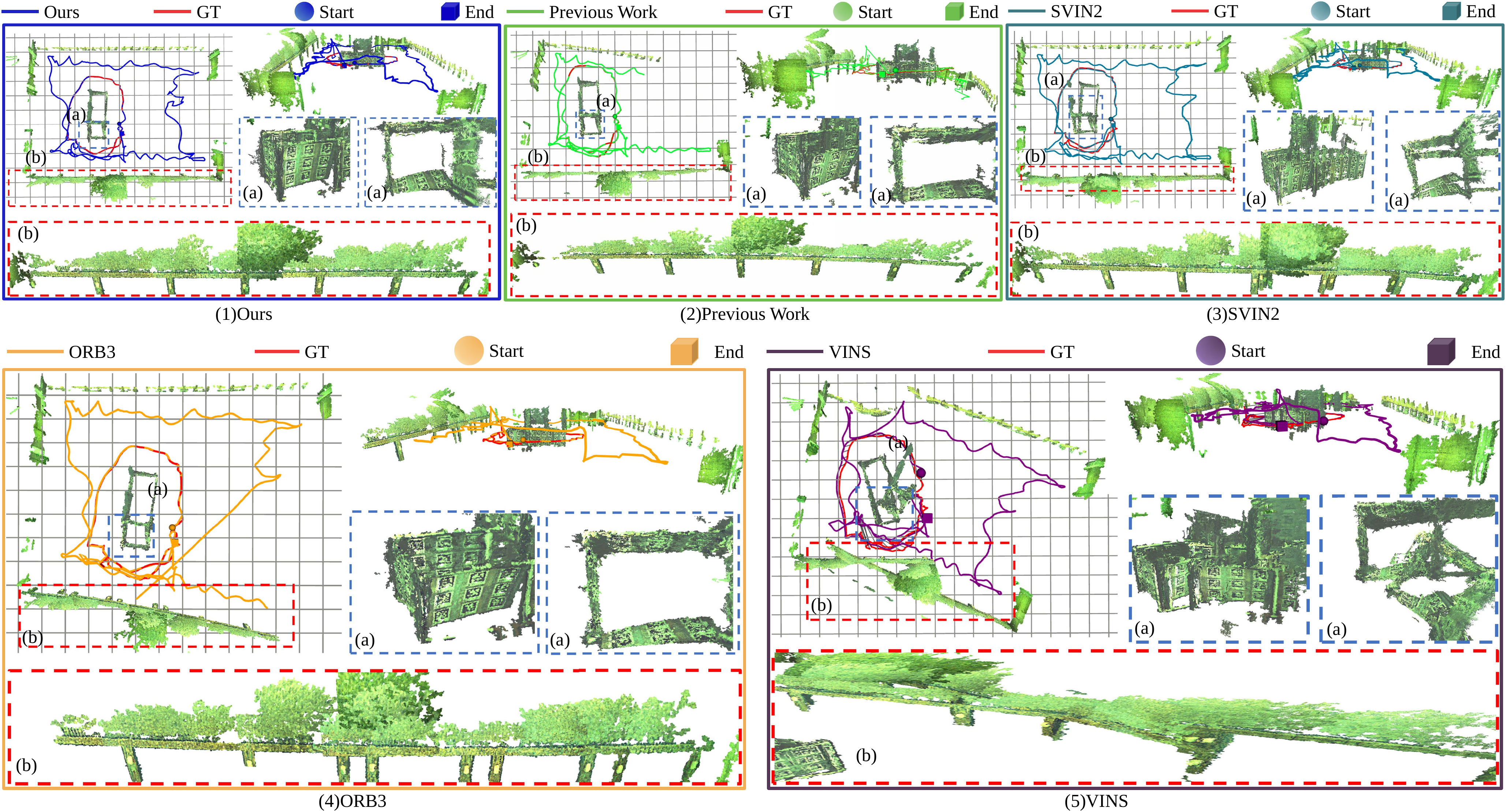}}
    \caption{Estimated trajectories and dense maps on WholeTank Hard sequence in SLAM mode. The figure legends and grid size are the same as Fig. \ref{fig:dense_odom_he}.}
    \label{fig:dense_slam_wh}
\end{figure*}

\begin{figure}
    \centering
    \addtocounter{figure}{-2}
    \subfigure{\includegraphics[width=\columnwidth]{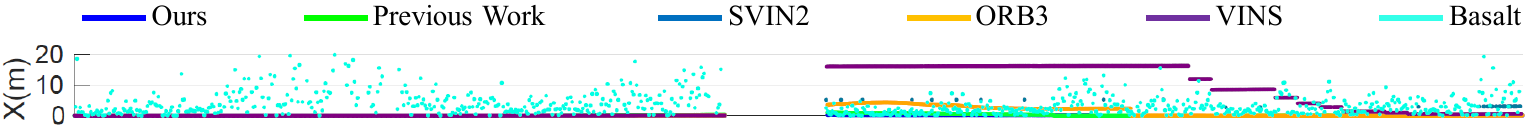}}
    \subfigure{\includegraphics[width=\columnwidth]{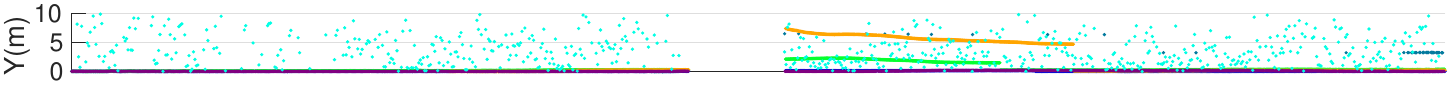}}
    \subfigure{\includegraphics[width=\columnwidth]{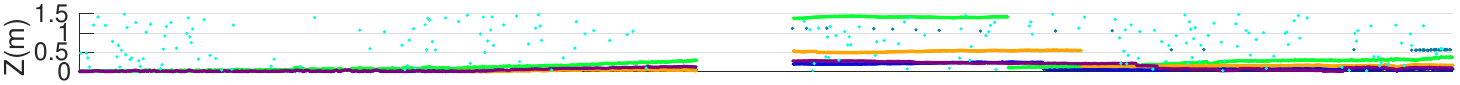}}
    \subfigure{\includegraphics[width=\columnwidth]{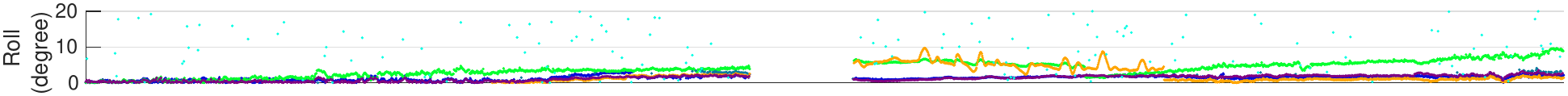}}
    \subfigure{\includegraphics[width=\columnwidth]{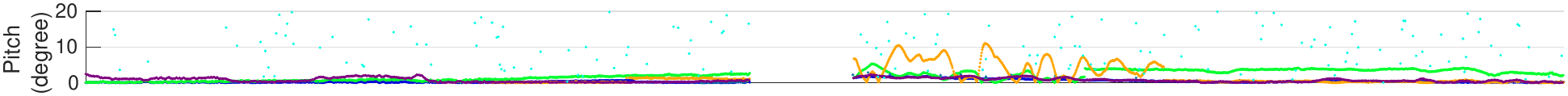}}
    \subfigure{\includegraphics[width=\columnwidth]{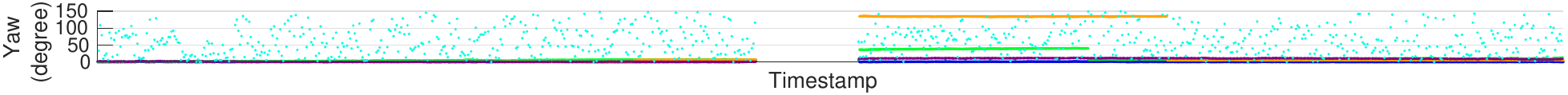}}
    \caption{SLAM errors of 6 axis on HalfTank Easy sequence. The gaps indicate the times without GT poses.}
    \label{fig:error_time_slam_he}
    \addtocounter{figure}{+1}
\end{figure}

Similar to the odometry evaluation, for each sequence, a method is run 10 times to compute the average RMSE and STD in Table \ref{table:slam_rmse}. The results of the 10 runs are also given in Fig. \ref{fig:slam_error_map}. We can see that our method surpasses others in most scenarios. Although the SLAM performances of SVIN2, ORB3 and VINS show improvement compared with their odometry, they still exhibit significant errors in translation, particularly in sequences characterized by challenging visual conditions. The loop closure mechanisms somewhat mitigate the drifts, but the lack of sufficient features or lighting significantly degrades the performance and reliability of visual-inertial systems in underwater settings. Basalt only achieves slightly better performance to our proposed method on the SE sequence, drifting on sequences with more challenging visual conditions. We hypothesize that the superior performance of Basalt on the SE sequence is due to its use of an offline loop closure approach, which allows for extensive frame matching and larger-scale optimization. In contrast, our approach employs an online loop correction in real-time and achieves comparable performance. Additionally, Basalt frequently encounters numerical faults, leading to system crashes when running on sequences with challenging visual conditions.

The temporal evolution of errors across the 6-DoF poses is given in Fig. \ref{fig:error_time_slam_he}.
% demonstrating a notable distinction from the odometrAs discussed before y results shown in Fig. \ref{fig:error_time_odom_he}.
Most methods, except SVIN2 and Basalt, exhibit a significant reduction in error following loop detection for enhanced trajectory accuracy.
Our proposed method consistently achieves superior accuracy both before and after loop closure. In comparison, our previous work realizes comparable accuracy in translation and yaw errors post loop closure. However, its errors in roll and pitch remain uncorrected.
SVIN2  does not detect the loop, and its loop-closure module, implemented as a separate ROS node from the tracking node, publishes poses at a low frequency. This results in only a few data points being represented.
After the loop closures, ORB3 approaches the accuracy of the proposed method in the translation and yaw axes. However, it exhibits significant errors in roll and pitch at the start of the second trajectory segment. As discussed before, ORB3 cannot localise the robot when traversing along the tank wall.
VINS exhibits a noticeable reduction in x-translation error and gradually corrects for drift, resulting in a stepwise decrease in error.
Basalt exhibits significant errors in trajectory due to false positive loop closure detections under challenging visual conditions.

\subsubsection{Qualitative Evaluation on HalfTank Easy Sequence}

% \begin{figure}
%     \centering
%     \subfigure{\includegraphics[width=\columnwidth]{images/error_timestamp_wh_slam/t_x-cropped}}\\
%     \subfigure{\includegraphics[width=\columnwidth]{images/error_timestamp_wh_slam/t_y-cropped}}\\
%     \subfigure{\includegraphics[width=\columnwidth]{images/error_timestamp_wh_slam/t_z-cropped}}\\
%     \subfigure{\includegraphics[width=\columnwidth]{images/error_timestamp_wh_slam/r_roll-cropped}}\\
%     \subfigure{\includegraphics[width=\columnwidth]{images/error_timestamp_wh_slam/r_pitch-cropped}}\\
%     \subfigure{\includegraphics[width=\columnwidth]{images/error_timestamp_wh_slam/r_yaw-cropped}}
%     \caption{Error over time of 6 axis on WholeTank Hard in odometry mode, from top to bottom: x, y, z, roll, pitch, yaw.}
%     \label{fig:error_time_slam_wh}
% \end{figure}

Fig. \ref{fig:dense_slam_he} illustrates the trajectories and dense reconstructions produced by the methods in SLAM mode for the HalfTank Easy sequence.
Our proposed method, alongside our previous work, estimates trajectories well aligned with GT, producing good-quality 3D maps.
% When we direct our attention to area (a) of the structure in Fig. \ref{fig:dense_slam_he} (1) and (2), and compare it with area (a) in , it becomes evident that
The map misalignments observed in the odometry mode in Fig. \ref{fig:dense_odom_he}(1) and (2) have also been corrected in the SLAM mode.
Meanwhile, ORB3 achieves significantly improved results compared to its odometry mode, yet it fails to localise around the tank wall. The results of SVIN2  and VINS  still show clear errors in the GT.
Basalt exhibits significant errors in the trajectory and dense reconstruction on the HE sequence. Furthermore, it consistently crashes on the WH sequence, preventing result acquisition. Therefore, the maximum error is manually set for the WH sequence. We hypothesize that this is caused by false positive loop closure detections due to poor image quality.

% SVIN2 \cite{rahman2022svin2} demonstrates clear misalignment in its trajectory when compared with the GT, as well as in the reconstruction of the structure, particularly in parts (a1) and (a2) of Fig. \ref{fig:dense_slam_he} (3). This misalignment is primarily attributed to translational drift. The resemblance of these results to those observed in odometry mode, as depicted in Fig. \ref{fig:dense_odom_he} (3), can be ascribed to SVIN2's \cite{rahman2022svin2} inability to detect loops in this specific sequence.

% ORB3 \cite{orbslam3} exhibits well-aligned trajectories and high-quality reconstruction, yet it fails to capture certain parts of the boundary due to tracking failures, as illustrated in Fig. \ref{fig:dense_slam_he} (4). When compared to its odometry results depicted in Fig. \ref{fig:dense_odom_he} (4), which primarily showcase the reconstruction of the structure, ORB3 presents a more comprehensive reconstruction of the tank in SLAM mode. This improvement is attributed to its ability to successfully merge back to the previous map upon revisiting the front side of the structure.

% The results from VINS Fusion\cite{lin2018VINS-Mono} make it difficult to discern the shape of the tank and the structure, as illustrated in Fig. \ref{fig:dense_slam_he} (5). This is primarily due to large drift caused by an unbounded bias, which remains uncorrected despite loop detection. However, the misalignment of the structure is much less severe than in odometry mode, as illustrated in Fig. \ref{fig:dense_odom_he} (5).

\subsubsection{Qualitative Evaluation on WholeTank Hard Sequence}

The quantitative SLAM results on WholeTank Hard sequence are also shown in Fig. \ref{fig:dense_slam_wh}, which provides similar insights into the SLAM performance. The absence of Basalt's results is due to its consistent crashes on this sequence.

\subsection{Validation in Real Offshore Environments}
The offshore experiments were conducted near an offshore wind turbine in the North Sea. The Falcon ROV, localized using the proposed SLAM method, was deployed to operate around the turbine foundation.
Due to the lack of ground truth data in open-sea environments, we use trajectory and reconstruction generated by COLMAP \cite{schoenberger2016sfm} as a reference to evaluate our performance. However, the challenging nature of the Offshore image sequences compromised the performance of COLMAP's Multi-View Stereo (MVS) component, resulting in noisy dense reconstructions. Consequently, only the Structure from Motion (SfM) generated trajectory from COLMAP was used as a reference. To assess reconstruction quality, point clouds obtained through stereo matching were incrementally fused using both the trajectory produced by our method and that of COLMAP's SfM.

\begin{figure}
    \centering
    \includegraphics[width=1.0\linewidth]{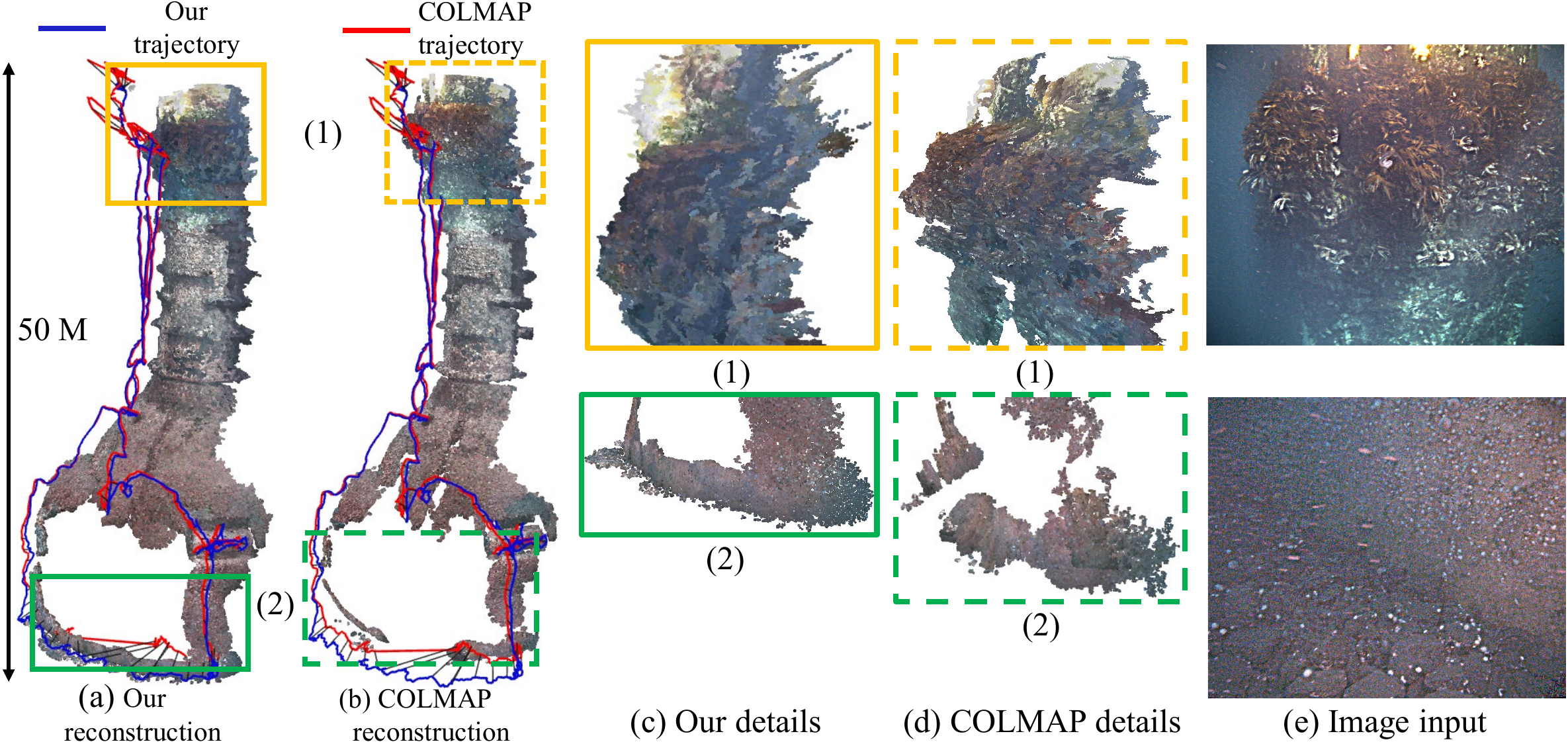}
    \caption{Offshore 3D reconstruction and estimated trajectory comparison with COLMAP.}
    \label{fig:offshore_large_compare}
    % \Sen{align images like Figure 1}
\end{figure}

Fig. \ref{fig:offshore_big} shows the dense point cloud covering an expansive region of about $50$-meter depth from the top to the bottom of a turbine base. Despite poor image quality in certain areas, the dense reconstruction remains consistent, verifying the accuracy of the poses estimated by the proposed SLAM technique and its viability in real offshore scenarios.
Additionally, we compare the reconstructions generated by our method and COLMAP in Fig. \ref{fig:offshore_large_compare}. The COLMAP trajectory exhibits misalignment with our trajectory in both the top and bottom areas. The top area, covered in algae, led to noisy results from COLMAP, evident in the inconsistent reconstruction shown in Fig. \ref{fig:offshore_large_compare} (1). The bottom area, containing motion blur and textureless images, also challenged COLMAP, resulting in a suboptimal reconstruction shown in Fig. \ref{fig:offshore_large_compare} (2). In contrast, our method's dense reconstruction is more consistent and accurate, even in these areas with poor image quality.

\begin{figure}
    \centering
    \includegraphics[width=1.0\linewidth]{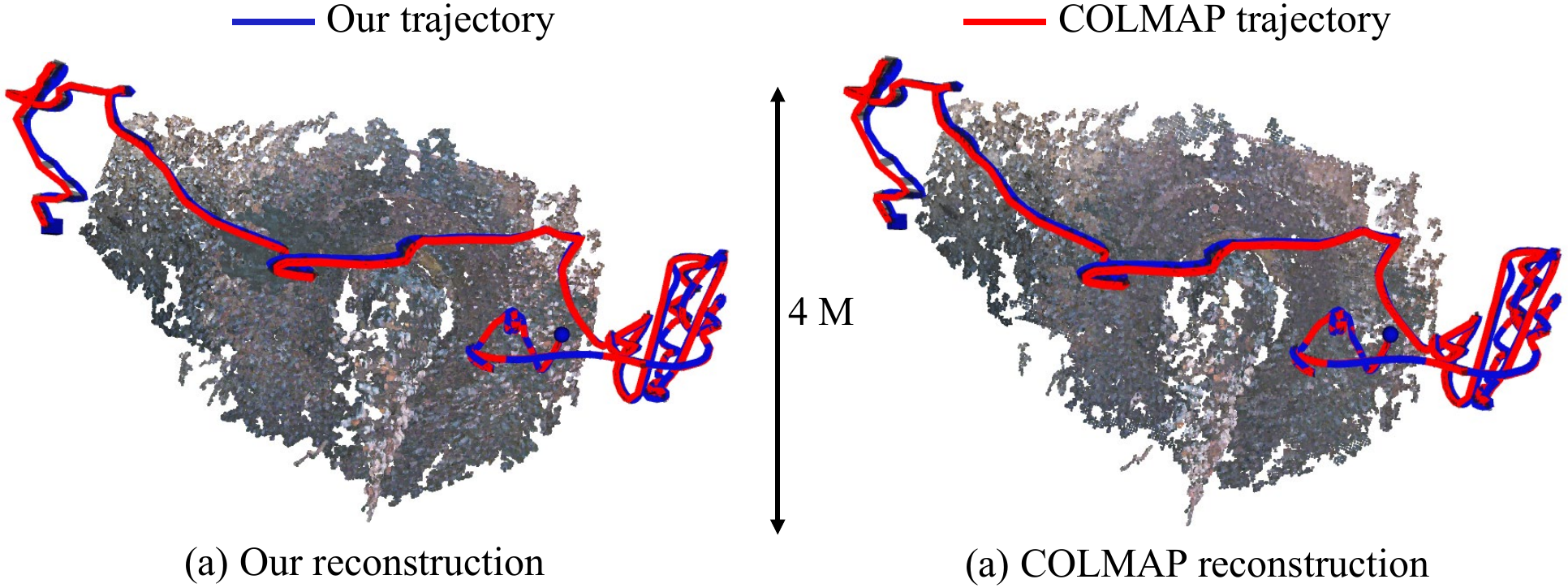}
    \caption{Offshore 3D reconstruction and estimated trajectory comparison with COLMAP.}
    \label{fig:offshore_medium}
    % \Sen{align images like Figure 1}
\end{figure}

Furthermore, we compared our results with COLMAP in the region surrounding a cable socket on the wind turbine base, as shown in Fig. \ref{fig:offshore_medium}. Both trajectories align well, and the dense reconstructions are consistent.
The offshore experiments demonstrate the robustness of the proposed method in challenging real-world ocean environments. % The dense reconstruction of the offshore structures is consistent and accurate, even in areas with poor image quality. This validates the effectiveness of the proposed method in real-world applications.

% \begin{figure}
%     \centering
%     \includegraphics[width=0.9\linewidth]{images/offshore/small_area.png}
%     \caption{Offshore validation at short-range area}
%     \label{fig:offshore_small}
% \end{figure}
% Fig. \ref{fig:offshore_small} displays the dense reconstruction of a limited-range area adjacent to the cable slot connecting to the wind turbine. It presents a consistent structure, wherein the shape of the cable slot can be readily identified.

\subsection{Evaluation on Sensor Calibration}
%    \Shida{Add simulation validation of the convergecy of extrinsic parameter, bias, gravity direction, DVL transducer orientation}

\subsubsection{Extrinsic Calibration of DVL, IMU and Camera}
A segment of the Easy Structure sequence which has relatively clear and richly featured images is chosen for the extrinsic calibration among the DVL, IMU and camera sensors. An identity matrix is set as the initial extrinsic transformation for the calibration optimization, assuming no prior information about the calibration parameters. Then the extrinsic parameters between the IMU and the DVL, $\mathbf{T}_{\mathtt{I}\mathtt{D}}$, and between the camera and the DVL, $\mathbf{T}_{\mathtt{D}\mathtt{C}}$, are optimized. The calibration process commences after the insertion of 10 keyframes and ends when 100 keyframes have been received.

\begin{figure}
    \renewcommand{\thesubfigure}{(\arabic{subfigure})} % Temporarily change the counter representation
    \centering
    \subfigure[Extrinsic parameters $\mathbf{T}_{\mathtt{I}\mathtt{D}}$]{
	\includegraphics[width=0.48\columnwidth]{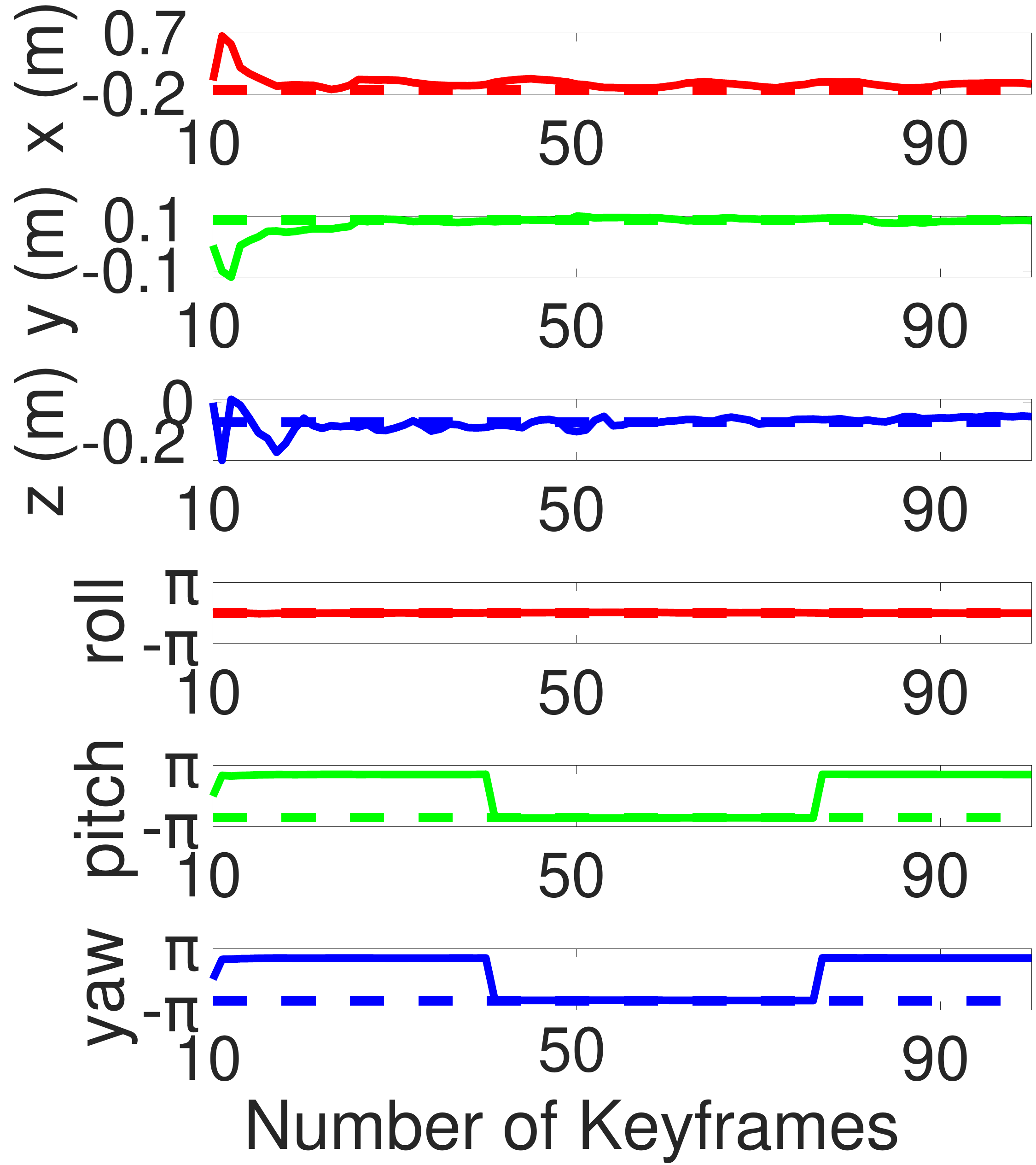}}
    \subfigure[Extrinsic parameters $\mathbf{T}_{\mathtt{D}\mathtt{C}}$]{
    \includegraphics[width=0.48\columnwidth]{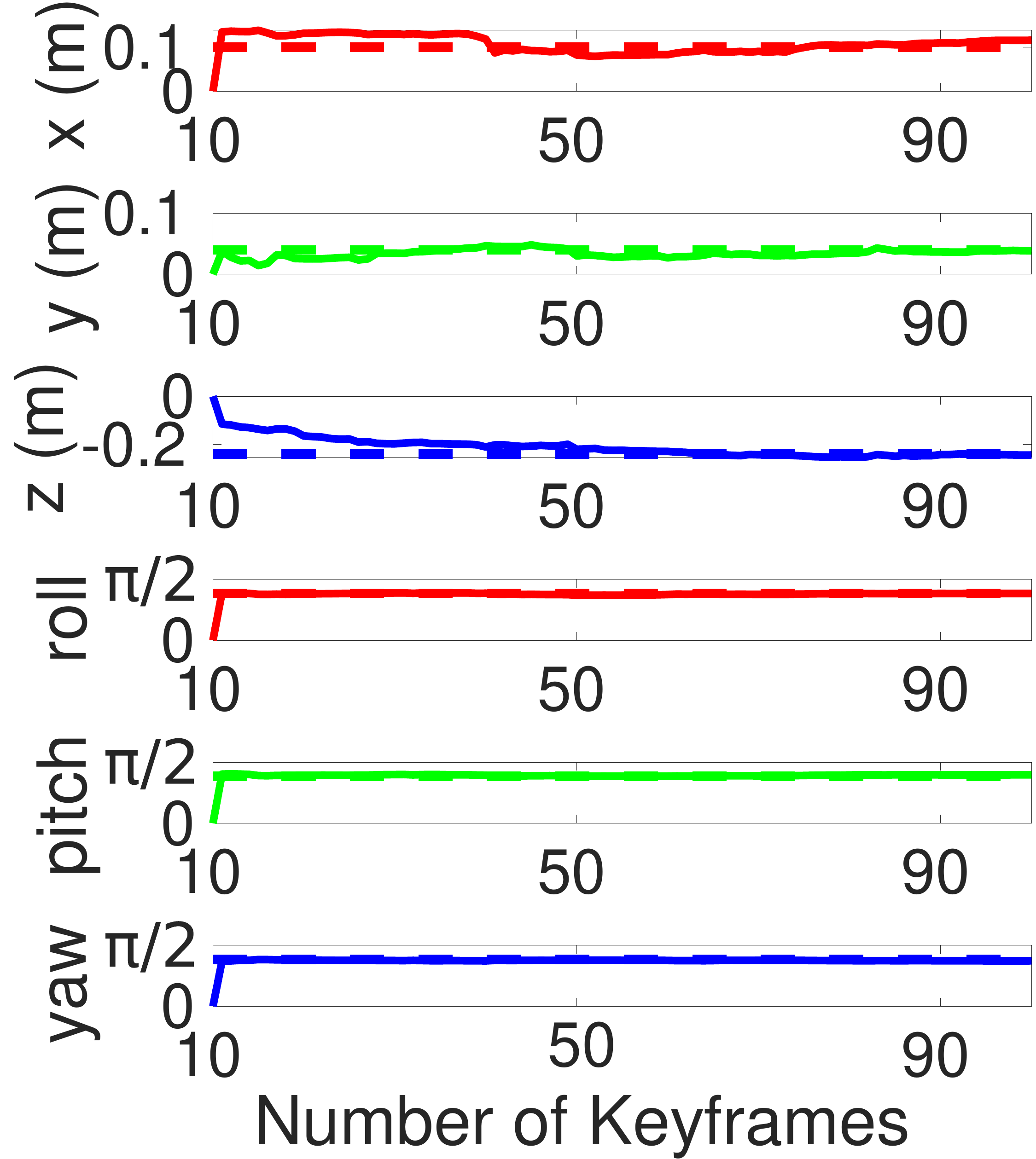}}
    \caption{Results of extrinsic calibration. Solid lines: estimation. Dashed lines: manual measure.}
%    \Sen{Draw parameters initially starting from identity. add legend solid and dashed lines. cut white space and make it more compact}
    \label{fig:extrinsic_calibration}
    \renewcommand{\thesubfigure}{\thefigure(\alph{subfigure})} % Change it back to the original format
\end{figure}

The extrinsic calibration result is shown in Fig. \ref{fig:extrinsic_calibration}. %The dash lines present the manually measured parameter values.
We can see the estimated extrinsic parameters gradually converge to the measured values. Note the slight offsets between the calibrated parameters and the manual measures are likely caused by the inevitable measurement errors, particularly for the orientations.
Nonetheless, the results demonstrate that the proposed method can achieve decent calibration even without prior information.

% are the fact manually measuring the orientation between sensors due to , so the rough measurement of the orientation might not be precise. Nonetheless, the results demonstrate that the proposed method can deliver decent calibration even without prior information.

\subsubsection{DVL Misalignment Calibration}

\begin{figure}
    \centering
    \includegraphics[width=1\linewidth]{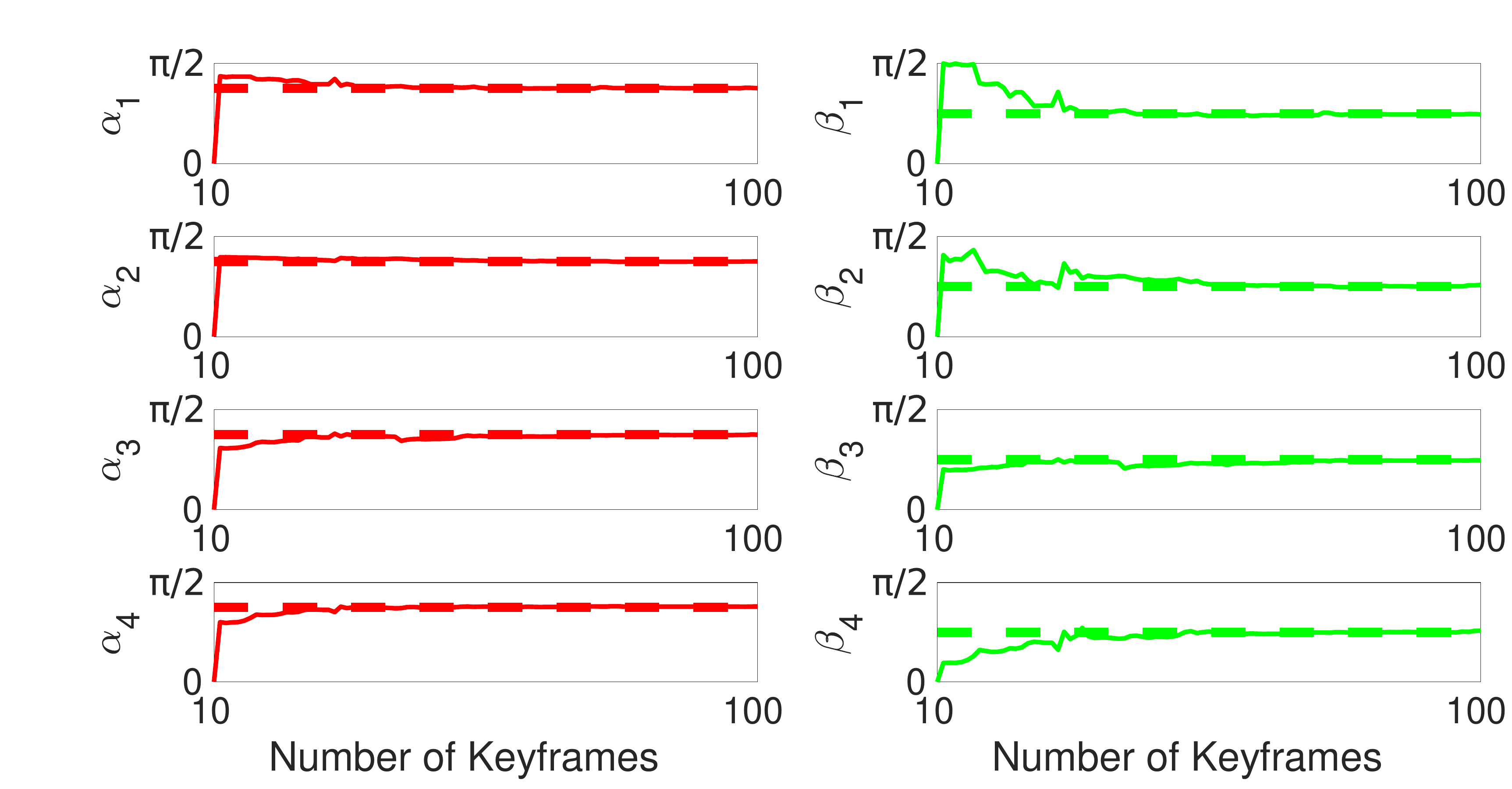}
    \caption{DVL misalignment calibration.  Solid lines: estimation. Dashed lines: manual measure.}
%    \Sen{add legend solid and dashed lines. cut white space and make it more compact}
    \label{fig:dvl_calibration}
\end{figure}

The DVL misalignment calibration is also performed on the Easy Structure sequence. The results are given in Fig. \ref{fig:dvl_calibration}.
% The dash line present the the default setting of each transducer.
Over time, we can observe the calibrated transducer alignment parameters converging to the default setting.
Since our DVL is rather new and well-maintained with no substantial misalignment, the calibration results closely match the default setting. These results demonstrate the capability of the proposed method to calibrate the transducer alignment parameters accurately.

\subsection{Ablation Study}

\subsubsection{Effect of Bias Correction and Accelerometer Residual}
\label{sec:ablation_acc_bias}

\begin{figure}
    \centering
    \includegraphics[width=\columnwidth]{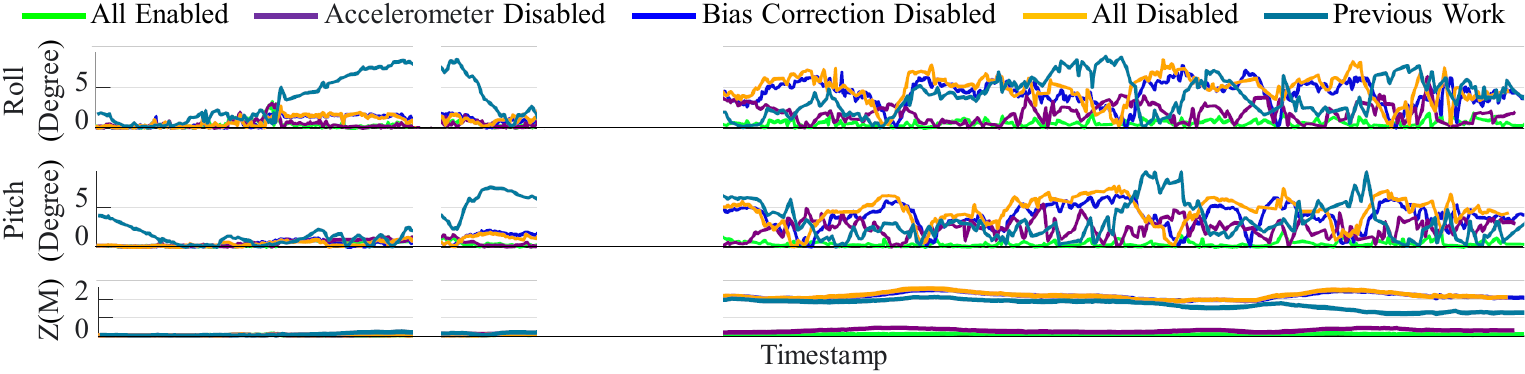}
    \caption{Abalation study of the effect of bias correction and accelerometer integration. The gaps indicate the times without GT poses.}
    \label{fig:ablation_bias_acc}
\end{figure}
 In previous sections, we hypothesized that the drift observed in our prior work was primarily due to the lack of accelerometer integration and the assumption of zero gyroscope bias. To further substantiate this hypothesis and analyze the factors contributing to the performance improvement of our proposed method over the previous work, we present a detailed ablation study in this section.

To evaluate their impact, we disabled the accelerometer-related residuals (defined in Equations \refeq{equa:imu_velocity_residual} and \refeq{equa:imu_translation_residual}) and bias correction in the optimization process. The results, illustrated in Fig. \ref{fig:ablation_bias_acc}, reveal increased drift in roll, pitch, and z axes when either accelerometer residuals or bias correction is disabled individually. The highest error occurs when both are disabled simultaneously. This is because the IMU residuals constrain roll and pitch through gravity, and gyroscope bias correction further reduces noise in gyroscope measurements. Disabling these components introduces errors in roll and pitch estimation, and since the BlueROV2 moves primarily in the y direction, roll drift directly translates into z-axis drift. Furthermore, we observe similar performance to our previous work when disabling both the accelerometer residuals and bias correction. This validates that the performance improvements compared to our previous work stem from: 1) the newly introduced tightly-coupled formulation, which enables bias correction, and 2) the newly added accelerometer residuals.

\subsubsection{Effect of Sensor Calibration}

To further investigate the impact of sensor calibration, an ablation study is performed in the UUV simulation environment \cite{Manhaes2016_uuvsim}. Noises to the extrinsic parameters and the DVL transducer orientation are manually added. Then, using the proposed calibration method, these parameters are recalibrated automatically. Subsequently, we compared the SLAM performance using both the calibrated and uncalibrated parameters, on three sequences. The results, presented in Tables \ref{table:ablation_sensor_calibration}, demonstrate that the calibrated parameters significantly reduce translation and rotation errors compared to the uncalibrated ones. These findings underscore the importance and effectiveness of the proposed sensor calibration methods.

\begin{table}
    \caption{Ablation study of the effect of sensor calibration}
    \label{table:ablation_sensor_calibration}
    \centering
    \begin{adjustbox}{width=1\columnwidth}
        \footnotesize \setlength{\tabcolsep}{2.7pt}
        \begin{tabular}{l || c | c | c ||  c | c | c }
            \hline
            & \multicolumn{3}{c||}{Translation Error (in meter)} & \multicolumn{3}{c}{Rotation Error (in degree)} \\
            \cline{2-7}
            & \makecell[c]{All \\ Calibrated}  & \makecell[c]{Extrinsics \\ Uncalibrated} & \makecell[c]{DVL \\ Uncalibrated} & \makecell[c]{All \\ Calibrated}  & \makecell[c]{Extrinsics \\ Uncalibrated} & \makecell[c]{DVL \\ Uncalibrated} \\
            \hline
            Seq 1 & \textbf{0.308} & 1.421 & 1.082 & \textbf{1.438} & 4.030 & 4.065 \\
            \hline
            Seq 2 & \textbf{0.039} & 0.167 & 0.300 & \textbf{0.544} & 2.136 & 0.998\\
            \hline
            Seq 3 & \textbf{0.070} & 1.118 & 2.187 & \textbf{0.427} & 11.675 & 13.511\\
            \hline
        \end{tabular}
    \end{adjustbox}
\end{table}

\subsubsection{Challenging Cases}

Thanks to the integration of DVL, our SLAM system can perform robustly in visually challenging environments over extended periods. During testing on the datasets, no failure cases were observed. However, for improved precision, we recommend that the SLAM system is initialized in visually favorable conditions and undergo sufficient motion to complete IMU initialization, which corrects bias and estimates the gravity direction. Insufficiently excited IMU initialization may lead to suboptimal precision, especially when visual conditions are poor and the system relies primarily on IMU and DVL for pose estimation.

\begin{figure}
    \includegraphics[width=1\columnwidth]{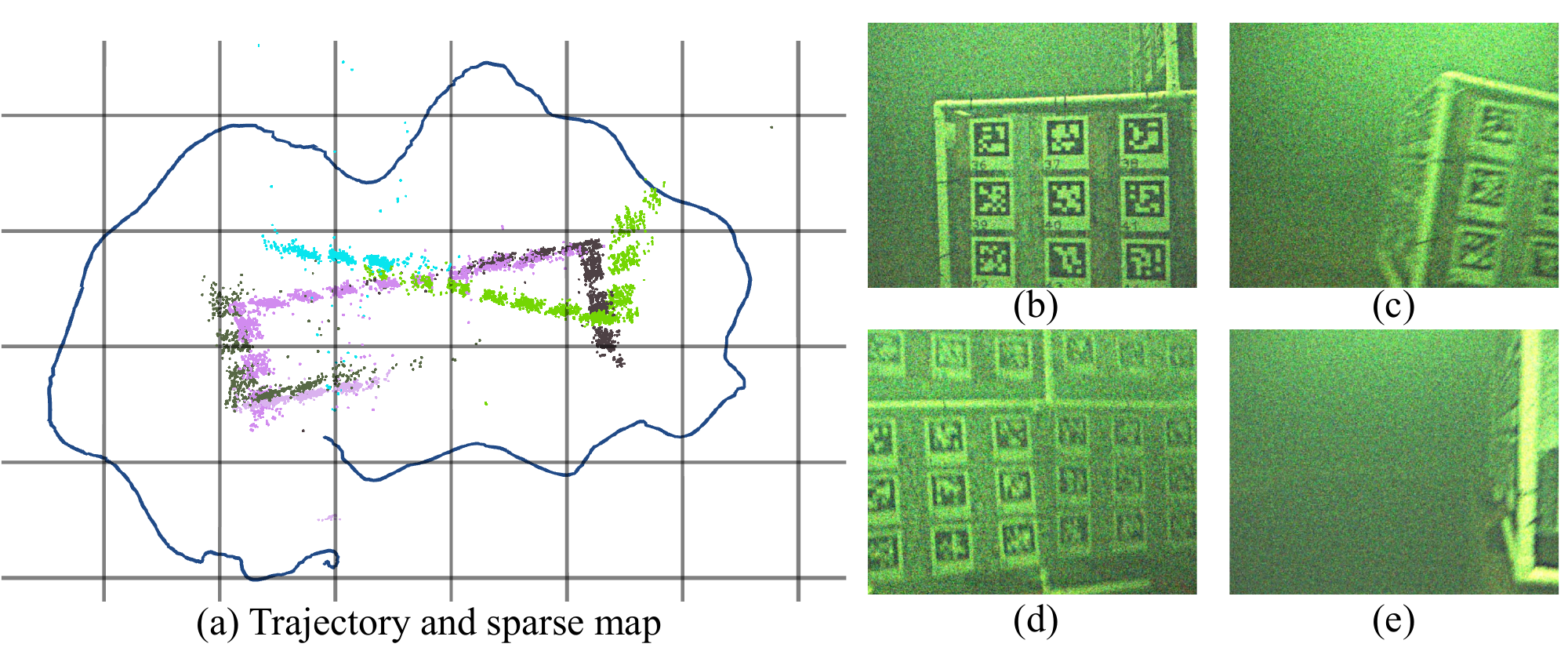}
    \caption{Trajectory and sparse reconstruction under challenging visual conditions but without proper initialization.}
    \label{fig:challenging_case}
\end{figure}

Fig. \ref{fig:challenging_case} (a) shows a challenging scenario where the SLAM system was initialized under poor visual conditions, resulting in a distorted sparse reconstruction. Three factors contribute to this distortion: (1) The image quality during initialization was poor, as illustrated in Fig. \ref{fig:challenging_case} (b) and (c); (2) initialization occurred during an aggressive motion, causing severe motion blur in the images (Fig. \ref{fig:challenging_case} (c)), which led to IMU initialization failure and default zero bias; (3) aggressive motion, low image quality, and persistent motion blur throughout the sequence forced the SLAM system to mainly rely on IMU and DVL for pose estimation. Since the IMU bias was not properly initialized, drift, especially in orientation, was introduced. These factors together caused pose drift and misalignment in the sparse map, as shown in Fig. \ref{fig:challenging_case} (a).

Therefore, we recommend initializing the SLAM system in visually optimal conditions with sufficient motion to ensure proper IMU initialization, which corrects bias and estimates gravity direction. Without proper IMU initialization, precision may degrade, particularly in challenging visual environments.

% \fi
\section{Conclusions}\label{sec:cons}

% \begin{table}
%     \begin{minipage}{0.5\textwidth} % Adjust width as needed
%     \begin{center}
%     \caption{Abalation study of the effect of calibration on translation error}
%     \begin{tabular}{l || c | c | c }
%         \hline
%         & \multicolumn{3}{c}{Translation Error (in meter)} \\
%         \cline{2-4}
%            & \makecell[c]{All \\ Calibrated}  & \makecell[c]{Extrinsics \\ Uncalibrated} & \makecell[c]{DVL \\ Uncalibrated} \\
%             \hline 
%             Seq 1 & \textbf{0.308} & 1.421 & 1.082 \\ 
%             \hline 
%             Seq 2 & \textbf{0.039} & 0.167 & 0.300 \\ 
%             \hline 
%             Seq 3 & \textbf{0.070} & 1.118 & 2.187 \\ 
%             \hline 
%     \end{tabular}
%     \end{center}
%     \end{minipage}\hfill 
%     \begin{minipage}{0.5\textwidth} % Adjust width as needed
%     \centering
%     \caption{Abalation study of effect of calibration on rotation error}
%     \begin{tabular}{l || c | c | c  }
%         \hline
%         & \multicolumn{3}{c}{Rotation Error (in degree)} \\
%         \cline{2-4}
%            & \makecell[c]{All \\ Calibrated}  & \makecell[c]{Extrinsics \\ Uncalibrated} & \makecell[c]{DVL \\ Uncalibrated} \\
%             \hline 
%             Seq 1 & \textbf{1.438} & 4.030 & 4.065\\ 
%             \hline 
%             Seq 2 & \textbf{0.544} & 2.136 & 0.998\\ 
%             \hline 
%             Seq 3 & \textbf{0.427} & 11.675 & 13.511\\ 
%             \hline 
%         \end{tabular}
%     \end{minipage}
% \end{table}

This paper proposes a novel underwater acoustic-visual-inertial SLAM which tightly fuses DVL, camera and IMU sensors in a graph optimization framework. DVL measurement is rigorously investigated, and its DVL pre-integration model is derived in detail. The proposed SLAM method leverages DVL sensing to enhance its accuracy and robustness in challenging underwater environments.
Meanwhile, novel techniques for DVL-camera-IMU extrinsic calibration and DVL transducer misalignment calibration are presented, further expedited with a rapid linear approximation method. These methods enable precise calibration of the extrinsic parameters among the DVL, camera, and IMU, as well as the orientation of the DVL transducer, even without the need for a dedicated underwater facility.
The proposed methods are evaluated qualitatively and qualitatively in a tank and at an offshore site. The results demonstrate that the proposed method not only achieves superior accuracy and robustness compared to state-of-the-art underwater SLAM methods and visual-inertial SLAM methods, but also has the capability to enable real-world underwater applications. These applications include precise navigation and mapping for deep-sea exploration, accurate localization for the maintenance and inspection of under- water pipelines and cables, detailed monitoring of coral reef health and other marine ecosystems, and efficient search and recovery operations in underwater archaeology and wreck exploration. In future, we will investigate temporal calibration for acoustic-visual-inertial systems.
% \clearpage
\section*{Appendix}
\subsection{DVL Translation Measurement Model Derivation}
\label{sec:appendix DVL Translation Measurement Model Derivation}
    Given the definitions of ${_{\mathtt{C}_0}\mathbf{p}_{{\mathtt{C}_0}\mathtt{D}_j}}$,${_{\mathtt{C}_0}\mathbf{p}_{{\mathtt{C}_0}\mathtt{D}_i}}$ and ${_{\mathtt{C}_0}\mathbf{p}_{\mathtt{D}_i\mathtt{D}_j}}$ in \eqref{equa:dvl_intg1}, we can reformulate \eqref{equa:dvl_intg1} as follows:
    {\allowdisplaybreaks
        \begin{align}
            &\mathbf{p}_{j} - \mathbf{R}_{j} \mathbf{R}_{\mathtt{D}\mathtt{C}}^T {_\mathtt{D}\mathbf{p}_{\mathtt{D}\mathtt{C}}}   =  \mathbf{p}_{i}   - \mathbf{R}_{i} \mathbf{R}_{\mathtt{D}\mathtt{C}}^T {_\mathtt{D}\mathbf{p}_{\mathtt{D}\mathtt{C}}} \\ & +  \sum_{k=i}^{j-1} \mathbf{R}_{i} \mathbf{R}_{\mathtt{I}\mathtt{C}}^T {\Delta\mathbf{R}_{\mathtt{I}_i\mathtt{I}_k}} \mathbf{R}_{\mathtt{I}\mathtt{D}}\ {(_{\mathtt{D}_i}{\tilde{\mathbf{v}}}} - \bm{\eta}^D) \Delta t \\ \xRightarrow[]{\text{(a)}} &  \mathbf{p}_{j} - \mathbf{R}_{j} \mathbf{R}_{\mathtt{D}\mathtt{C}}^T {_\mathtt{D}\mathbf{p}_{\mathtt{D}\mathtt{C}}} - \mathbf{p}_{i} + \mathbf{R}_{i} \mathbf{R}_{\mathtt{D}\mathtt{C}}^T {_\mathtt{D}\mathbf{p}_{\mathtt{D}\mathtt{C}}} \\ & =   \sum_{k=i}^{j-1} \mathbf{R}_{i} \mathbf{R}_{\mathtt{I}\mathtt{C}}^T {\Delta\mathbf{R}_{\mathtt{I}_i\mathtt{I}_k}} \mathbf{R}_{\mathtt{I}\mathtt{D}}\ {(_{\mathtt{D}_i}{\tilde{\mathbf{v}}}} - \bm{\eta}^D) \Delta t \\
            \xRightarrow[]{\text{(b)}} &{\mathbf{R}_{\mathtt{I}\mathtt{C}}} {\mathbf{R}_{i}}^T \mathbf{p}_{j} - {\mathbf{R}_{\mathtt{I}\mathtt{C}}} {\mathbf{R}_{i}}^T \mathbf{R}_{j} \mathbf{R}_{\mathtt{D}\mathtt{C}}^T {_\mathtt{D}\mathbf{p}_{\mathtt{D}\mathtt{C}}} - {\mathbf{R}_{\mathtt{I}\mathtt{C}}} {\mathbf{R}_{i}}^T\mathbf{p}_{i} \\ & + {\mathbf{R}_{\mathtt{I}\mathtt{C}}} {\mathbf{R}_{i}}^T \mathbf{R}_{i} \mathbf{R}_{\mathtt{D}\mathtt{C}}^T {_\mathtt{D}\mathbf{p}_{\mathtt{D}\mathtt{C}}} = \underbrace{\sum_{k=i}^{j-1}   {\Delta\mathbf{R}_{\mathtt{I}_i\mathtt{I}_k}} \mathbf{R}_{\mathtt{I}\mathtt{D}}\ {(_{\mathtt{D}_i}{\tilde{\mathbf{v}}}} - \bm{\eta}^D) \Delta t}_{{\Delta {_{\mathtt{D}_i}}{\mathbf{p}}_{\mathtt{D}_i\mathtt{D}_j}}}
            \\ \xRightarrow[]{\text{(c)}} &{\mathbf{R}_{\mathtt{I}\mathtt{D}}\mathbf{R}_{\mathtt{D}\mathtt{C}}} {\mathbf{R}_{i}}^T \mathbf{p}_{j} - {\mathbf{R}_{\mathtt{I}\mathtt{D}}\mathbf{R}_{\mathtt{D}\mathtt{C}}} {\mathbf{R}_{i}}^T \mathbf{R}_{j} \mathbf{R}_{\mathtt{D}\mathtt{C}}^T {_\mathtt{D}\mathbf{p}_{\mathtt{D}\mathtt{C}}} - {\mathbf{R}_{\mathtt{I}\mathtt{D}}\mathbf{R}_{\mathtt{D}\mathtt{C}}} {\mathbf{R}_{i}}^T\mathbf{p}_{i} \\ & + {\mathbf{R}_{\mathtt{I}\mathtt{D}}} {_\mathtt{D}\mathbf{p}_{\mathtt{D}\mathtt{C}}} =  {\Delta {_{\mathtt{D}_i}}{\mathbf{p}}_{\mathtt{D}_i\mathtt{D}_j}} \\
            \xRightarrow[]{\text{(d)}} & \underbrace{\mathbf{R}_{\mathtt{I}\mathtt{D}}\big({_\mathtt{D}\mathbf{p}_{\mathtt{D}\mathtt{C}}}-\mathbf{R}_{\mathtt{D}\mathtt{C}}\mathbf{R}_{i}^T\mathbf{R}_{j}\mathbf{R}_{\mathtt{D}\mathtt{C}}^T{_\mathtt{D}\mathbf{p}_{\mathtt{D}\mathtt{C}}} + \mathbf{R}_{\mathtt{D}\mathtt{C}} (\mathbf{R}_{i}^T\mathbf{p}_{j}-\mathbf{R}_{i}^T\mathbf{p}_{i})\big) }_{h_{D_t}(\mathcal{X}_i,\mathcal{X}_j)} \\&=  {\Delta {_{\mathtt{D}_i}}{\mathbf{p}}_{\mathtt{D}_i\mathtt{D}_j}}
        \label{equa:dvl_translation_measurement_derivation}
        \end{align}}
    Step (a) moves $\mathbf{p}_{i}   - \mathbf{R}_{i} \mathbf{R}_{\mathtt{D}\mathtt{C}}^T {_\mathtt{D}\mathbf{p}_{\mathtt{D}\mathtt{C}}}$ to the left side of the equation. Step (b) multiplies ${\mathbf{R}_{\mathtt{I}\mathtt{C}}} {\mathbf{R}_{i}}^T$ at the both sides of the equation. Step (c) substitutes ${\mathbf{R}_{\mathtt{I}\mathtt{C}}}$ with ${\mathbf{R}_{\mathtt{I}\mathtt{D}}\mathbf{R}_{\mathtt{D}\mathtt{C}}}$ as defined at \eqref{equa:extrinsic_IC}. Step (d) reformulates the equation as the DVL translation measurement model defined at \eqref{equa:dvl_decoupled}.
\subsection{DVL Pre-integration Derivation}
\label{sec:appendix DVL Preintegration Derivation}
    The relative DVL translation incremental $\Delta {_{\mathtt{D}_i}}{\mathbf{p}}_{\mathtt{D}_i\mathtt{D}_j}$ can be further reformulated as:
    % \begin{equation}
    {\allowdisplaybreaks
        \begin{align}
            & \Delta {_{\mathtt{D}_i}}{\mathbf{p}}_{\mathtt{D}_i\mathtt{D}_j} = {\sum_{k=i}^{j-1}} {{\Delta \mathbf{R}_{\mathtt{I}_i\mathtt{I}_k}}} \mathbf{R}_{\mathtt{I}\mathtt{D}}\ ({_{\mathtt{D}_i}{\tilde{\mathbf{v}}} - \bm{\eta}^D}) \Delta t \\
            \stackrel{\text{(a)}}{\approx}  & \sum_{k=i}^{j-1} {\Delta{\hat{\mathbf{R}}}_{\mathtt{I}_i\mathtt{I}_k}} (\mathbf{I} - \delta{\hat{\bm{\phi}}}_{\mathtt{I}_i\mathtt{I}_k}^\land) \mathbf{R}_{\mathtt{I}\mathtt{D}}\ {_{\mathtt{D}_i}{\tilde{\mathbf{v}}}} \Delta t - {\Delta{\hat{\mathbf{R}}}_{\mathtt{I}_i\mathtt{I}_k}} \mathbf{R}_{\mathtt{I}\mathtt{D}}\ {\bm{\eta}^D} \Delta t \\
            \stackrel{\text{(b)}}{=} & \sum_{k=i}^{j-1} {\Delta{\hat{\mathbf{R}}}_{\mathtt{I}_i\mathtt{I}_k}} \mathbf{R}_{\mathtt{I}\mathtt{D}}\ {_{\mathtt{D}_i}{\tilde{\mathbf{v}}}} \Delta t \\
            & -  {\Delta{\hat{\mathbf{R}}}_{\mathtt{I}_i\mathtt{I}_k}} \delta{\hat{\bm{\phi}}}_{\mathtt{I}_i\mathtt{I}_k}^\land  \mathbf{R}_{\mathtt{I}\mathtt{D}}\ {_{\mathtt{D}_i}{\tilde{\mathbf{v}}}} \Delta t - {\Delta{\hat{\mathbf{R}}}_{\mathtt{I}_i\mathtt{I}_k}} \mathbf{R}_{\mathtt{I}\mathtt{D}}\ {\bm{\eta}^D} \Delta t \\
            \stackrel{\text{(c)}}{=} & \underbrace{\sum_{k=i}^{j-1} {\Delta{\hat{\mathbf{R}}}_{\mathtt{I}_i\mathtt{I}_k}} \mathbf{R}_{\mathtt{I}\mathtt{D}}\ {_{\mathtt{D}_i}{\tilde{\mathbf{v}}}} \Delta t}_{\Delta {_{\mathtt{D}_i}}{\bar{\mathbf{p}}}_{\mathtt{D}_i\mathtt{D}_j}}  - \\
            & \underbrace{\sum_{k=i}^{j-1} [-{\Delta{\hat{\mathbf{R}}}_{\mathtt{I}_i\mathtt{I}_k}} (\mathbf{R}_{\mathtt{I}\mathtt{D}}\ {_{\mathtt{D}_i}{\tilde{\mathbf{v}}}})^\land \cdot \delta{\hat{\bm{\phi}}}_{\mathtt{I}_i\mathtt{I}_k} \Delta t  + {\Delta{\hat{\mathbf{R}}}_{\mathtt{I}_i\mathtt{I}_k}} \mathbf{R}_{\mathtt{I}\mathtt{D}}\ {\bm{\eta}^D} \Delta t}_{\delta {_{\mathtt{D}_i}}{\bar{\mathbf{p}}}_{\mathtt{D}_i\mathtt{D}_j}} ]
        \label{equa:dvl_preintegration_derivation}
        \end{align}
    % \end{equation}
    }
    Step (a) first replaces ${\Delta \mathbf{R}_{\mathtt{I}_i\mathtt{I}_k}}$ with ${\Delta{\hat{\mathbf{R}}}_{\mathtt{I}_i\mathtt{I}_k}} \text{Exp}( -\delta{\hat{\bm{\phi}}}_{\mathtt{I}_i\mathtt{I}_k})$ as in \eqref{equa:gyros_integration2} then applies the first order approximation of $\text{Exp}( -\delta{\hat{\bm{\phi}}}_{\mathtt{I}_i\mathtt{I}_k})$ as $\text{Exp}( \bm{\theta}) \approx (\mathbf{I}+\bm{\theta}^\land)$ \cite{forster2016manifold}. Step (b) expands the brace and reformulates the equation. Step (c) applies the property of $\bm{u}^\land\bm{v} = -\bm{v}^\land\bm{u}$ and reformulates to $\Delta {_{\mathtt{D}_i}}{\bar{\mathbf{p}}}_{\mathtt{D}_i\mathtt{D}_j}$ and $\delta {_{\mathtt{D}_i}}{\bar{\mathbf{p}}}_{\mathtt{D}_i\mathtt{D}_j}$ defined in \eqref{equa:dvl_decoupled}.
\subsection{Jacobian of Extrinsic Calibration Approximation Iteration Derivation}
    \label{sec:Jacobian of Extrinsic Calibration Approximation Iteration Derivation}
    Following the definition of derivative, we have
    {\allowdisplaybreaks
    \begin{align}
            \frac{\partial {\Delta {_{\mathtt{D}_i}}{\bar{\mathbf{p}}}_{\mathtt{D}_i\mathtt{D}_j}} }{\partial {\bm{\phi}_{\mathtt{I}\mathtt{D}}}} =  & \lim_{\Delta {\bm{\phi}_{\mathtt{I}\mathtt{D}}} \to 0}  \frac{ \left[\begin{aligned} & \sum_{k=i}^{j-1} {\Delta{\hat{\mathbf{R}}}_{\mathtt{I}_i\mathtt{I}_k}} \text{Exp}(\Delta {\bm{\phi}_{\mathtt{I}\mathtt{D}}}) \mathbf{R}_{\mathtt{I}\mathtt{D}}\ {_{\mathtt{D}_i}{\tilde{\mathbf{v}}}} \Delta t \\ & -  {\Delta{\hat{\mathbf{R}}}_{\mathtt{I}_i\mathtt{I}_k}} \mathbf{R}_{\mathtt{I}\mathtt{D}}\ {_{\mathtt{D}_i}{\tilde{\mathbf{v}}}} \Delta t
            \end{aligned} \right]}{\Delta {\bm{\phi}_{\mathtt{I}\mathtt{D}}}} \\ \stackrel{\text{(a)}}{\approx} &\lim_{\Delta {\bm{\phi}_{\mathtt{I}\mathtt{D}}} \to 0} \frac{ \left[\begin{aligned} & \sum_{k=i}^{j-1} {\Delta{\hat{\mathbf{R}}}_{\mathtt{I}_i\mathtt{I}_k}} (\mathbf{I} + {\Delta \bm{\phi}_{\mathtt{I}\mathtt{D}}^\land}) \mathbf{R}_{\mathtt{I}\mathtt{D}}\ {_{\mathtt{D}_i}{\tilde{\mathbf{v}}}} \Delta t \\ & - {\Delta{\hat{\mathbf{R}}}_{\mathtt{I}_i\mathtt{I}_k}} \mathbf{R}_{\mathtt{I}\mathtt{D}}\ {_{\mathtt{D}_i}{\tilde{\mathbf{v}}}} \Delta t
            \end{aligned} \right]}{\Delta {\bm{\phi}_{\mathtt{I}\mathtt{D}}}} \\ \stackrel{\text{(b)}}{=} & \lim_{\Delta {\bm{\phi}_{\mathtt{I}\mathtt{D}}} \to 0} \frac{\sum_{k=i}^{j-1} {\Delta{\hat{\mathbf{R}}}_{\mathtt{I}_i\mathtt{I}_k}}  {\Delta \bm{\phi}_{\mathtt{I}\mathtt{D}}^\land} \mathbf{R}_{\mathtt{I}\mathtt{D}}\ {_{\mathtt{D}_i}{\tilde{\mathbf{v}}}} \Delta t}{\Delta {\bm{\phi}_{\mathtt{I}\mathtt{D}}}} \\ \stackrel{\text{(c)}}{=}& \lim_{\Delta {\bm{\phi}_{\mathtt{I}\mathtt{D}}} \to 0} \frac{\sum_{k=i}^{j-1} -{\Delta{\hat{\mathbf{R}}}_{\mathtt{I}_i\mathtt{I}_k}} (\mathbf{R}_{\mathtt{I}\mathtt{D}}\ {_{\mathtt{D}_i}{\tilde{\mathbf{v}}}})^\land {\Delta {\bm{\phi}_{\mathtt{I}\mathtt{D}}}}  \Delta t}{\Delta {\bm{\phi}_{\mathtt{I}\mathtt{D}}}} \\ \stackrel{\text{(d)}}{=}&  \sum_{k=i}^{j-1} -{\Delta{\hat{\mathbf{R}}}_{\mathtt{I}_i\mathtt{I}_k}} (\mathbf{R}_{\mathtt{I}\mathtt{D}}\ {_{\mathtt{D}_i}{\tilde{\mathbf{v}}}})^\land  \Delta t
        \label{equa:extrinsic_calibration_approximation_iteration_derivation}
        \end{align}
    }
    We assume left multiplication update is adopted during iteration and $\text{Exp}(\Delta {\bm{\phi}_{\mathtt{I}\mathtt{D}}})$ stand for the incremental update. Step (a) applies the first order approximation of $\text{Exp}(\Delta {\bm{\phi}_{\mathtt{I}\mathtt{D}}})$. Step (b) applies $\bm{u}^\land\bm{v} = -\bm{v}^\land\bm{u}$. Step (c) and (d) reformulate the equation as the form in \eqref{equa:extrinsic_linearization}.
\subsection{Jacobian of Misalignment Calibration Approximation Iteration Derivation}
    \label{sec:Jacobian of Misalignment Calibration Approximation Iteration Derivation}
    We have the below derivation for \eqref{equa:misalignment_linearization}:
    \begin{equation}
        \begin{aligned}
            \frac{\partial {\Delta {_{\mathtt{D}_i}}{\bar{\mathbf{p}}}_{\mathtt{D}_i\mathtt{D}_j}} }{\partial {_{\mathtt{D}_i}{\tilde{\mathbf{v}}}}} = & \lim_{{\Delta {\tilde{\mathbf{v}}}} \to 0} \frac{\left[ \begin{aligned} &\sum_{k=i}^{j-1} {\Delta{\hat{\mathbf{R}}}_{\mathtt{I}_i\mathtt{I}_k}} \mathbf{R}_{\mathtt{I}\mathtt{D}}\ ({_{\mathtt{D}_i}{\tilde{\mathbf{v}}} + {\Delta {\tilde{\mathbf{v}}}}}) \Delta t \\ & -  {\Delta{\hat{\mathbf{R}}}_{\mathtt{I}_i\mathtt{I}_k}} \mathbf{R}_{\mathtt{I}\mathtt{D}}\ {_{\mathtt{D}_i}{\tilde{\mathbf{v}}}} \Delta t
            \end{aligned} \right]}{{\Delta {\tilde{\mathbf{v}}}}} \\ \stackrel{\text{(a)}}{=} & \lim_{{\Delta {\tilde{\mathbf{v}}}} \to 0} \frac{\sum_{k=i}^{j-1} {\Delta{\hat{\mathbf{R}}}_{\mathtt{I}_i\mathtt{I}_k}} \mathbf{R}_{\mathtt{I}\mathtt{D}}\  {\Delta {\tilde{\mathbf{v}}}} \Delta t }{{\Delta {\tilde{\mathbf{v}}}}} \\ \stackrel{\text{(b)}}{=} & \sum_{k=i}^{j-1} {\Delta{\hat{\mathbf{R}}}_{\mathtt{I}_i\mathtt{I}_k}} \mathbf{R}_{\mathtt{I}\mathtt{D}}\ \Delta t
        \end{aligned}
        \label{equa:misalignment_calibration_approximation_iteration_derivation}
    \end{equation}
% Step (a) and (b) simply reformulate the equation to the format .

\section*{Acknowledgment}
The authors would like to thank Dr Jonatan Scharff Willners, Joshua Roe and Sean Katagiri for their support on the data collection and hardware.
% \newpage

%

\bibliography{bib/bibtran}

\begin{thebibliography}{10}
\providecommand{\url}[1]{#1}
\csname url@samestyle\endcsname
\providecommand{\newblock}{\relax}
\providecommand{\bibinfo}[2]{#2}
\providecommand{\BIBentrySTDinterwordspacing}{\spaceskip=0pt\relax}
\providecommand{\BIBentryALTinterwordstretchfactor}{4}
\providecommand{\BIBentryALTinterwordspacing}{\spaceskip=\fontdimen2\font plus
\BIBentryALTinterwordstretchfactor\fontdimen3\font minus
  \fontdimen4\font\relax}
\providecommand{\BIBforeignlanguage}[2]{{%
\expandafter\ifx\csname l@#1\endcsname\relax
\typeout{** WARNING: IEEEtran.bst: No hyphenation pattern has been}%
\typeout{** loaded for the language `#1'. Using the pattern for}%
\typeout{** the default language instead.}%
\else
\language=\csname l@#1\endcsname
\fi
#2}}
\providecommand{\BIBdecl}{\relax}
\BIBdecl

\bibitem{mur2017orb2}
R.~Mur-Artal and J.~D. Tard{\'o}s, ``{ORB-SLAM2}: An open-source {SLAM} system
  for monocular, stereo, and rgb-d cameras,'' \emph{IEEE Transactions on
  Robotics}, vol.~33, no.~5, pp. 1255--1262, 2017.

\bibitem{engel2017dso}
J.~Engel, V.~Koltun, and D.~Cremers, ``Direct sparse odometry,'' \emph{IEEE
  Transactions on Pattern Analysis and Machine Intelligence}, vol.~40, no.~3,
  pp. 611--625, 2017.

\bibitem{luo2022hybrid}
D.~Luo, Y.~Zhuang, and S.~Wang, ``Hybrid sparse monocular visual odometry with
  online photometric calibration,'' \emph{The International Journal of Robotics
  Research}, vol.~41, no. 11-12, pp. 993--1021, 2022.

\bibitem{lin2018VINS-Mono}
T.~Qin, P.~Li, and S.~Shen, ``{VINS-Mono}: A robust and versatile monocular
  visual-inertial state estimator,'' \emph{IEEE Transactions on Robotics},
  vol.~34, no.~4, pp. 1004--1020, 2018.

\bibitem{orbslam3}
C.~Campos, R.~Elvira, J.~J.~G. Rodr{\'\i}guez, J.~M. Montiel, and J.~D.
  Tard{\'o}s, ``{ORB-SLAM3}: An accurate open-source library for visual,
  visual-inertial, and multimap {SLAM},'' \emph{IEEE Transactions on Robotics},
  vol.~37, no.~6, pp. 1874--1890, 2021.

\bibitem{rahman2022svin2}
S.~Rahman, A.~Quattrini~Li, and I.~Rekleitis, ``Svin2: A multi-sensor
  fusion-based underwater {SLAM} system,'' \emph{The International Journal of
  Robotics Research}, vol.~41, no. 11-12, pp. 1022--1042, 2022.

\bibitem{Rudolph2012dvl_model}
D.~Rudolph and T.~A. Wilson, ``Doppler velocity log theory and preliminary
  considerations for design and construction,'' in \emph{2012 Proceedings of
  IEEE Southeastcon}, 2012, pp. 1--7.

\bibitem{huang2023tightly_visualdvl}
Y.~Huang, P.~Li, S.~Yan, Y.~Ou, Z.~Wu, M.~Tan, and J.~Yu, ``Tightly-coupled
  visual-dvl fusion for accurate localization of underwater robots,'' in
  \emph{2023 IEEE/RSJ International Conference on Intelligent Robots and
  Systems (IROS)}.\hskip 1em plus 0.5em minus 0.4em\relax IEEE, 2023, pp.
  8090--8095.

\bibitem{zhao2023tightly_icewater}
L.~Zhao, M.~Zhou, and B.~Loose, ``Tightly-coupled visual-dvl-inertial odometry
  for robot-based ice-water boundary exploration,'' in \emph{2023 IEEE/RSJ
  International Conference on Intelligent Robots and Systems (IROS)}.\hskip 1em
  plus 0.5em minus 0.4em\relax IEEE, 2023, pp. 7127--7134.

\bibitem{Thoms2023dvl_slam}
A.~Thoms, G.~Earle, N.~Charron, and S.~Narasimhan, ``Tightly coupled,
  graph-based dvl/imu fusion and decoupled mapping for {SLAM}-centric maritime
  infrastructure inspection,'' \emph{IEEE Journal of Oceanic Engineering},
  vol.~48, no.~3, pp. 663--676, 2023.

\bibitem{vargas2021robust}
E.~Vargas, R.~Scona, J.~S. Willners, T.~Luczynski, Y.~Cao, S.~Wang, and Y.~R.
  Petillot, ``Robust underwater visual {SLAM} fusing acoustic sensing,'' in
  \emph{2021 IEEE International Conference on Robotics and Automation}.\hskip
  1em plus 0.5em minus 0.4em\relax IEEE, 2021, pp. 2140--2146.

\bibitem{xu2021underwater}
S.~Xu, T.~Luczynski, J.~S. Willners, Z.~Hong, K.~Zhang, Y.~R. Petillot, and
  S.~Wang, ``Underwater visual acoustic {SLAM} with extrinsic calibration,'' in
  \emph{IEEE/RSJ International Conference on Intelligent Robots and
  Systems}.\hskip 1em plus 0.5em minus 0.4em\relax IEEE, 2021, pp. 7647--7652.

\bibitem{strobl2006handeye}
K.~H. Strobl and G.~Hirzinger, ``Optimal hand-eye calibration,'' in
  \emph{IEEE/RSJ International Conference on Intelligent Robots and
  Systems}.\hskip 1em plus 0.5em minus 0.4em\relax IEEE, 2006, pp. 4647--4653.

\bibitem{xu2022dvl_calibration}
B.~Xu and Y.~Guo, ``A novel dvl calibration method based on robust invariant
  extended kalman filter,'' \emph{IEEE Transactions on Vehicular Technology},
  vol.~71, no.~9, pp. 9422--9434, 2022.

\bibitem{fu2022dvl_calibration}
Q.~Fu, Q.~Shen, D.~Wei, F.~Wu, and G.~Yan, ``Multiposition alignment for
  rotational ins based on real-time estimation of inner lever arms,''
  \emph{IEEE Transactions on Instrumentation and Measurement}, vol.~71, pp.
  1--8, 2022.

\bibitem{underwater_vslam_ekf1}
R.~Eustice, O.~Pizarro, and H.~Singh, ``Visually augmented navigation in an
  unstructured environment using a delayed state history,'' in \emph{IEEE
  International Conference on Robotics and Automation}, vol.~1, 2004, pp.
  25--32 Vol.1.

\bibitem{ozog2013dvl_camera_piecewise_planar}
P.~Ozog and R.~M. Eustice, ``Real-time {SLAM} with piecewise-planar surface
  models and sparse 3d point clouds,'' in \emph{IEEE/RSJ International
  Conference on Intelligent Robots and Systems}.\hskip 1em plus 0.5em minus
  0.4em\relax IEEE, 2013, pp. 1042--1049.

\bibitem{kim2013hull_inspection}
A.~Kim and R.~M. Eustice, ``Real-time visual {SLAM} for autonomous underwater
  hull inspection using visual saliency,'' \emph{IEEE Transactions on
  Robotics}, vol.~29, no.~3, pp. 719--733, 2013.

\bibitem{underwater_vslam_graph2}
S.~Hong and J.~Kim, ``Three-dimensional visual mapping of underwater ship hull
  surface using piecewise-planar {SLAM},'' \emph{International Journal of
  Control, Automation and Systems}, vol.~18, pp. 564--574, 2020.

\bibitem{westman2018camera_dvl_extrinsic}
E.~Westman and M.~Kaess, ``Underwater {AprilTag SLAM} and calibration for high
  precision robot localization,'' \emph{Technical Report}, 2018.

\bibitem{rahman2018svin}
S.~Rahman, A.~Q. Li, and I.~Rekleitis, ``Sonar visual inertial {SLAM} of
  underwater structures,'' in \emph{IEEE International Conference on Robotics
  and Automation}.\hskip 1em plus 0.5em minus 0.4em\relax IEEE, 2018, pp.
  5190--5196.

\bibitem{okvis}
S.~Leutenegger, P.~Furgale, V.~Rabaud, M.~Chli, K.~Konolige, and R.~Siegwart,
  ``Keyframe-based visual-inertial {SLAM} using nonlinear optimization,''
  \emph{Proceedings of Robotis Science and Systems}, 2013.

\bibitem{rahman2019svin2}
S.~Rahman, A.~Q. Li, and I.~Rekleitis, ``Svin2: an underwater {SLAM} system
  using sonar, visual, inertial, and depth sensor,'' in \emph{IEEE/RSJ
  International Conference on Intelligent Robots and Systems}.\hskip 1em plus
  0.5em minus 0.4em\relax IEEE, 2019, pp. 1861--1868.

\bibitem{joshi2023switching}
B.~Joshi, H.~Damron, S.~Rahman, and I.~Rekleitis, ``Sm/vio: Robust underwater
  state estimation switching between model-based and visual inertial
  odometry,'' in \emph{2023 IEEE International Conference on Robotics and
  Automation (ICRA)}.\hskip 1em plus 0.5em minus 0.4em\relax IEEE, 2023, pp.
  5192--5199.

\bibitem{gu2019camera_imu_cali_underwater}
C.~Gu, Y.~Cong, and G.~Sun, ``Environment driven underwater camera-{IMU}
  calibration for monocular visual-inertial {SLAM},'' in \emph{International
  Conference on Robotics and Automation}.\hskip 1em plus 0.5em minus
  0.4em\relax IEEE, 2019, pp. 2405--2411.

\bibitem{yang2020camera_sonar_extrinsic}
D.~Yang, B.~He, M.~Zhu, and J.~Liu, ``An extrinsic calibration method with
  closed-form solution for underwater opti-acoustic imaging system,''
  \emph{IEEE Transactions on Instrumentation and Measurement}, vol.~69, no.~9,
  pp. 6828--6842, 2020.

\bibitem{li2022dvl_calibration}
D.~Li, J.~Xu, B.~Zhu, and H.~He, ``A calibration method of {DVL} in integrated
  navigation system based on particle swarm optimization,'' \emph{Measurement},
  vol. 187, p. 110325, 2022.

\bibitem{Luo2023dvl_calibration}
L.~Luo, Y.~Huang, G.~Wang, Y.~Zhang, and L.~Tang, ``An on-line full-parameters
  calibration method for {SINS/DVL} integrated navigation system,'' \emph{IEEE
  Sensors Journal}, pp. 1--1, 2023.

\bibitem{furgale2014}
P.~Furgale, ``{Representing Robot Pose: The good, the bad, and the ugly},''
  \url{https://paulfurgale.info/news/2014/6/9/representing-robot-pose-the-good-the-bad-and-the-ugly}.

\bibitem{Brokloff1994dvl_model}
N.~Brokloff, ``Matrix algorithm for doppler sonar navigation,'' in
  \emph{Proceedings of OCEANS'94}, vol.~3, 1994, pp. III/378--III/383 vol.3.

\bibitem{forster2016manifold}
C.~Forster, L.~Carlone, F.~Dellaert, and D.~Scaramuzza, ``On-manifold
  preintegration for real-time visual--inertial odometry,'' \emph{IEEE
  Transactions on Robotics}, vol.~33, no.~1, pp. 1--21, 2016.

\bibitem{triggs2000bundle}
B.~Triggs, P.~F. McLauchlan, R.~I. Hartley, and A.~W. Fitzgibbon, ``Bundle
  adjustment—a modern synthesis,'' in \emph{International Workshop on Vision
  Algorithms Corfu, Greece}.\hskip 1em plus 0.5em minus 0.4em\relax Springer,
  2000, pp. 298--372.

\bibitem{xu2023observability}
S.~Xu, J.~S. Willners, Z.~Hong, K.~Zhang, Y.~R. Petillot, and S.~Wang,
  ``Observability-aware active extrinsic calibration of multiple sensors,'' in
  \emph{IEEE International Conference on Robotics and Automation}.\hskip 1em
  plus 0.5em minus 0.4em\relax IEEE, 2023, pp. 2091--2097.

\bibitem{luczynski2019stereo}
T.~{\L}uczy{\'n}ski, P.~{\L}uczy{\'n}ski, L.~Pehle, M.~Wirsum, and A.~Birk,
  ``Model based design of a stereo vision system for intelligent deep-sea
  operations,'' \emph{Measurement}, vol. 144, pp. 298--310, 2019.

\bibitem{wang2016iros_april}
J.~Wang and E.~Olson, ``{AprilTag} 2: Efficient and robust fiducial
  detection,'' in \emph{Proceedings of the {IEEE/RSJ} International Conference
  on Intelligent Robots and Systems}, October 2016.

\bibitem{schoenberger2016sfm}
J.~L. Sch\"{o}nberger and J.-M. Frahm, ``Structure-from-motion revisited,'' in
  \emph{Conference on Computer Vision and Pattern Recognition (CVPR)}, 2016.

\bibitem{usenko2019basalt}
V.~Usenko, N.~Demmel, D.~Schubert, J.~St{\"u}ckler, and D.~Cremers,
  ``Visual-inertial mapping with non-linear factor recovery,'' \emph{IEEE
  Robotics and Automation Letters}, vol.~5, no.~2, pp. 422--429, 2019.

\bibitem{Manhaes2016_uuvsim}
\BIBentryALTinterwordspacing
M.~M.~M. Manh{\~{a}}es, S.~A. Scherer, M.~Voss, L.~R. Douat, and
  T.~Rauschenbach, ``{UUV} simulator: A gazebo-based package for underwater
  intervention and multi-robot simulation,'' in \emph{{OCEANS} 2016
  {MTS}/{IEEE} Monterey}.\hskip 1em plus 0.5em minus 0.4em\relax {IEEE}, sep
  2016. [Online]. Available:
  \url{https://doi.org/10.1109%2Foceans.2016.7761080}
\BIBentrySTDinterwordspacing

\end{thebibliography}
\bibliographystyle{IEEEtran}

\end{document}